\documentclass{article}
\usepackage{xcolor}

\usepackage{iclr2025_conference,times}

\usepackage{amsmath,amsfonts,bm}
\usepackage{multicol}

\def\eqref#1{equation~\ref{#1}}

\def\1{\bm{1}}

\def\vtheta{{\bm{\theta}}}

\def\vo{{\bm{o}}}

\def\vq{{\bm{q}}}

\DeclareMathAlphabet{\mathsfit}{\encodingdefault}{\sfdefault}{m}{sl}
\SetMathAlphabet{\mathsfit}{bold}{\encodingdefault}{\sfdefault}{bx}{n}

\usepackage{amsmath,amssymb}          
\usepackage{graphicx}                 
\usepackage{subcaption}               
\usepackage{booktabs,multirow,makecell,colortbl}  
\usepackage{xcolor}                   
\usepackage{xspace}                   
\usepackage{enumitem}                 
\usepackage[most]{tcolorbox}          
\tcbuselibrary{listings,skins,breakable}
\usepackage{wrapfig}                  
\usepackage{marvosym}                 
\usepackage{url}
\usepackage{hyperref}                 
\usepackage{cleveref}                 
\usepackage{lineno}                   
\usepackage{algorithm}                
\usepackage{algpseudocode}            

\definecolor{uclablue}{rgb}{0.15,0.45,0.68}
\hypersetup{
    breaklinks,
    colorlinks=true,
    citecolor=uclablue,
}

\definecolor{table-color}{HTML}{F5FFFA}   
\definecolor{titleblue}{HTML}{5AA1BF}     
\definecolor{lightblue}{HTML}{EAF4FB}     

\newcommand{\myie}{\textit{i.e.,}\xspace}
\newcommand{\myeg}{\textit{e.g.,}\xspace}
\newcommand{\custompara}[1]{{\vspace{1mm}\noindent\textbf{#1}\xspace}}
\robustify\bfseries

\let\titleold\title
\renewcommand{\title}[1]{\titleold{#1}\renewcommand{\thetitle}{#1}}
\def\maketitlesupplementary{%
    {\newpage
            \par\nolinenumbers
            \centering
            \Large
            \textbf{\thetitle}\\
            \vspace{0.3em}
            Supplementary Material \\
            \linenumbers}%
}

\newcommand{\promptsep}{
    \nopagebreak
    \vspace{2pt}
    \textcolor{gray!50}{\rule{\linewidth}{0.2pt}}
    \vspace{1pt}
}

\newcounter{promptcount}
\newtcolorbox{custom_template_box}[2][]{
    enhanced,
    colback=lightblue, colframe=titleblue,
    colbacktitle=titleblue, coltitle=white,
    fonttitle=\bfseries\small,
    boxrule=1pt, arc=2mm, boxsep=4pt,
    left=6pt,right=6pt,top=4pt,bottom=4pt,
    code={\refstepcounter{promptcount}},
    title={#2}, #1
}
\crefname{promptcount}{prompt}{prompts}
\Crefname{promptcount}{Prompt}{Prompts}

\crefname{equation}{Eq.}{Eqs.}    \Crefname{equation}{Eq.}{Eqs.}
\crefname{table}{Table}{Tables}     \Crefname{table}{Table}{Tables}
\crefname{figure}{Fig.}{Figs.}    \Crefname{figure}{Fig.}{Figs.}
\creflabelformat{equation}{#2(#1)#3}
\creflabelformat{table}{#2#1#3}
\creflabelformat{figure}{#2#1#3}
\crefformat{section}{\S#2#1#3}
\Crefformat{section}{\S#2#1#3}
\crefmultiformat{section}{\S#2#1#3}{, \S#2#1#3}{, \S#2#1#3}{, \S#2#1#3}

\algrenewcommand\algorithmiccomment[1]{\hfill$\triangleright$~\textit{\small#1}}
\newcommand{\mycomment}[1]{\hfill$\triangleright$~\textit{\small #1}}     
\newcommand{\mylinecomment}[1]{$\triangleright$~\textit{\small #1}}       

\DeclareMathOperator*{\mean}{mean}

\newcommand{\nextline}{\\}

\makeatletter
\def\@maketitle{
    \vbox{
        \hsize\textwidth
        {\centering {\Large\bf \@title\par}}
        \begingroup
        \leftskip=0pt \rightskip=0pt \parindent=0pt
        \rule{\z@}{6pt}{\centering\par{\@author}\par}\vskip 0.5em
        \endgroup
    }
    \thispagestyle{firstpage}
}
\makeatother

\title{\texttt{Uni-OPD}: Unifying On-Policy Distillation with a Dual-Perspective Recipe}

\author{
    Wenjin Hou$^{1,\,\scalebox{1.1}{$\ast$}}$,
    Shangpin Peng$^{2,\,3,\,\scalebox{1.1}{$\ast$}}$,
    Weinong Wang$^{3,\,\dagger}$,
    Zheng Ruan$^{3}$,
    Yue Zhang$^{1}$,
    Zhenglin Zhou$^{1}$
    \\[2pt]
    Mingqi Gao$^{3}$,
    Yifei Chen$^{3}$,
    Kaiqi Wang$^{3}$,
    Hongming Yang$^{3}$,
    Chengquan Zhang$^{3}$,
    Zhuotao Tian$^{2}$
    \\[2pt]
    Han Hu$^{3,\,\ddagger}$,
    \hspace{0.3em}
    Yi Yang$^{1}$,
    \hspace{0.3em}
    Fei Wu$^{1}$,
    \hspace{0.3em}
    Hehe Fan$^{1,\,}$\textsuperscript{\Letter}
    \\[2pt]
    $^1$\textbf{Zhejiang University}
    \qquad
    $^2$\textbf{Shenzhen Loop Area Institute}
    \qquad
    $^3$\textbf{LLM Department, Tencent}
    \\[2pt]
    {
        \tt\small
        houwj17@gmail.com \qquad
        weinong.wang@hotmail.com \qquad
        hehefan@zju.edu.cn
    }
}

\begin{document}

\maketitle

\let\oldthefootnote\thefootnote
\let\thefootnote\relax\footnotetext{
    \hangindent=1.4em
    $^{\scalebox{1.0}{\hspace{-0.7em} $\ast$}}$Equal contribution. \hspace{1em}$^{\star}$Work was done when Wenjin Hou and Shangpin Peng interned at Tencent.
    \\
    $^{\dagger}$Project leader.
    \hspace{1em}$^{\ddagger}$Project supervisor.
    \hspace{1em}\textsuperscript{\Letter}Corresponding author.
}
\let\thefootnote\oldthefootnote

\begin{abstract}
    On-policy distillation (OPD) has recently emerged as an effective post-training paradigm for consolidating the capabilities of specialized expert models into a single student model.
    Despite its empirical success, the conditions under which OPD yields reliable improvement remain poorly understood.
    In this work, we identify two fundamental bottlenecks that limit effective OPD: insufficient student exploration of informative states and unreliable teacher supervision for student rollouts.
    Building on this insight, we propose \texttt{Uni-OPD}, a unified OPD framework that generalizes across LLMs and MLLMs, centered on a dual-perspective optimization strategy.
    Specifically, from the student's perspective, we adopt two data balancing strategies
    to promote exploration of informative student-generated states during training.
    From the teacher's perspective, we show that reliable supervision hinges on whether aggregated token-level guidance remains order-consistent with the outcome reward.
    To this end, we propose an outcome-guided margin calibration mechanism to restore order consistency between correct and incorrect trajectories.
    We conduct extensive experiments on 5 ability domains and 16 benchmarks covering diverse settings, including single-teacher and multi-teacher distillation across LLMs and MLLMs, strong-to-weak distillation, and cross-modal distillation. Our results verify the effectiveness and versatility of \texttt{Uni-OPD} and provide practical insights into reliable OPD.\footnote{Code is available at \url{https://github.com/WenjinHou/Uni-OPD}.}~\looseness=-1
\end{abstract}

\begin{figure}[h]
    \centering
    \includegraphics[width=1.0\linewidth]{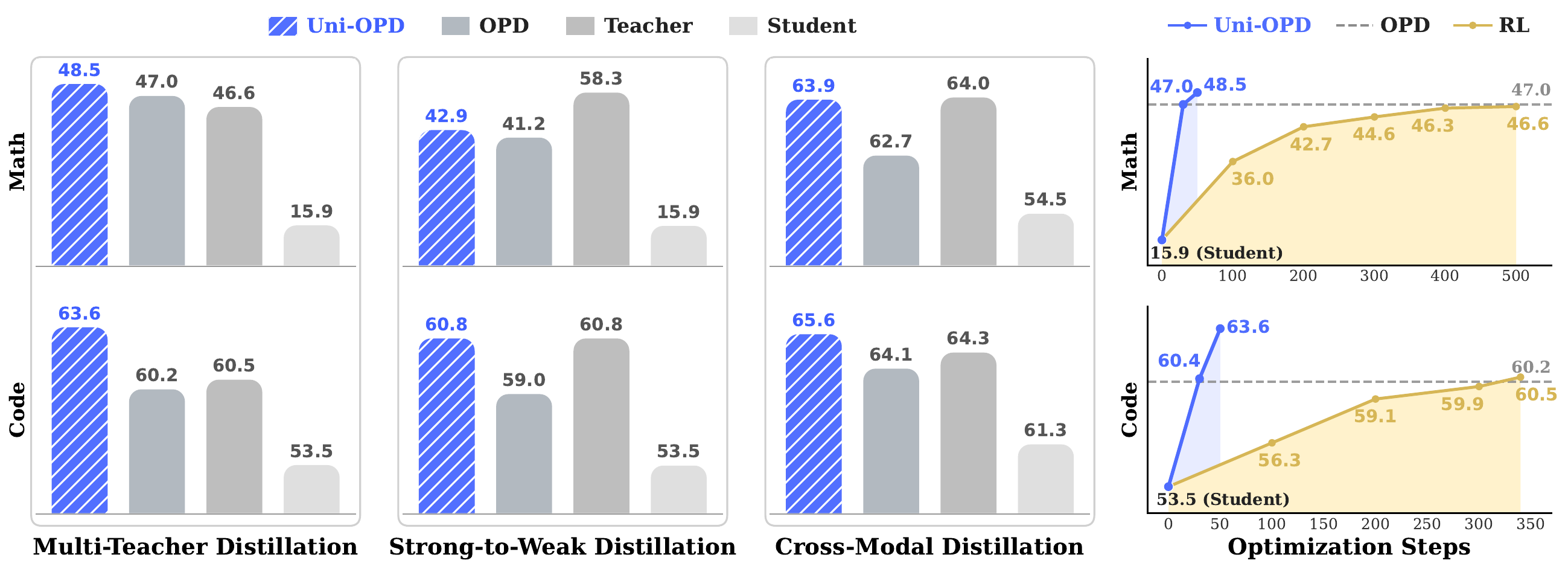}
    \vspace{-1.6em}
    \caption{
        \textbf{Overall performance comparisons and convergence behavior.}
        Results are shown for settings including multi-teacher, strong-to-weak, and cross-modal distillation on math reasoning and code generation tasks.
        \texttt{Uni-OPD} consistently outperforms OPD and converges faster than RL, demonstrating its effectiveness across diverse settings.
    }
    \label{fig:teaser}
\end{figure}

\section{Introduction}
\label{sec: introduction}

Injecting complex reasoning abilities, domain knowledge, and human preferences into LLMs and MLLMs remains a core challenge in the post-training stage.
Conventional approaches typically follow a two-stage paradigm: supervised fine-tuning (SFT) first, followed by reinforcement learning (RL)~\citep{Deepseek_R1_2025, xu2025redstar, GLM_5_2026, zhao2026large}.
While SFT leverages expert data for training, its inherently off-policy nature introduces substantial exposure bias~\citep{qin2025survey,song2026survey}.
Entering rarely covered erroneous states during inference may lead to compounding errors.
Alternatively, on-policy RL (\myeg GRPO~\citep{shao2024deepseekmath}) alleviates distribution shift through online sampling.
However, it mainly relies on sequence-level or terminal rewards, making fine-grained credit assignment difficult and limiting the stability of long-term training~\citep{team2026kimi}.

Recently, on-policy distillation (OPD) has emerged as a promising post-training paradigm for efficiently transferring the knowledge and capabilities of domain experts into a single, unified model.
It combines the strengths of RL and SFT, namely on-policy sampling and token-level supervision. Concretely, OPD trains the student on its own sampled trajectories with teacher feedback under a reverse KL objective~\citep{lu2025onpolicydistillation, DeepSeek_V4_2026}.

Despite its empirical success, current OPD research remains largely confined to LLM distillation~\citep{zhou2025openonerec, G_OPD_2026,xiao2026mimo, yang2026nemotron, wu2026lightning}.
Although a few recent works extend OPD to MLLMs, they are restricted to limited subsets of tasks within a single modality, such as video~\citep{li2026video} or speech~\citep{cao2026x}.
To this end, we first aim to develop a unified OPD framework for both LLMs and MLLMs, enabling effective knowledge distillation across tasks and modalities.

\custompara{Key observations.}
Beyond unifying the framework, we raise a more fundamental question: \textit{what makes OPD a reliable optimization paradigm?}
We posit that effective OPD depends on two factors. \textbf{First}, the student must sufficiently explore informative states, \myie diverse and appropriately difficult self-generated trajectories.
\textbf{Second}, the teacher's token-level supervision must remain reliable when applied to student rollouts.
In particular, the reliability of token-level guidance is significantly enhanced when its trajectory-level aggregation remains order-consistent with outcome reward (\myie correct trajectories receive higher aggregated scores than incorrect ones).
The outcome reward thus provides a global anchor for calibrating unreliable teacher supervision.
These observations motivate a dual-perspective optimization strategy that jointly improves student exploration and the reliability of teacher signals.

\custompara{Our recipe.}
Building on these insights, we introduce \texttt{Uni-OPD}, a dual-perspective strategy for optimizing OPD from the fundamental roles of the student and the teacher.
In this unified framework, we adopt two complementary data-balancing strategies, namely offline difficulty-aware and online correctness-aware balancing, to promote exploration of informative student-generated states. We further present a novel outcome-guided margin calibration mechanism to obtain reliable teacher supervision. Extensive experiments on LLMs and MLLMs verify our recipe.

To summarize, our contributions are threefold:
\begin{itemize}[
        label=\raisebox{0.5ex}{\tiny$\bullet$},
        leftmargin=1em,
        itemsep=2pt, 
        parsep=0pt, 
        topsep=0pt, 
        partopsep=0pt 
    ]
    \item \textbf{Key bottlenecks of OPD.} We identify two core bottlenecks in OPD: insufficient exploration of informative states and unreliable teacher supervision for student rollouts. Our analysis reveals that reliable teacher supervision largely depends on whether token-level guidance remains order-consistent with the outcome reward.
    \item \textbf{Dual-perspective optimization recipe.} We present a dual-perspective optimization recipe for unified OPD that jointly improves student exploration and teacher supervision. Concretely, we combine offline and online data balancing with an outcome-guided margin calibration mechanism, leading to more effective optimization.
    \item \textbf{Comprehensive experimental validation.} We conduct extensive experiments on 5 domains and 16 benchmarks covering diverse settings, including single-teacher and multi-teacher distillation across LLMs and MLLMs, strong-to-weak distillation, and cross-modal distillation (\myie combining text-only and multimodal tasks). Our results verify the effectiveness and versatility of \texttt{Uni-OPD} and provide practical insights into reliable OPD.
\end{itemize}

\section{Related Work}
\vspace{-2mm}

\custompara{Knowledge distillation for LLMs and MLLMs.}
Knowledge distillation~\citep{hinton2015distilling,xu2024survey} aims to transfer knowledge from a larger teacher model to a smaller student model. Conventional approaches typically rely on \textit{off-policy forward Kullback--Leibler (KL) divergence} on a static dataset to align the student's generation distribution with that of the teacher~\citep{liu2024ddk,guo2025learning,he2025kd,liu2025less,ko2025distillm}. Another line of work treats supervised fine-tuning (SFT) on tokens generated by the teacher as an alternative off-policy distillation strategy for eliciting reasoning capabilities during LLM and MLLM post-training~\citep{Deepseek_R1_2025, zhang2025bee, HoneyBee_2026, zhang2025openmmreasoner, team2026kimi, xiao2026mimo}. Though effective, these off-policy methods essentially imitate the teacher's behavior, limiting the student's ability to surpass the teacher and making the student prone to exposure bias~\citep{song2026survey}.

\custompara{On-policy distillation.}
OPD~\citep{agarwal2024policy,lu2025onpolicydistillation} allows a superior teacher to provide feedback on the student's \textit{on-policy trajectories}. This paradigm effectively alleviates exposure bias and elevates the student's upper performance bound. Owing to these merits, OPD has become an efficient way to merge capabilities from multiple experts into a single student model~\citep{xiao2026mimo, yang2026nemotron}, as well as to support strong-to-weak distillation~\citep{Qwen3_VL_2025, GLM_5_2026}. Building on this paradigm, current studies on OPD have branched into several key directions. From the lens of the teacher, recent work explores teacher-free self-distillation paradigms~\citep{kujanpaa2024efficient, shenfeld2026self, zhao2026self, hubotter2026reinforcement,ye2026policy, zhang2026fast, stein2026gates}, develops black-box OPD methods~\citep{ye2025black, xiong2026ovd}, and facilitates distillation across different model families~\citep{OPD_model_family_2025}. Complementary efforts focus on unified training frameworks~\citep{zhang2026kdflow} and stable optimization strategies~\citep{jin2026entropy,kim2026explain, li2026rethinking, xu2026paced} combined with RL~\citep{yang2026self, qu2026pope, jang2026stable, wang2026openclaw}. Few works extend OPD to multimodal domains~\citep{bousselham2025vold,ko2026scaling,li2026video,cao2026x}.
In this work, we push OPD with a dual-perspective recipe that promotes student exploration and teacher reliability, generalizing across LLMs and MLLMs. More detailed related work is provided in the~\cref{supp:sec:related_work}.
\section{Methodology}
\vspace{-1mm}
\label{sec: methodology}

\begin{figure}
    \centering
    \includegraphics[width=1.0\linewidth]{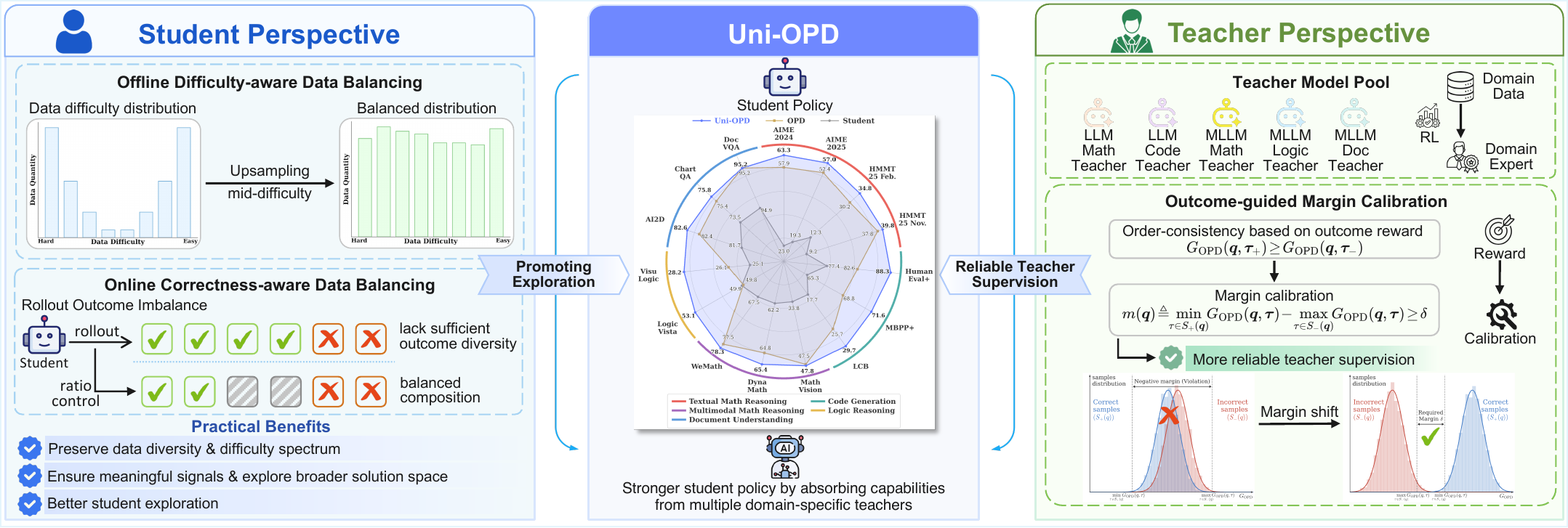}
    \vspace{-1.6em}
    \caption{
        \textbf{Overview of the \texttt{Uni-OPD} framework.}
        (\textit{Left}) Offline difficulty-aware and online correctness-aware data balancing promote student exploration.
        (\textit{Right}) Outcome-guided margin calibration mechanism improves the reliability of teacher supervision.
        (\textit{Middle}) The resulting student policy merges complementary capabilities from multiple domain-specific teachers more effectively than standard OPD, leading to stronger overall performance.
    }
    \label{fig:overview}
    \vspace{-2mm}
\end{figure}

We propose \texttt{Uni-OPD}, a unified framework that advances OPD across LLMs and MLLMs, as shown in~\cref{fig:overview}.
Our design is driven by two fundamental bottlenecks in OPD: \textit{insufficient exploration of informative student-generated states} and \textit{unreliable teacher supervision for student rollouts}.
\texttt{Uni-OPD} addresses them with a dual-perspective recipe that enhances student exploration and calibrates teacher supervision to align with the outcome reward.
We first introduce the preliminaries in~\cref{subsec:preliminary}, followed by an overview of \texttt{Uni-OPD} in~\cref{subsec: Overview}.
We then detail the exploration strategy in~\cref{subsec: Student_Perspective} and the supervision calibration mechanism in~\cref{subsec:Teacher_Perspective}.

\subsection{Preliminaries}
\label{subsec:preliminary}

\custompara{On-policy distillation.}
OPD retains the \textit{on-policy} nature of optimization while providing \textit{token-level credit assignment}, enabling effective post-training.
During training, the student policy $\pi_{\vtheta}$ samples its trajectories and is optimized by minimizing the reverse Kullback-Leibler (KL) divergence to the teacher policy $\pi_{\mathrm{T}}$ over these samples:

\vspace{-1.0em}
\begin{equation}
    \label{eq: OPD_objective}
    \mathcal{J}_{\text{OPD}}(\boldsymbol{\theta})
    =
    \min_{\boldsymbol{\theta}}
    \mathbb{E}_{\boldsymbol{q} \sim D,\, \boldsymbol{\tau} \sim {\pi}_{\boldsymbol{\theta}}(\cdot \mid \boldsymbol{q})}
    \!\Big[\mathcal{D}_{\mathrm{KL}}\!\Big(
        {\pi}_{\boldsymbol{\theta}}(\boldsymbol{\tau} \!\mid\! \boldsymbol{q})
        \,\big\|\,\pi_{\text{T}}(\boldsymbol{\tau} \!\mid\! \boldsymbol{q})
        \Big)\Big],
\end{equation}
\par\nobreak\vspace{-0.5em}

where $\vq$ is the input question, $\boldsymbol{\tau} = (o_1, \dots, o_{|\boldsymbol{\tau}|})$ is a trajectory sampled by the student, $o_t$ is the token at decoding step $t$, and $|\boldsymbol{\tau}|$ is the length of the trajectory. The gradient of OPD can be derived as:

\vspace{-1.0em}
\begin{equation}
    \label{eq: OPD_gradient}
    \nabla_{\boldsymbol{\theta}}
    \mathcal{J}_{\text{OPD}}(\boldsymbol{\theta})
    =
    \mathbb{E}_{\boldsymbol{q} \sim D,\, \boldsymbol{\tau} \sim {\pi}_{\boldsymbol{\theta}}(\cdot \mid \boldsymbol{q})}
    \!\Big[
        \sum_{t=1}^{|\boldsymbol{\tau}|}\!
        \big(
        \log {\pi}_{\boldsymbol{\theta}}(o_{t} \mid \boldsymbol{q}, \boldsymbol{o}_{<t})
        -
        \log \pi_{\text{T}}(o_{t} \mid \boldsymbol{q}, \boldsymbol{o}_{<t})
        \big)
        \,\nabla_{\boldsymbol{\theta}} \log {\pi}_{\boldsymbol{\theta}}(o_{t} \mid \boldsymbol{q}, \boldsymbol{o}_{<t})
        \Big],
\end{equation}
\par\nobreak\vspace{-0.5em}

where $\vo_{<t}$ denotes the prefix before step $t$. The gradient naturally induces a token-level reward at step $t$, analogous to standard RL:

\vspace{-1.0em}
\begin{equation}
    r^{\mathrm{OPD}}_{t}
    =
    \log \pi_{\text{T}}(o_{t} \mid \boldsymbol{q}, \boldsymbol{o}_{<t})
    -
    \log {\pi}_{\boldsymbol{\theta}}(o_{t} \mid \boldsymbol{q}, \boldsymbol{o}_{<t})
    =
    \log \frac{\pi_{\text{T}}(o_{t} \mid \boldsymbol{q}, \boldsymbol{o}_{<t})}
    {{\pi}_{\boldsymbol{\theta}}(o_{t} \mid \boldsymbol{q}, \boldsymbol{o}_{<t})}.
    \label{eq: OPD_token_reward}
\end{equation}
\par\nobreak\vspace{-0.5em}

This formulation provides fine-grained credit assignment signals at the token level.

\custompara{Analyzing teacher supervision in OPD.}
As shown in~\cref{eq: OPD_token_reward}, OPD relies on the teacher to provide fine-grained supervision for student-generated trajectories.
For effective optimization, this signal should align with overall trajectory correctness.
In practice, this alignment is not guaranteed and can fail in several typical ways:
\textbf{(a) OOD degradation}: when student rollouts enter sparse or out-of-distribution regions relative to the teacher, $\log \pi_{\mathrm{T}}(o_t \mid \cdot)$ may become noisy, disrupting the ranking between correct and incorrect trajectories.
\textbf{(b) Overestimation of incorrect trajectories}: incorrect trajectories may receive abnormally high scores when their local token patterns align with the teacher's high-confidence regions.
\textbf{(c) Underestimation of correct trajectories}: correct trajectories may receive abnormally low scores when their generation paths deviate from the teacher's dominant regions, thereby suppressing useful reasoning paths.
These phenomena suggest that teacher supervision is not always reliable, motivating us to introduce an outcome reward as a global anchor for calibrating trajectory-level supervision.

\subsection{The Overview of Uni-OPD}
\label{subsec: Overview}

In this work, we propose \texttt{Uni-OPD}, a unified OPD framework that generalizes across both LLMs and MLLMs, as illustrated in~\cref{fig:overview}.
Formally, given $M$ expert teachers $\{\pi_{\mathrm{T}_1}, \pi_{\mathrm{T}_2}, \dots, \pi_{\mathrm{T}_M}\}$ who specialize in different domains, and letting $w_i$ denote the weight assigned to teacher $\pi_{\mathrm{T}_i}$, we define the objective as:

\vspace{-1.4em}
\begin{equation}
    \label{eq: multi_teacher_objective}
    \mathcal{J}_{\text{Uni-OPD}}(\boldsymbol{\theta})
    =
    \sum_{i=1}^{M} w_i \, \mathcal{D}_{\mathrm{KL}}\!\left({\pi}_{\boldsymbol{\theta}} \,\|\, \pi_{\mathrm{T}_i}\right),
\end{equation}
\par\nobreak\vspace{-0.9em}

This formulation provides a unified objective for both single-teacher and multi-teacher distillation by aggregating supervision from multiple experts.
Building on this objective, we optimize OPD from the two fundamental roles.
From the student's perspective, we introduce a data-balancing strategy that promotes exploration via offline difficulty-aware and online correctness-aware selection.
From the teacher's perspective, we develop an outcome-guided margin calibration mechanism to correct unreliable token-level supervision by enforcing consistency with outcome rewards.
These designs stabilize optimization and improve the reliability of OPD.
\subsection{Joint Offline and Online Data Balancing Strategy for Student Exploration}

\label{subsec: Student_Perspective}

From the student's perspective, sufficient diversity and an appropriate level of difficulty in the generated trajectories are essential for effective OPD.
To this end, based on our empirical study, we propose complementary data-balancing strategies for both offline data construction and online sampling.

\begin{figure}[t!]
    \centering
    \includegraphics[width=\linewidth]{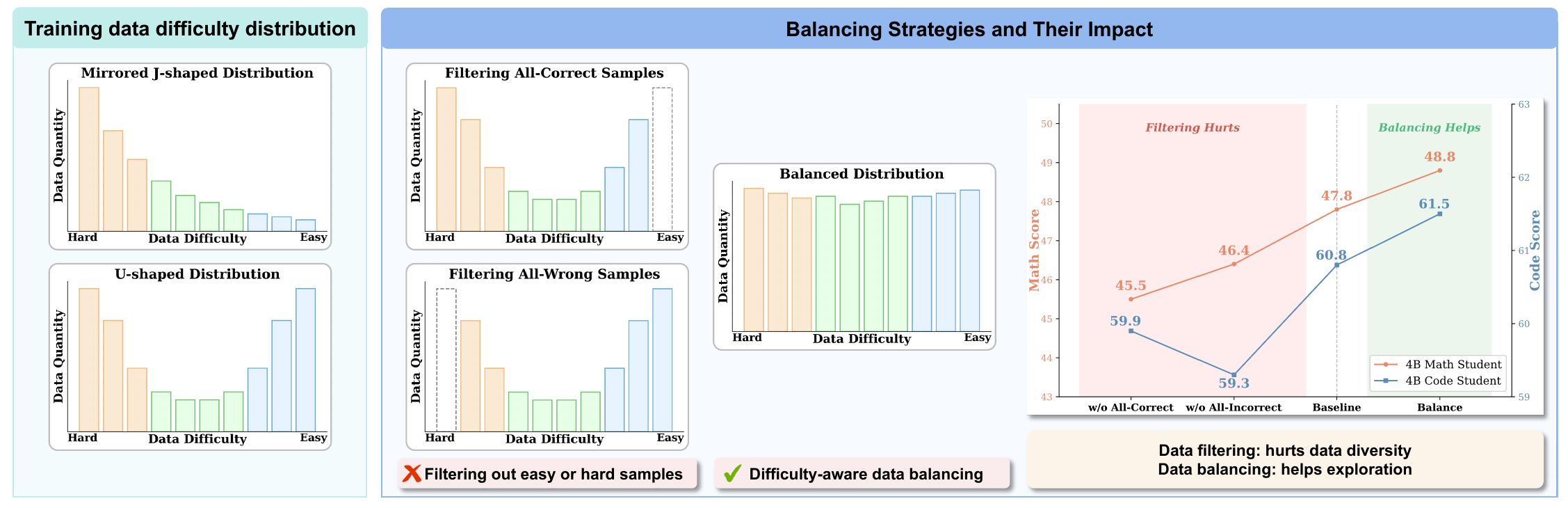}
    \vspace{-1.6em}
    \caption{
        \textbf{Data difficulty distribution and its impact on OPD performance.}
        (\textit{Left}) Training data often exhibits mirrored J-shaped or U-shaped difficulty distributions.
        (\textit{Right}) A naive strategy is to filter out overly easy or overly hard samples (\myie all-correct or all-wrong cases), but this reduces diversity.
        In contrast, our difficulty-balancing strategy upsamples mid-difficulty samples to preserve a balanced spectrum and empirically outperforms filtering.
    }
    \vspace{-1mm}
    \label{fig: offline_data_balancing}
\end{figure}

\custompara{Offline difficulty-aware data balancing.} 
A prevalent practice in RL is to estimate prompt difficulty via multiple rollouts and then filter out samples that are either overly easy (\myie always correct) or overly hard (\myie always incorrect)~\citep{Polaris2025,zhou2023lima}.
However, for small-scale models, training data often exhibits a mirrored J-shaped or U-shaped distribution (see~\cref{fig: offline_data_balancing}).
Strictly removing these easy or hard samples can substantially reduce data diversity and limit exploration of informative student-generated states.
Our empirical findings show that such filtering leads to substantial performance degradation in OPD.

Based on this observation, we adopt a difficulty-aware balancing strategy that selectively upsamples mid-difficulty samples (\myie correct in only some of multiple rollouts).
As shown in~\cref{fig: offline_data_balancing}, this strategy reshapes the data distribution into a more uniform form while preserving both diversity and difficulty.
In addition, it consistently improves performance on math reasoning and code generation.
Overall, these results show that maintaining data diversity and a balanced difficulty spectrum enables the student to generate more informative trajectories, thereby exploring a broader solution space.

\begin{wrapfigure}{r}{0.37\textwidth}
    \centering
    \vspace{-1.0em}
    \includegraphics[width=\linewidth]{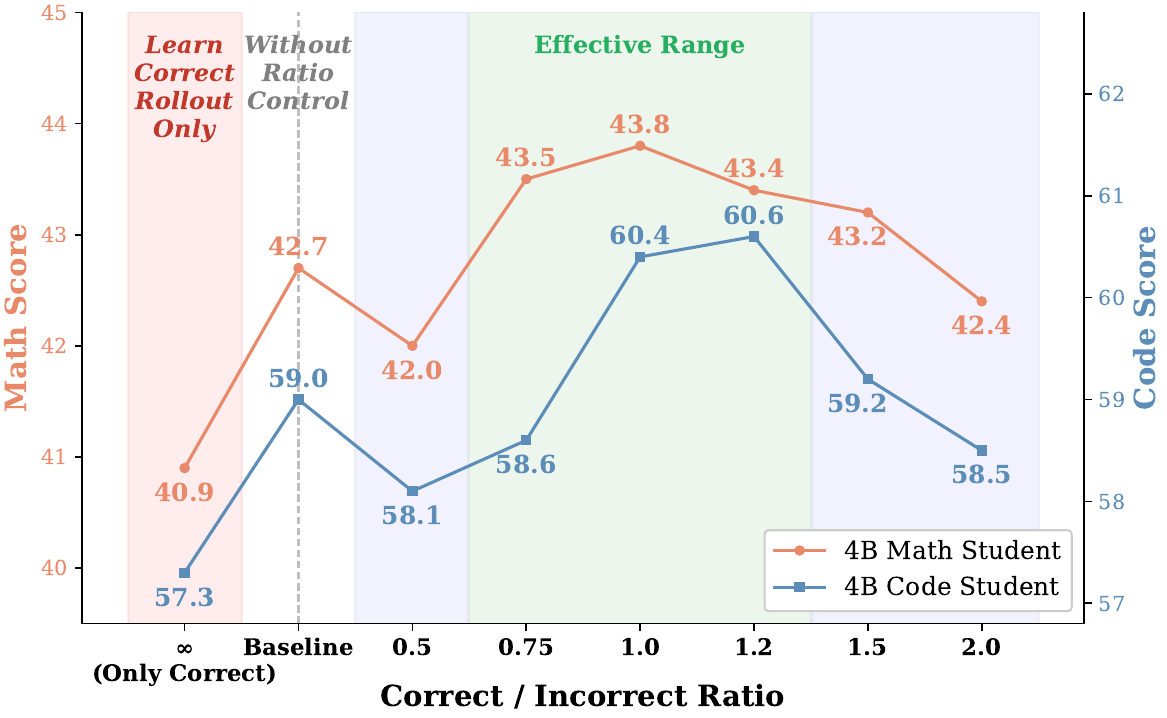}
    \vspace{-1.6em}
    \caption{
        Impact of online correct and incorrect ratio on student final performance.
    }
    \label{fig: online_data_balancing}
    \vspace{-1.5em}
\end{wrapfigure}

\vspace{1mm} 
\textbf{Online correctness-aware data balancing.}
\xspace
After applying offline difficulty-aware balancing, we further observe that insufficient exploration can cause the model to collapse to local optima during training, especially when rollout groups lack sufficient outcome diversity (\myeg only incorrect trajectories).
To mitigate this issue, we explicitly enforce a balanced composition of correct and incorrect trajectories within each rollout group during training.
This prevents degenerate cases in which all samples share the same outcome and thus yield uninformative gradients.
By maintaining such a balance, we ensure that the student consistently receives meaningful contrastive signals for stable on-policy learning.
As shown in~\cref{fig: online_data_balancing}, an appropriate outcome balance achieves better performance than using only correct samples or an excessively high correct/incorrect ratio.

\subsection{Outcome-guided Margin Calibration for Teacher Supervision}
\label{subsec:Teacher_Perspective}
A basic premise of OPD is that the teacher exhibits a directional likelihood preference over positive and negative trajectories. In particular, relative to the student, the teacher should assign higher likelihood to correct trajectories and lower likelihood to incorrect ones. Under this premise, the resulting distillation signal should remain consistent with outcome-level correctness at the trajectory level. We next formalize this principle through a trajectory-level distillation return and develop an outcome-guided calibration strategy based on it.

\begin{figure}[t!]
    \centering
    \includegraphics[width=\linewidth]{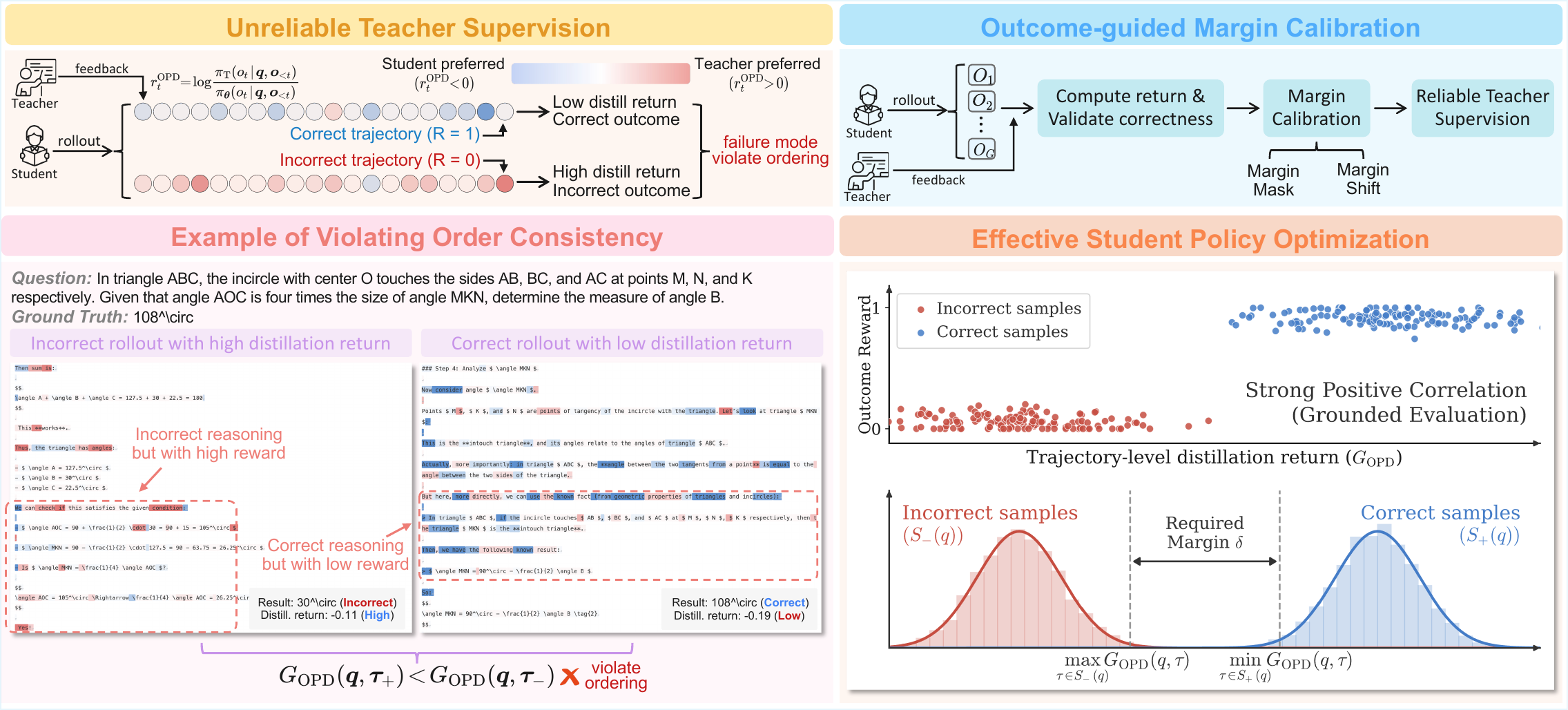}
    \vspace{-1.6em}
        \caption{
        \textbf{Demonstration of unreliable teacher supervision and outcome-guided margin calibration mechanism.}
        (\textit{Left}) Standard teacher supervision in OPD suffers from misalignment between trajectory-level return and outcome rewards, yielding unreliable supervision signals. 
        (\textit{Right}) Our method uses outcome rewards as a global anchor to calibrate returns through margin-based adjustment, restoring order consistency and improving optimization stability.
    }
    \vspace{-0.5em}
    \label{fig: teacher_supervision}
\end{figure}

\custompara{Trajectory-level distillation return.}
To characterize the overall supervision signal along a rollout trajectory, we define the \textit{trajectory-level distillation return} as the average log-probability gap between the teacher and the student:

\vspace{-1.0em}
\begin{equation}
    G_{\mathrm{OPD}}(\boldsymbol{q},\boldsymbol{\tau})
    \triangleq
    \frac{1}{|\boldsymbol{\tau}|}
    \sum_{t=1}^{|\boldsymbol{\tau}|}
    \log\frac{\pi_{\mathrm{T}}(o_{t} \mid \boldsymbol{q}, \boldsymbol{o}_{<t})}{\pi_{\boldsymbol{\theta}}(o_{t}\mid \boldsymbol{q}, \boldsymbol{o}_{<t})}
    =
    \frac{1}{|\boldsymbol{\tau}|}
    \sum_{t=1}^{|\boldsymbol{\tau}|} r^{\mathrm{OPD}}_{t}\, ,
    \label{eq: distillation_return}
\end{equation}
\par\nobreak\vspace{-0.5em}

This quantity measures the teacher's \emph{average log-likelihood preference} over the student along trajectory $\boldsymbol{\tau}$. When $G_{\mathrm{OPD}}(\vq, \boldsymbol{\tau}) > 0$, the teacher assigns higher confidence than the student on average, encouraging the student to move toward this trajectory. Conversely, when $G_{\mathrm{OPD}}(\vq, \boldsymbol{\tau}) < 0$, the student is discouraged from moving toward this trajectory. The normalization by trajectory length ensures comparability across trajectories of different lengths.

\custompara{Order consistency as a trajectory-level criterion.}
For a given question $\vq$, let $R(\vq, \boldsymbol{\tau}) \in \{0,1\}$ denote the outcome reward of a sampled trajectory $\boldsymbol{\tau}$, where $R(\vq, \boldsymbol{\tau}) = 1$ indicates that the final answer in $\boldsymbol{\tau}$ is correct for question $\vq$, and $R(\vq, \boldsymbol{\tau}) = 0$ otherwise. We then define the positive and negative trajectory sets as:
\begin{align}
    S_{+}(\vq) \triangleq \{\boldsymbol{\tau} \mid R(\vq, \boldsymbol{\tau}) = 1\},
    \qquad &
    S_{-}(\vq) \triangleq \{\boldsymbol{\tau} \mid R(\vq, \boldsymbol{\tau}) = 0\}.
    \label{eq: set_def}
    \\
    \intertext{
        Following the trajectory-level bandit formulation in~\citep{ouyang2022training}, we treat the prompt as the context and the entire generated trajectory as a macro-action. Under this view, the associated outcome reward naturally serves as a one-step trajectory-level return, denoted as $G_{\mathrm{RL}}(\vq, \boldsymbol{\tau}) = R(\vq, \boldsymbol{\tau})$. Therefore, the outcome-level RL return induces the following oracle ordering: 
    }
    G_{\mathrm{RL}}(\vq, \boldsymbol{\tau}_{+}) \ge G_{\mathrm{RL}}(\vq, \boldsymbol{\tau}_{-})\,,
    \qquad &
    \forall \boldsymbol{\tau}_{+} \in S_{+}(\vq),\; \forall \boldsymbol{\tau}_{-} \in S_{-}(\vq)\,.
    \label{eq: rl_order}
    \\
    \intertext{
        The derivation process is provided in~\cref{supp:subsec:order_consistency}. This motivates a trajectory-level reliability criterion for OPD. Under the distillation premise, the trajectory-level distillation return $G_{\mathrm{OPD}}(\vq, \boldsymbol{\tau})$ should preserve the same outcome-induced ordering as $G_{\mathrm{RL}}(\vq, \boldsymbol{\tau})$. Specifically, for any prompt $\vq$, we expect:
    }
    G_{\mathrm{OPD}}(\vq, \boldsymbol{\tau}_{+}) \ge G_{\mathrm{OPD}}(\vq, \boldsymbol{\tau}_{-})\,,
    \qquad &
    \forall \boldsymbol{\tau}_{+} \in S_{+}(\vq),\; \forall \boldsymbol{\tau}_{-} \in S_{-}(\vq)\,.
    \label{eq: opd_order}
\end{align}

\custompara{Teacher supervision may violate ordering.}
In practice, however, the teacher's supervision is not always reliable. As discussed in~\cref{subsec:preliminary}, teacher scoring may degrade in sparse out-of-distribution regions, overestimate incorrect trajectories, or underestimate correct ones due to spurious local patterns. Such failures may persist even after token-level supervision is aggregated to the trajectory level. A mean-based criterion is therefore insufficient, since the mismatch is often concentrated in a few extreme samples: a single overly confident negative trajectory or a severely underestimated positive trajectory can already distort the supervision signal for the entire prompt group.

\custompara{Outcome-guided margin calibration.}
Based on the above analysis, during OPD training, the constraint in~\cref{eq: opd_order} should hold between positive and negative trajectories within each prompt. To this end, we consider the margin between the lowest-scoring correct trajectory and the highest-scoring incorrect trajectory, which directly characterizes whether the ordering is violated in the most adversarial case. We define the prompt-level margin as

\vspace{-1.0em}
\begin{equation}
    m(\boldsymbol{q}) \triangleq
    \min_{\boldsymbol{\tau}\in S_{+}(\boldsymbol{q})}
    \!
    G_{\mathrm{OPD}}(\boldsymbol{q},\boldsymbol{\tau})
    -
    \max_{\boldsymbol{\tau}\in S_{-}(\boldsymbol{q})}
    \!
    G_{\mathrm{OPD}}(\boldsymbol{q},\boldsymbol{\tau})\, .
    \label{eq: prompt_level_margin}
\end{equation}
\par\nobreak\vspace{-0.5em}

By construction, $m(\vq) \ge 0$ indicates strict order consistency on prompt $\vq$, since even the worst positive trajectory still outperforms the best negative one (see Fig.~\ref{fig: teacher_supervision}). Thus, $m(\vq) \ge 0$ means that all positive trajectories are ranked above all negative ones for prompt $\vq$. To improve robustness, we further require:

\vspace{-1.0em}
\begin{equation}
    m(\boldsymbol{q}) \ge \delta\, ,
    \label{eq: ge_condition}
\end{equation}
\par\nobreak\vspace{-0.5em}

where $\delta > 0$ defines a safety margin against estimation noise and finite-sample fluctuations. Since $S_{+}(\vq)$ and $S_{-}(\vq)$ are determined by outcome rewards, this criterion uses the outcome signal as a global anchor to calibrate the teacher's trajectory-level scores. This formulation enables direct interventions on the margin, allowing us to suppress ordering violations or enlarge the separation between positive and negative trajectories.

\custompara{Margin calibration strategy.}
Based on~\cref{eq: ge_condition}, we present two calibration strategies: \textbf{margin mask} and \textbf{margin shift}.
Specifically, the margin mask keeps only the prompt groups satisfying $m(\vq) \ge \delta$ and discards the rest, so that training is performed only with reliable supervision.
Margin shift instead repairs an unreliable group with the smallest additive correction. For groups with $m(\vq) < \delta$, we define:
\begin{equation}
    \lambda(\vq) = \delta - m(\vq),
    \qquad
    \widetilde{G}_{\mathrm{OPD}}(\vq, \boldsymbol{\tau})
    =
    G_{\mathrm{OPD}}(\vq, \boldsymbol{\tau})
    +
    \lambda(\vq)\,\mathbf{1}\{R(\vq, \boldsymbol{\tau}) = 1\}.
    \label{eq: margin_shift_lift}
\end{equation}
This shift preserves the relative ordering within $S_{+}(\vq)$ and guarantees
\begin{equation}
    \min_{\boldsymbol{\tau} \in S_{+}(\vq)}
    \widetilde{G}_{\mathrm{OPD}}(\vq, \boldsymbol{\tau})
    -
    \max_{\boldsymbol{\tau} \in S_{-}(\vq)}
    \widetilde{G}_{\mathrm{OPD}}(\vq, \boldsymbol{\tau})
    =
    \delta\,.
\end{equation}
In this way, margin shift restores outcome-consistent ordering with a minimal group-level correction, while margin mask provides a more conservative alternative when the supervision signal is too unreliable to calibrate.

\begin{table*}[!t]
    \centering \small
    \caption{
        \textbf{Performance of \textit{Qwen3-4B Student} under math reasoning and code generation benchmarks.}
        Teacher models (\myie \textit{Qwen3-4B-Math-RL} and \textit{Qwen3-4B-Code-RL}) are developed through domain-specific RL. The performance of teacher models is denoted by the ``RL'' type. \textbf{Bold} values indicate the best score within each group. \textbf{Avg.} denotes the average score within each domain.
    }
    \vspace{-2mm}
    \label{tab: llm_full_4B_teacher_4B_student}
    \setlength{\tabcolsep}{7pt}
    \begin{tabular}{lccccccccc}
        \toprule
        \multirow{3}{*}{\vspace{-2mm}\textbf{Method}} &
        \multicolumn{5}{c}{\textbf{Math Reasoning}}   &
        \multicolumn{4}{c}{\textbf{Code Generation}}
        \\
        \cmidrule(lr){2-6}
        \cmidrule(lr){7-10}
        \small                                        &
        \makecell{AIME \nextline 2024}                &
        \makecell{AIME \nextline 2025}                &
        \makecell{HMMT \nextline 25 Feb.}             &
        \makecell{HMMT \nextline 25 Nov.}             &
        \textbf{Avg.}                                 &
        \makecell{Human \nextline Eval+}              &
        MBPP+                                         &
        LCB                                           &
        \textbf{Avg.}
        \\
        \midrule

        Student (4B)                                  & 23.0          & 19.3          & 12.3          & 9.2           & 15.9          & 77.4          & 65.3          & 17.7          & 53.5          \\
        Teacher (RL)                                  & 60.1          & 55.1          & 32.5          & 38.5          & 46.6          & 85.2          & 69.8          & 26.6          & 60.5          \\
        \midrule
        \multicolumn{10}{c}
        {\emph{\quad \textbf{Single--Teacher Distillation}}}                                                                                                                                          \\

        ExPO                                          & 58.7          & 55.2          & 32.4          & 37.0          & 45.8          & 84.8          & 70.2          & 28.0          & 61.0          \\
        OPD                                           & 57.9          & 52.4          & 30.2          & 37.8          & 44.6          & 82.6          & 68.8          & 25.7          & 59.0          \\
        ExOPD                                         & 62.7          & 56.1          & 33.9          & 39.3          & 48.0          & 86.9          & 70.7          & 28.6          & 62.1          \\
        \rowcolor{table-color} \texttt{Uni-OPD}       & \textbf{63.3} & \textbf{57.0} & \textbf{34.8} & \textbf{39.8} & \textbf{48.7} & \textbf{88.3} & \textbf{71.6} & \textbf{29.7} & \textbf{63.2} \\
        \midrule
        \multicolumn{10}{c}
        {\emph{\quad \textbf{Multi--Teacher Distillation}}}                                                                                                                                           \\
        SFT                                           & 58.5          & 53.3          & 30.7          & 34.8          & 44.3          & 86.4          & 69.6          & 26.4          & 60.8          \\
        ExPO                                          & 57.5          & 54.5          & 31.7          & 36.3          & 45.0          & 86.7          & 72.0          & 29.0          & 62.6          \\
        OPD                                           & 60.9          & 55.2          & 33.4          & 38.3          & 47.0          & 86.3          & 70.9          & 23.4          & 60.2          \\
        ExOPD                                         & 61.0          & 56.0          & 34.4          & 39.2          & 47.7          & 86.3          & 70.6          & 29.0          & 62.0          \\
        \rowcolor{table-color} \texttt{Uni-OPD}       & \textbf{62.3} & \textbf{57.2} & \textbf{34.9} & \textbf{39.6} & \textbf{48.5} & \textbf{88.0} & \textbf{72.6} & \textbf{30.1} & \textbf{63.6} \\
        \bottomrule
    \end{tabular}
\end{table*}

\section{Experiments and Analysis}
\label{sec: experiments}

In this section, we conduct comprehensive experiments across both textual and multimodal domains to evaluate the effectiveness of \texttt{Uni-OPD}. We first detail the experimental configurations (\cref{subsec: experimental_setup}). Subsequently, we assess how the proposed recipe improves OPD performance across diverse distillation scenarios for LLMs and MLLMs, including single-teacher and multi-teacher distillation (\cref{subsec: general_distillation}), strong-to-weak distillation (\cref{subsec: strong_to_weak_distillation}), and cross-modal distillation (\cref{subsec: cross_modal_distillation}). Finally, we provide a rigorous ablation study to further analyze the core strategies of our method (\cref{subsec: ablation_study}).

\vspace{-2mm}
\subsection{Experimental Setup}
\label{subsec: experimental_setup}

\begin{table*}[!t]
    \centering \small
    \caption{
        \textbf{Performance of \textit{Qwen3-VL-4B-Instruct Student} under math reasoning, logic reasoning, and document understanding benchmarks.}
        Teacher models (\myie \textit{Qwen3-VL-4B-Instruct-Math-RL}, \textit{Qwen3-VL-4B-Instruct-Logic-RL} and \textit{Qwen3-VL-4B-Instruct-Document-RL}) are developed through domain-specific RL. \textbf{Bold} values indicate the best score within each group. \textbf{Avg.} denotes the mean score within each category.
    }
    \vspace{-2mm}
    \label{tab: vlm_full_4B_Teacher_4B_Student}
    \setlength{\tabcolsep}{3pt}
    \begin{tabular}{lccccccccccccc}
        \toprule
        \multirow{3}{*}{\textbf{Method}}             &
        \multicolumn{4}{c}{\textbf{Math Reasoning}}  &
        \multicolumn{4}{c}{\textbf{Logic Reasoning}} &
        \multicolumn{5}{c}{\textbf{Document Understanding}}
        \\
        \cmidrule(lr){2-5}
        \cmidrule(lr){6-9}
        \cmidrule(lr){10-14}

                                                     & Math          & Dyna          & We            & \multirow{2}{*}{\textbf{Avg.}} & LogicVista    & LogicVista    & Visu          & \multirow{2}{*}{\textbf{Avg.}} & \multirow{2}{*}{AI2D} & Chart         & Doc           & Info          & \multirow{2}{*}{\textbf{Avg.}} \\
                                                     & Vision        & Math          & Math          &                                & Accuracy      & Format        & Logic         &                                &                       & QA            & VQA           & VQA           &                                \\
        \midrule

        Student (4B)                                 & 33.8          & 62.2          & 67.5          & 54.5                           & 49.9          & 66.4          & 25.1          & 47.0                           & 81.7                  & 73.5          & 94.9          & 79.8          & 82.5                           \\
        Teacher (RL)                                 & 47.2          & 65.3          & 79.5          & 64.0                           & 52.5          & 73.8          & 27.4          & 51.2                           & 82.5                  & 76.4          & 95.1          & 81.6          & 83.9                           \\
        \midrule
        \multicolumn{14}{c}{\emph{\quad \textbf{Single--Teacher Distillation}}}                                                                                                                                                                                                                                                 \\

        OPD                                          & 47.5          & 64.8          & 77.5          & 63.3                           & 49.8          & 73.0          & 26.1          & 49.6                           & 82.4                  & 75.4          & 95.2          & \textbf{81.4} & 83.6                           \\
        \rowcolor{table-color}
        \texttt{Uni-OPD}                             & \textbf{47.8} & \textbf{65.4} & \textbf{78.3} & \textbf{63.9}                  & \textbf{53.1} & \textbf{73.8} & \textbf{28.2} & \textbf{51.7}                  & \textbf{82.6}         & \textbf{75.8} & \textbf{95.2} & 81.2          & \textbf{83.7}                  \\
        \midrule
        \multicolumn{14}{c}{\emph{\quad \textbf{Multi--Teacher Distillation}}}                                                                                                                                                                                                                                                  \\
        OPD                                          & 41.0          & 60.9          & 71.7          & 57.9                           & 51.3          & 72.3          & 26.3          & 50.0                           & 82.6                  & 75.0          & 95.1          & 81.3          & 83.4                           \\
        \rowcolor{table-color}
        \texttt{Uni-OPD}                             & \textbf{45.5} & \textbf{62.3} & \textbf{76.1} & \textbf{61.0}                  & \textbf{54.0} & \textbf{75.2} & \textbf{27.5} & \textbf{52.5}                  & \textbf{83.0}         & \textbf{75.7} & \textbf{95.3} & \textbf{81.6} & \textbf{83.9}                  \\
        \bottomrule
    \end{tabular}
\end{table*}
\custompara{Models.}
We conduct experiments on the Qwen3 family~\citep{Qwen3_2025,Qwen3_VL_2025}. For textual experiments, we use Qwen3-4B/1.7B student models. In the same-sized setting, we apply domain-specific RL to Qwen3-4B to obtain specialized teachers. In the strong-to-weak setting, we use Qwen3-30B-A3B-Instruct-2507 as the strong teacher. For multimodal experiments, we use Qwen3-VL-2B-Instruct and Qwen3-VL-4B-Instruct as student models, and obtain multimodal teachers through domain-specific RL. Detailed training setups are in~\cref{supp:subsec:training_setup}.

\custompara{Training datasets.}
We use task-specific training data to construct and distill specialized teachers. For textual tasks, we use 57K math reasoning samples filtered from DeepMath \citep{he2025deepmath} (difficulty level $\geq 6$) and 25K code generation samples from the Code subset of Eurus-2-RL-Data \citep{cui2025process}. For multimodal tasks, we use math reasoning, logic reasoning, and document understanding data mainly from OpenMMReasoner-RL-74K \citep{zhang2025openmmreasoner}. Detailed training data configurations are provided in~\cref{supp:subsec:training_data}.

\custompara{Baselines.}
We compare \texttt{Uni-OPD} against several representative baselines for LLM distillation:
(1) \textbf{SFT}, which performs supervised fine-tuning on teacher-generated trajectories via cross-entropy loss;
(2) \textbf{ExPO}~\citep{G_OPD_2026}, a weight-space extrapolation method that merges domain-specific teachers and extrapolates their weights relative to the student model;
(3) \textbf{ExOPD}, a reward-level extrapolation approach that scales the reward factor ($>1$) to enable the student to surpass the performance boundaries of its teachers.
For MLLM experiments, since OPD remains largely underexplored in this setting, we use \textbf{vanilla OPD} as the primary baseline.

\custompara{Evaluation benchmarks.}
We evaluate \texttt{Uni-OPD} on a comprehensive benchmark suite spanning textual and multimodal capabilities, organized along five capability axes: \textbf{Textual Math Reasoning:} AIME24~\citep{aime2024}, AIME25~\citep{aime2025}, HMMT25 (February and November)~\citep{balunovic2025matharena}; \textbf{Textual Code Generation:} HumanEval+~\citep{liu2023your}, MBPP+~\citep{liu2023your}, and LiveCodeBench (v6 only, Feb. 25$\sim$May 25)~\citep{jain2024livecodebench}; \textbf{Multimodal Math Reasoning:} MathVision~\citep{wang2024measuring}, DynaMath~\citep{DynaMath_2025}, and WeMath~\citep{qiao2025we}; \textbf{Multimodal Logic Reasoning:} LogicVista~\citep{xiao2024logicvista} and VisuLogic~\citep{xu2025visulogic};
\textbf{Document Understanding:} AI2D~\citep{kembhavi2016diagram}, ChartQA~\citep{masry2022chartqa}, DocVQA~\citep{mathew2021docvqa}, and InfoVQA~\citep{mathew2022infographicvqa}.
Detailed information is in~\cref{supp:subsec:evaluation_benchmarks}.

\subsection{Single-Teacher and Multi-Teacher Distillation on LLMs and MLLMs}
\label{subsec: general_distillation}

As an effective and flexible paradigm for consolidating capabilities from one or multiple teachers into a unified student model, we first evaluate \texttt{Uni-OPD} on both LLMs and MLLMs across diverse domains.
Specifically, for LLMs, following G-OPD \citep{G_OPD_2026}, we conduct experiments on math reasoning and code generation.
For MLLMs, we further consider three domains: math reasoning, logic reasoning, and document understanding.

\custompara{Main results.} As shown in~\cref{tab: llm_full_4B_teacher_4B_student}, \texttt{Uni-OPD} achieves the best overall performance on LLM distillation under both single-teacher and multi-teacher settings.
In single-teacher distillation, \texttt{Uni-OPD} consistently outperforms OPD and ExOPD, obtaining the highest scores of 48.7 on math reasoning and 63.2 on code generation.
More importantly, under multi-teacher distillation, \texttt{Uni-OPD} effectively merges the distinct capabilities of multiple teachers into a single student model, yielding gains of 1.5\% and 3.4\% over OPD on math reasoning and code generation.

A similar trend is observed for MLLMs in~\cref{tab: vlm_full_4B_Teacher_4B_Student}. Under single-teacher distillation, \texttt{Uni-OPD} delivers the best average performance in all three domains, reaching 63.9 on math reasoning, 51.7 on logic reasoning, and 83.7 on document understanding.
For multi-teacher distillation, \texttt{Uni-OPD} consistently outperforms OPD, improving the average score from 57.9 to 61.0 on math reasoning, from 50.0 to 52.5 on logic reasoning, and from 83.4 to 83.9 on document understanding.
The consistent gains across settings validate the robustness of \texttt{Uni-OPD}.

\subsection{Strong-to-Weak Distillation}
\label{subsec: strong_to_weak_distillation}

\begin{table*}[!t]
    \centering \small
    \caption{
        \textbf{Results for strong-to-weak distillation setting under math reasoning and code generation benchmarks.}
        The teacher model is \textit{Qwen3-30B-A3B-Instruct-2507}, and the student models are the smaller \textit{Qwen3-4B} and \textit{Qwen3-1.7B}. \textbf{Bold} values indicate the best score within each group. \textbf{Avg.} denotes the average score within each domain.
    }
    \vspace{-2mm}
    \label{tab: llm_strong_to_weak_distillation}
    \setlength{\tabcolsep}{7pt}
    \begin{tabular}{lccccccccc}
        \toprule
        \multirow{3}{*}{\vspace{-2mm}\textbf{Method}} &
        \multicolumn{5}{c}{\textbf{Math Reasoning}}   &
        \multicolumn{4}{c}{\textbf{Code Generation}}
        \\
        \cmidrule(lr){2-6}
        \cmidrule(lr){7-10}
                                                      &
        \makecell{AIME \nextline 2024}                &
        \makecell{AIME \nextline 2025}                &
        \makecell{HMMT \nextline 25 Feb.}             &
        \makecell{HMMT \nextline 25 Nov.}             &
        \textbf{Avg.}                                 &
        \makecell{Human \nextline Eval+}              &
        MBPP+                                         &
        LCB                                           &
        \textbf{Avg.}                                                                                                                                                                                 \\
        \midrule

        Teacher                                       & 72.1          & 61.4          & 42.5          & 57.1          & 58.3          & 81.9          & 77.2          & 23.4          & 60.8          \\
        \midrule
        \multicolumn{10}{c}{\emph{\quad \textbf{Qwen3-4B Student}}}                                                                                                                                   \\

        Student                                       & 23.0          & 19.3          & 12.3          & 9.2           & 15.9          & 77.4          & 65.3          & 17.7          & 53.5          \\
        OPD                                           & \textbf{56.5} & 46.4          & 28.5          & 33.4          & 41.2          & 82.9          & \textbf{72.4} & 21.6          & 59.0          \\
        \rowcolor{table-color}
        \texttt{Uni-OPD}                              & 55.9          & \textbf{50.2} & \textbf{29.8} & \textbf{35.6} & \textbf{42.9} & \textbf{83.1} & 71.3          & \textbf{28.0} & \textbf{60.8} \\
        \midrule
        \multicolumn{10}{c}{\emph{\quad \textbf{Qwen3-1.7B Student}}}                                                                                                                                 \\

        Student                                       & 13.9          & 11.1          & 5.6           & 4.9           & 8.9           & 61.9          & 53.4          & 11.9          & 42.4          \\
        OPD                                           & \textbf{35.7} & 27.6          & 17.2          & 14.6          & 23.8          & 67.1          & 56.7          & 23.4          & 49.1          \\
        \rowcolor{table-color}
        \texttt{Uni-OPD}                              & 35.2          & \textbf{30.7} & \textbf{17.7} & \textbf{16.4} & \textbf{25.0} & \textbf{71.5} & \textbf{58.6} & \textbf{28.0} & \textbf{52.7} \\
        \bottomrule
    \end{tabular}
\end{table*}

Strong-to-weak distillation is particularly important for the practical post-training of small models \citep{Qwen3_VL_2025}. We further investigate whether \texttt{Uni-OPD} can better facilitate the transfer of reasoning capabilities from a larger, stronger teacher model (\myeg Qwen3-30B-A3B-Instruct-2507) to significantly smaller students (\myeg Qwen3-4B and Qwen3-1.7B). In this setting, the student is trained on both math and code data, with teacher feedback provided across both domains, which can be viewed as a multi-teacher scenario.

\custompara{Main results.}
The results for the strong-to-weak distillation setting are presented in~\cref{tab: llm_strong_to_weak_distillation}.
Notably, \texttt{Uni-OPD} yields significant performance gains across both the 4B and 1.7B student settings.
When distilled from the highly capable 30B teacher, \texttt{Uni-OPD} consistently outperforms standard OPD.
Specifically, for the 4B student, \texttt{Uni-OPD} achieves average scores of 42.9 in mathematical reasoning and 60.8 in code generation, surpassing standard OPD by 1.7 and 1.8 points, respectively.
This trend holds even for the highly constrained 1.7B student, where \texttt{Uni-OPD} lifts performance to 25.0 on math reasoning and 52.7 on code generation.
These results demonstrate that \texttt{Uni-OPD} effectively bridges the capacity gap, enabling smaller students to more effectively absorb and replicate complex reasoning behaviors from superior teachers.

\subsection{Cross-Modal Distillation}
\label{subsec: cross_modal_distillation}
\begin{table*}[htbp]
    \centering \small
    \caption{
        \textbf{Results for cross-modal distillation under textual code generation and multimodal math reasoning benchmarks.}
        The student model is \textit{Qwen3-VL-4B-Instruct}. The teacher models are developed from the same MLLM backbone via domain-specific RL on textual code and multimodal math domains, \myie \textit{Qwen3-VL-4B-Instruct-Code-RL} and \textit{Qwen3-VL-4B-Instruct-Math-RL}, respectively. \textbf{Bold} values indicate the best score within each group. \textbf{Avg.} denotes the average score within each domain.
    }
    \vspace{-2mm}
    \label{tab: vlm_llm_opd_code_math}
    \setlength{\tabcolsep}{10pt}
    \begin{tabular}{lccccccccc}
        \toprule
        \multirow{2.5}{*}{\vspace{-3mm}\textbf{Method}}        &
        \multicolumn{4}{c}{\textbf{Code Generation (Textual)}} &
        \multicolumn{4}{c}{\textbf{Math Reasoning (Multimodal)}}
        \\
        \cmidrule(lr){2-5}
        \cmidrule(lr){6-9}
        \small                                                 &
        \makecell{Human \nextline Eval+}                       &
        MBPP+                                                  &
        LCB                                                    &
        \textbf{Avg.}                                          &
        \makecell{Math \nextline Vision}                       &
        \makecell{Dyna \nextline Math}                         &
        \makecell{We \nextline Math}                           &
        \textbf{Avg.}
        \\
        \midrule
        \makecell[l]{Student}                                  & 76.8          & 70.0          & 37.0          & 61.3          & 33.8          & 62.2          & 67.5          & 54.5          \\
        \makecell[l]{Teacher}                                  & 82.2          & 70.5          & 40.1          & 64.3          & 47.2          & 65.3          & 79.5          & 64.0          \\
        \midrule
        OPD                                                    & 83.1          & 70.6          & 38.6          & 64.1          & 46.1          & 65.4          & 76.6          & 62.7          \\
        \rowcolor{table-color} \texttt{Uni-OPD}                & \textbf{84.1} & \textbf{71.4} & \textbf{41.3} & \textbf{65.6} & \textbf{46.6} & \textbf{66.5} & \textbf{78.5} & \textbf{63.9} \\
        \bottomrule
    \end{tabular}
\end{table*}
Cross-modal distillation is an important yet underexplored setting in OPD.
Unlike conventional distillation settings, where capability transfer typically occurs within the same modality, here \textbf{we investigate whether textual and multimodal capabilities can be unified into a single student policy}. Specifically, we use \textit{Qwen3-VL-4B-Instruct} as the student model, and construct domain-specific teachers from the same MLLM backbone via RL on textual code data and multimodal math data, respectively.
As a result, although the student is multimodal, one of the transferred capabilities is learned from a teacher specialized in a purely textual domain, enabling capability transfer across modality boundaries.
This setting is beneficial for integrating and transferring cross-modal capabilities.

\custompara{Main results.} As shown in~\cref{tab: vlm_llm_opd_code_math}, \texttt{Uni-OPD} achieves consistent gains over standard OPD across both textual code generation and multimodal math reasoning in this cross-modal setting. Specifically, it improves the average score from 64.1 to 65.6 on code generation and from 62.7 to 63.9 on math reasoning. On the textual side, the gains are consistent across all three code benchmarks, with the largest improvement on LCB (38.6 $\rightarrow$ 41.3). On the multimodal side, \texttt{Uni-OPD} further improves MathVision (46.1 $\rightarrow$ 46.6) and DynaMath (65.4 $\rightarrow$ 66.5), while maintaining strong performance on WeMath. These results suggest that \texttt{Uni-OPD} can effectively absorb and coordinate capabilities originating from both textual and multimodal domains within a unified student model, rather than improving one domain at the expense of the other. For a broader view of cross-modal distillation, we further provide results on code and logic reasoning in~\cref{supp:sec:further_evaluations}.

\subsection{Ablation Study}
\label{subsec: ablation_study}
\begin{table*}[htbp]
    \centering \small
    \caption{
        \textbf{Results of \texttt{Uni-OPD} variants with a \textit{Qwen3-4B Student} on math reasoning and code generation.}
        We ablate core strategies (\myie offline data balancing, online data balancing, and margin calibration) to assess their effectiveness using the Qwen3-4B-RL and Qwen3-30B-A3B-Instruct teacher models.
    }
    \vspace{-2mm}
    \label{tab: ablation}
    \setlength{\tabcolsep}{4pt}
    \begin{tabular}{lccccccccc}
        \toprule
        \multirow{3}{*}[-0.7mm]{\textbf{Configuration}} &
        \multicolumn{5}{c}{\textbf{Math Reasoning}}     &
        \multicolumn{4}{c}{\textbf{Code Generation}}
        \\
        \cmidrule(lr){2-6}
        \cmidrule(lr){7-10}
                                                        &
        \makecell{AIME \nextline 2024}                  &
        \makecell{AIME \nextline 2025}                  &
        \makecell{HMMT \nextline 25 Feb.}               &
        \makecell{HMMT \nextline 25 Nov.}               &
        \textbf{Avg.}                                   &
        \makecell{Human \nextline Eval+}                &
        MBPP+                                           &
        LCB                                             &
        \textbf{Avg.}
        \\
        \midrule
        \multicolumn{10}{c}{\emph{\quad \textbf{Qwen3-4B RL Teacher}}}                                                                   \\

        OPD                                             & 60.9 & 55.2 & 33.4 & 38.3 & 47.0          & 86.3 & 70.9 & 23.4 & 60.2          \\
        \rowcolor{table-color}
        \texttt{Uni-OPD}                                & 62.3 & 57.2 & 34.9 & 39.6 & \textbf{48.5} & 88.0 & 72.6 & 30.1 & \textbf{63.6} \\
        w/o offline data balancing                      & 62.6 & 56.5 & 32.5 & 38.5 & 47.5          & 88.0 & 71.1 & 27.9 & 62.3          \\
        w/o online data balancing                       & 62.5 & 56.7 & 33.2 & 38.9 & 47.8          & 88.0 & 71.8 & 28.0 & 62.6          \\
        w/o margin calibration                          & 63.0 & 54.7 & 33.4 & 38.1 & 47.3          & 86.4 & 71.6 & 25.7 & 61.2          \\
        \midrule
        \multicolumn{10}{c}{\emph{\quad \textbf{Qwen3-30B-A3B-Instruct-2507 Teacher}}}                                                   \\
        OPD                                             & 56.5 & 46.4 & 28.5 & 33.4 & 41.2          & 82.9 & 72.4 & 21.6 & 59.0          \\
        \rowcolor{table-color}
        \texttt{Uni-OPD}                                & 55.9 & 50.2 & 29.8 & 35.6 & \textbf{42.9} & 83.1 & 71.3 & 28.0 & \textbf{60.8} \\
        w/o offline data balancing                      & 57.1 & 46.3 & 28.8 & 36.8 & 42.2          & 80.6 & 70.3 & 28.0 & 59.6          \\
        w/o online data balancing                       & 57.0 & 47.6 & 26.8 & 37.0 & 42.1          & 81.6 & 71.4 & 28.0 & 60.3          \\
        w/o margin calibration                          & 54.9 & 48.1 & 29.1 & 35.8 & 42.0          & 82.8 & 70.4 & 25.7 & 59.6          \\
        \bottomrule
    \end{tabular}
\end{table*}

In~\cref{tab: ablation}, we conduct comprehensive ablation studies to evaluate the individual contributions of each strategy in our \texttt{Uni-OPD}.
Applying our proposed operations results in a significant improvement in accuracy over the vanilla OPD. In particular, the average gains reach +1.5/+3.4 points on math/code with the Qwen3-4B-RL teacher, and +1.7/+1.8 points with the Qwen3-30B-A3B-Instruct teacher.
\textbf{Offline and online data balancing address insufficient exploration}: without either of them, the student policy struggles to be exposed to diverse and challenging trajectories.
\textbf{Margin calibration improves supervision reliability}: without it, token-level feedback can become misaligned with outcome rewards, leading to less stable training and suboptimal performance.

\begin{wraptable}{r}{0.49\textwidth}
    \vspace{-4.5mm}
    \centering
    \small
    \caption{
        \textbf{Comparison results for different margin calibration.}
        We directly incorporate them into OPD to examine which strategy better benefits OPD training.
    }
    \vspace{-2mm}
    \label{tab: ablation_calibration}
    \setlength{\tabcolsep}{4pt}
    \begin{tabular}{lccccc}
        \toprule
        \textbf{Method}                   &
        \makecell{AIME \nextline 2024}    &
        \makecell{AIME \nextline 2025}    &
        \makecell{HMMT \nextline 25 Feb.} &
        \makecell{HMMT \nextline 25 Nov.} &
        \textbf{Avg.}
        \\
        \midrule
        Student (4B)                      & 23.0 & 19.3 & 12.3 & 9.2  & 15.9 \\
        OPD                               & 57.9 & 52.4 & 30.2 & 37.8 & 44.6 \\
        + margin mask                     & 62.3 & 56.2 & 34.3 & 38.1 & 47.7 \\
        \rowcolor{table-color}
        + margin shift                    & 62.7 & 56.3 & 34.4 & 39.2 & 48.1 \\
        \bottomrule
    \end{tabular}
\end{wraptable}

\vspace{1mm} 
\textbf{Margin mask vs. margin shift.}
\xspace We consider various strategies to calibrate the return signals for improving teacher supervision. In this work, we explore two simple variants, namely margin mask and margin shift. As shown in~\cref{tab: ablation_calibration}, directly incorporating either mechanism into OPD yields consistent performance gains over the baseline, underscoring the necessity of reliable teacher supervision. Among them, margin shift achieves slightly better results and is therefore adopted in our main experiments. More ablations are in~\cref{supp:subsec:further_ablation}.

\subsection{Qualitative Evaluation}
\begin{figure}[!t]
    \centering
    \includegraphics[width=1.0\linewidth]{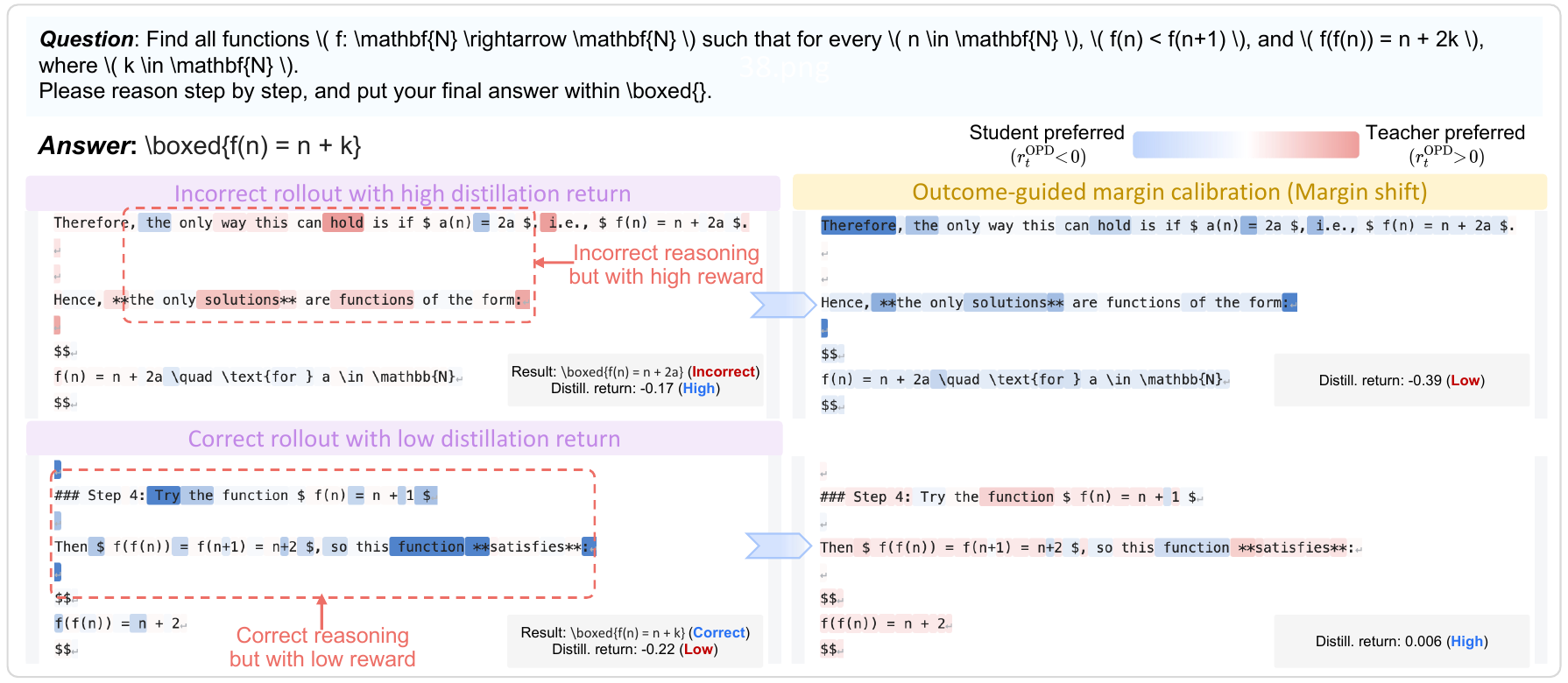}
    \vspace{-1.6em}
    \caption{
        \textbf{Heatmap visualization of failure modes in OPD and the effect of margin shift.}
        \textit{Left}: an incorrect rollout with a high distillation return (top) and a
        correct rollout with a low one (bottom).
        \textit{Right}: the same two rollouts after our margin shift, with the outcome ordering restored.
    }
    \label{fig: adv_heatmap_same_prompt}
    \vspace{-3mm}
\end{figure}

To intuitively illustrate the effectiveness of our outcome-guided margin calibration, we use a token-level reward heatmap for visualization.
As shown in~\cref{fig: adv_heatmap_same_prompt}, we display the two failure modes under the same question: the overestimation of incorrect trajectories (\emph{top-left}) and the underestimation of correct trajectories (\emph{bottom-left}).
Each token is colored by its reward value: blue tokens indicate student-preferred $(r^{\mathrm{OPD}}_{t}\!<\!0)$, and red tokens indicate teacher-preferred $(r^{\mathrm{OPD}}_{t}\!>\!0)$, with saturation proportional to magnitude.
On the \emph{top-left}, an \emph{incorrect} rollout still accumulates a high distillation return: most of its tokens are saturated red, since they fall on regions where the teacher dominates the student.
On the \emph{bottom-left}, a \emph{correct} rollout receives a low distillation return: its tokens are
already well-covered by the student, so the teacher provides little additional return (predominantly faint colors with some blue).
The \emph{right} column shows the same two rollouts after our outcome-guided margin calibration. Concretely, the per-token rewards are uniformly shifted so that the trajectory-level aggregation aligns with the outcome reward.

\subsection{Analysis and Takeaways}
Based on our comprehensive and systematic study on both LLMs and MLLMs across single-teacher, multi-teacher, strong-to-weak, and cross-modal distillation settings, we deliver three takeaways to further advance OPD.

\begin{itemize}[
        label=\raisebox{0.5ex}{\tiny$\bullet$},
        leftmargin=1em,
        itemsep=2pt, 
        parsep=0pt, 
        topsep=0pt, 
        partopsep=0pt 
    ]
    \item \textbf{Balancing reasoning capability and efficiency.}
          \texttt{Uni-OPD} achieves the best performance with substantially fewer optimization steps than RL (\cref{fig:teaser}), and consistently delivers strong reasoning capability across diverse domains (\crefrange{tab: llm_full_4B_teacher_4B_student}{tab: vlm_llm_opd_code_math}, and~\crefrange{tab: llm_full_4B_teacher_1.7B_student}{tab: vlm_llm_opd_code_logic} in the Appendix).

    \item \textbf{Teacher value comes from the capability gap, not absolute strength alone.}
          In OPD, even with the same 4B backbone, a domain-specific RL teacher injects new capabilities and knowledge that drive the student to improve and even surpass the teacher (\cref{tab: llm_full_4B_teacher_4B_student,tab: vlm_full_4B_Teacher_4B_Student}).
          Moreover, our dual-perspective recipe further translates this gap into student gains, consistently boosting performance across all model sizes.

    \item \textbf{OPD distills reasoning as a modality-agnostic capability.}
          Trained jointly on textual and multimodal data, the multimodal student under \texttt{Uni-OPD} improves textual code generation and multimodal math/logic reasoning (\cref{tab: vlm_llm_opd_code_math,tab: vlm_llm_opd_code_logic}).
          The per-token signal carries reasoning patterns largely independent of modality, enabling a unified, single-stage path that enhances both textual and multimodal reasoning within one multimodal model.

    \item \textbf{OPD cleanly merges specialized capabilities, with related ones reinforcing each other.}
          Beyond two teachers, \texttt{Uni-OPD} extends to three, jointly improving all capabilities (\cref{tab: vlm_full_4B_Teacher_4B_Student,tab: vlm_full_4B_Teacher_2B_Student}). OPD thus offers a scalable path for merging many specialists into one reasoner, with related ones synergizing via shared reasoning structure.
\end{itemize}

\custompara{Reproducibility statement.}
To facilitate a clear understanding of our contributions and support broader adoption of our work, we provide extensive materials. In the main text, we detail the key components of our method in~\cref{sec: methodology} and report the main experimental results in~\cref{sec: experiments}. In the supplementary materials, we further elaborate on Data Balancing Details (\cref{supp:sec:data_balancing_details}), Margin Calibration Details (\cref{supp:sec:margin_calibration_details}), Training Details (\cref{supp:sec:training_details}), and Evaluation Details (\cref{supp:sec:evaluation_details}), which together should be sufficient to reproduce our results.
All code, training data, complete scripts, and model checkpoints will be open-sourced upon publication to accelerate future research.

\section{Conclusion and Future Work}
\label{sec:conclusion}

In this paper, we present \texttt{Uni-OPD}, a unified OPD framework that generalizes across LLMs and MLLMs. We identify two key bottlenecks for effective OPD: insufficient student exploration of informative states and unreliable teacher supervision for student rollouts. To address them, we propose a dual-perspective optimization strategy: (i) offline difficulty-aware and online correctness-aware data balancing for student exploration, and (ii) outcome-guided margin calibration for teacher supervision. Extensive experiments on 16 benchmarks covering multi-teacher, strong-to-weak, and cross-modal settings demonstrate the effectiveness and versatility of \texttt{Uni-OPD}. We hope it can provide a practical foundation for future research on scalable and reliable distillation across models and modalities.

For future work, our findings suggest several promising directions:
(a) extending \texttt{Uni-OPD} to larger-scale teacher distillation settings;
(b) applying \texttt{Uni-OPD} to broader capability merging scenarios, such as agentic planning, tool use, and long-horizon decision making;
and (c) uncovering the mechanistic principles of OPD, particularly how it shapes training dynamics and parameter geometry.

\clearpage
\bibliographystyle{iclr2025_conference}
{
    \small\setlength{\bibsep}{8pt} 
    \bibliography{main}
}

\clearpage
\appendix
\counterwithin{figure}{section}
\counterwithin{table}{section}
\maketitlesupplementary
\setcounter{page}{1}

\section*{Appendix Outline}
This material provides supplementary details to the main paper, including the following sections:

\vspace{-0.5em}
\begin{itemize}[
        label=\raisebox{0.5ex}{\tiny$\bullet$},
        leftmargin=1em,
        itemsep=0pt, 
        parsep=2pt, 
        partopsep=0pt 
    ]
    \item (\ref{supp:sec:data_balancing_details}) \textbf{Data Balancing Details}
          \begin{itemize}[label=-]
              \item (\ref{supp:subsec:offline_data_balancing}) Offline Difficulty-Aware Data Balancing
              \item (\ref{supp:subsec:online_data_balancing}) Online Correctness-Aware Data Balancing
          \end{itemize}
    \item (\ref{supp:sec:margin_calibration_details}) \textbf{Margin Calibration Details}
          \begin{itemize}[label=-]
              \item (\ref{supp:subsec:order_consistency}) Order Consistency of Trajectory-level Returns
              \item (\ref{supp:subsec:consistency_violation}) Order-Consistency Violations and Their Harm
              \item (\ref{supp:subsec:violation_statistics}) Statistical Evidence for Order-Consistency Violations
              \item (\ref{supp:subsec:margin_calibration}) Outcome-Guided Margin Calibration
          \end{itemize}
    \item (\ref{supp:sec:training_details}) \textbf{Training Details}
          \begin{itemize}[label=-]
              \item (\ref{supp:subsec:training_setup}) Training Setup
              \item (\ref{supp:subsec:training_data}) Training Data
              \item (\ref{supp:subsec:reward_acquisition}) Training Reward Acquisition
              \item (\ref{supp:subsec:training_pseudocode}) Training Pseudocode
              \item (\ref{supp:subsec:training_dynamics}) Training Dynamics
              \item (\ref{supp:subsec:training_complexity}) Training Complexity
          \end{itemize}
    \item (\ref{supp:sec:evaluation_details}) \textbf{Evaluation Details}
          \begin{itemize}[label=-]
              \item (\ref{supp:subsec:evaluation_benchmarks}) Evaluation Benchmarks
              \item (\ref{supp:subsec:evaluation_setup}) Evaluation Setup
          \end{itemize}
    \item (\ref{supp:sec:further_evaluations}) \textbf{Further Evaluations}
          \begin{itemize}[label=-]
              \item (\ref{supp:subsec:more_evaluation_results}) More Evaluation Results
              \item (\ref{supp:subsec:downstream_task_evaluation}) Downstream Task Evaluation
              \item (\ref{supp:subsec:further_ablation}) Further Ablation
          \end{itemize}
    \item (\ref{supp:sec:related_work}) \textbf{Related Work}
          \begin{itemize}[label=-]
              \item (\ref{supp:subsec:multimodal_large_language_models}) Multimodal Large Language Models
              \item (\ref{supp:subsec:reinforcement_learning}) Reinforcement Learning
              \item (\ref{supp:subsec:on_policy_distillation}) On-Policy Distillation
          \end{itemize}
    \item (\ref{supp:sec:case_studies}) \textbf{Case Studies}
\end{itemize}

\clearpage
\section{Data Balancing Details}
\label{supp:sec:data_balancing_details}

In this section, we provide a detailed exposition of data balancing strategies in our proposed \texttt{Uni-OPD} framework, including formulations and implementations. These strategies operate on complementary levels: offline difficulty-aware balancing (\cref{supp:subsec:offline_data_balancing}) shapes the \emph{prompt-level} difficulty distribution before training, while online correctness-aware balancing (\cref{supp:subsec:online_data_balancing}) regulates the \emph{intra-group} composition of correct and incorrect trajectories during rollout.

\subsection{Offline Difficulty-Aware Data Balancing}
\label{supp:subsec:offline_data_balancing}

In this section, we provide a detailed description of our offline difficulty-aware data balancing strategy.

\custompara{Offline rollout sampling.}
Before training, we perform a one-time offline rollout pass over the entire training dataset using the student model (\myeg Qwen3-4B). For each training prompt, the student is prompted to generate $N\!=\!8$ independent candidate responses, which serve as the basis for subsequent difficulty estimation.

The rollouts are produced with vLLM~\citep{vLLM_2023} under the same prompt template that will later be used at training time, so that the estimated difficulty reflects the actual input format the student will see. 
The decoding configuration is kept fixed throughout this offline phase: we use temperature $=1.0$, top-$p=0.95$, top-$k=50$, and a maximum response length of $16{,}384$ tokens. 
For each instance, we then verify the correctness of its $N$ candidate responses with the task-specific verifier (described in~\cref{supp:subsec:reward_acquisition}) and record the number of correct ones. 
The resulting empirical pass rate $k/N$ serves as our proxy for the instance's difficulty: a lower pass rate indicates a harder example, while a higher pass rate indicates an easier one.

\custompara{Limitations of aggressive difficulty filtering.}
Prior work on online RL optimization, such as GRPO, often relies on a heuristic pre-training filter that simply discards ``trivial'' samples such as all-correct cases, because these instances yield zero advantage and therefore provide essentially no learning signal. 
POLARIS~\citep{Polaris2025}, for example, reports that removing the easiest samples leads to consistent performance gains, and argues that keeping an unfiltered dataset can actively hinder training.

In the token-level reward OPD setting, however, we find that such aggressive filtering is counterproductive. 
Empirically, removing any specific difficulty tier, whether the easiest or the hardest, consistently hurts final performance. 
A plausible explanation is that each tier contributes a distinct pattern of token-level credit: easy instances calibrate the student's baseline behavior, intermediate instances provide the richest contrastive signals between correct and incorrect trajectories, and hard instances expose the student to diverse, non-trivial solution paths. 
Dropping any tier, therefore, both distorts the overall distribution of token-level credit and narrows the space of solution patterns to which the student is exposed.

\custompara{Difficulty-aware data balancing.}
Motivated by this observation, we adopt a difficulty-aware balancing scheme that deliberately preserves the full spectrum of difficulty while reweighting its different regions, rather than truncating them. Concretely, after the offline rollout pass, we examine the empirical distribution over the number of correct responses out of $N$. 
Across our training sources, we observe two recurring shapes: (i) a \textbf{U-shaped} distribution, where both very easy and very hard instances dominate while intermediate ones are sparse; and (ii) a \textbf{mirrored-J-shaped} distribution, where easy instances dominate and the mass decays toward the hard end.

We treat the two shapes slightly differently. For U-shaped distributions, we upsample instances of intermediate difficulty, namely those with $1$--$7$ correct responses out of $N=8$, so as to fill in the under-represented middle region. For mirrored-J-shaped distributions, we instead upsample all non-trivial instances, \myie everything with $1$--$8$ correct responses, to counteract the long tail of easy samples. In both cases, the effect of the reweighting is to flatten the overall difficulty distribution and to ensure that the token-level credit signals arriving during training are more evenly spread across difficulty levels. Empirically, we find that this simple rebalancing consistently leads to better final performance than either no filtering or the conventional drop-the-easy-cases strategy.

\subsection{Online Correctness-Aware Data Balancing}
\label{supp:subsec:online_data_balancing}

In this section, we detail the online correctness-aware data balancing strategy that operates during rollout.
While the offline difficulty-aware balancing in~\cref{supp:subsec:offline_data_balancing} controls the \emph{prompt-level} difficulty distribution before training, the composition of correct and incorrect trajectories \emph{within a rollout group} still varies dramatically as the student evolves.
This subsection describes how we regulate such intra-group composition online.

\custompara{Motivation.}
In OPD, for each prompt $\boldsymbol{q}$ we sample $N$ on-policy trajectories $\{\boldsymbol{\tau}_i\}_{i=1}^{N}$ and split them into a positive set $S_{+}(\boldsymbol{q})$ and a negative set $S_{-}(\boldsymbol{q})$ based on the outcome reward $R_{i}$.
As training proceeds, many prompts exhibit degenerate outcome distributions: either $|S_{-}(\boldsymbol{q})|\!\ll\!N$ (the student nearly masters $\boldsymbol{q}$) or $|S_{+}(\boldsymbol{q})|\!\ll\!N$ (the student often fails on $\boldsymbol{q}$).
In both cases, the outcome-level contrast vanishes and the outcome-guided margin calibration in~\cref{supp:subsec:margin_calibration} cannot provide any corrective signal, since the prompt-level margin $m(\boldsymbol{q})$ is undefined.
If left unregulated, such degenerate groups dominate the batch and drive the student into local optima with shrinking exploration.

\custompara{Online correctness-aware balancing.}
To preserve sufficient outcome diversity throughout training, we maintain a target correct-to-total ratio $\gamma^\star\!\in\!(0,1)$ at the batch level (we use $\gamma^\star\!\approx\!0.5$ by default, so positive and negative trajectories are roughly balanced).
At each training step, given a freshly rolled-out batch $\mathcal{B}$, we let $\gamma(\mathcal{B})=\sum_{\boldsymbol{\tau}_i\!\in\!\mathcal{B}}\mathbf{1}\{R_i\!=\!1\}/|\mathcal{B}|$ denote the current correct-to-total ratio across the whole batch.
Whenever $|\gamma(\mathcal{B})-\gamma^\star|\!>\!\epsilon$ for a tolerance $\epsilon$, we downweight the over-represented side (correct or incorrect trajectories) by subsampling within each group, so that the overall batch ratio is pulled back to the $\gamma^\star\!\pm\!\epsilon$ interval.
Subsampling is performed uniformly inside each group, which keeps the intra-group difficulty distribution intact and avoids biasing the prompt-level difficulty spectrum inherited from offline balancing.
\section{Margin Calibration Details}
\label{supp:sec:margin_calibration_details}

In this section, we provide a detailed exposition of the outcome-guided margin calibration in \texttt{Uni-OPD}. We first recall the order-consistency criterion for trajectory-level returns (\cref{supp:subsec:order_consistency}), then formalize what it means for teacher supervision to violate this criterion and analyze the resulting harm (\cref{supp:subsec:consistency_violation}). We next present a statistical study showing that such violations are prevalent in practice (\cref{supp:subsec:violation_statistics}), which motivates the two calibration strategies described in~\cref{supp:subsec:margin_calibration}.

\subsection{Order Consistency of Trajectory-level Returns}
\label{supp:subsec:order_consistency}

This section provides a brief explanation for the order-consistency conditions in~\cref{eq: rl_order,eq: opd_order} of the main paper. 
The key observation is two-fold. First, treating the entire reasoning rollout as a single macro-action gives $G_{\mathrm{RL}}(\vq,\boldsymbol{\tau}) \!=\! R(\vq,\boldsymbol{\tau})$, so $G_{\mathrm{RL}}$ respects the outcome-induced ordering by construction. Second, under the distillation premise, the trajectory-level distillation return $G_{\mathrm{OPD}}(\vq,\boldsymbol{\tau})$ is expected to preserve the same ordering, although this is a desideratum rather than a definitional consequence.

\custompara{Trajectory-as-one-action view of outcome-based RL.}
In outcome-based RL for reasoning, supervision is provided only at the trajectory level: a rollout $\boldsymbol{\tau}$ receives a single scalar reward $R(\boldsymbol{q},\boldsymbol{\tau})$ determined by the final answer. 
Under this view, the trajectory-level return reduces to the outcome reward itself, \myie

\vspace{-1.3em}
\begin{equation}
    G_{\mathrm{RL}}(\boldsymbol{q},\boldsymbol{\tau}) = R(\boldsymbol{q},\boldsymbol{\tau})\,.
    \label{eq: supp_rl_return}
\end{equation}
\par\nobreak\vspace{-0.8em}

\custompara{Order consistency under binary rewards.}
For the binary outcome reward adopted in this work, any $\boldsymbol{\tau}_{+}\!\in\!S_{+}(\boldsymbol{q})$ satisfies $R(\boldsymbol{q},\boldsymbol{\tau}_{+})\!=\!1$, while any $\boldsymbol{\tau}_{-}\!\in\!S_{-}(\boldsymbol{q})$ satisfies $R(\boldsymbol{q},\boldsymbol{\tau}_{-})\!=\!0$. 
Combined with~\cref{eq: supp_rl_return}, we have

\vspace{-1.3em}
\begin{equation}
    G_{\mathrm{RL}}(\boldsymbol{q},\boldsymbol{\tau}_{+})=1 \;\ge\; 0=G_{\mathrm{RL}}(\boldsymbol{q},\boldsymbol{\tau}_{-})\,,
\end{equation}
\par\nobreak\vspace{-0.5em}

for all $\boldsymbol{\tau}_{+}\!\in\!S_{+}(\boldsymbol{q})$ and $\boldsymbol{\tau}_{-}\!\in\!S_{-}(\boldsymbol{q})$, which recovers~\cref{eq: rl_order} directly.

\custompara{Extension to soft outcome rewards.}
The same argument extends to soft outcome rewards, where $R(\boldsymbol{q},\boldsymbol{\tau})\!\in\![0,1]$ (or any bounded interval) measures a graded notion of correctness, \myeg partial credit or a verifier's confidence score. 
As long as the trajectory partition is defined by thresholding the outcome reward, \myie $S_{+}(\boldsymbol{q})\!=\!\{\boldsymbol{\tau}\mid R(\boldsymbol{q},\boldsymbol{\tau})\!\ge\!\eta\}$ and $S_{-}(\boldsymbol{q})\!=\!\{\boldsymbol{\tau}\mid R(\boldsymbol{q},\boldsymbol{\tau})\!<\!\eta\}$ for some threshold $\eta$, then by~\cref{eq: supp_rl_return} every positive trajectory attains a return no smaller than that of any negative trajectory, and~\cref{eq: rl_order} still holds. 
In particular, the binary case is recovered as the special instance $\eta\!=\!1$, $R\!\in\!\{0,1\}$.

\custompara{From RL return to distillation return.}
The distillation return $G_{\mathrm{OPD}}(\boldsymbol{q},\boldsymbol{\tau})$ defined in~\cref{eq: distillation_return} plays the same role for OPD training as $G_{\mathrm{RL}}$ does for outcome-based RL: it is the trajectory-level supervision signal broadcast to all tokens in the rollout. 
The distillation premise in~\cref{subsec:Teacher_Perspective} posits that, relative to the student, the teacher assigns a higher log-likelihood to correct trajectories than incorrect ones. 
In other words, the teacher's trajectory-level preference is expected to be aligned with the outcome reward, so that $G_{\mathrm{OPD}}$ should inherit the same outcome-level ordering as $G_{\mathrm{RL}}$, leading to~\cref{eq: opd_order}. 
Unlike the RL return, however, $G_{\mathrm{OPD}}$ is derived from the teacher--student log-probability gap rather than the outcome reward itself, so the ordering is a desired property rather than a guaranteed one. 
The order-consistency condition in~\cref{eq: opd_order} provides a principled target, and subsequent margin mask and margin shift strategies (\cref{supp:subsec:margin_calibration}) are designed to enforce it whenever the teacher's supervision violates this property in practice.
\subsection{Order-Consistency Violations and Their Harm}
\label{supp:subsec:consistency_violation}

The order-consistency criterion in~\cref{eq: opd_order} states a desideratum: within each prompt, the trajectory-level distillation return should rank every correct trajectory above every incorrect one. 
In this section, we first formalize what it means for teacher supervision to violate this criterion.
We then explain why such a violation harms training by analyzing how the trajectory-level return steers the OPD update.
Throughout, we consider a non-degenerate prompt $\vq$, \myie one for which both $S_{+}(\vq)$ and $S_{-}(\vq)$ are non-empty.

\custompara{Order-consistency violation.}
We say that the teacher's trajectory-level supervision \textit{violates order consistency} on prompt $\vq$ if the ordering condition in~\cref{eq: opd_order} fails, \myie there exists at least one inverted positive--negative pair:
\begin{equation}
    \exists\,\boldsymbol{\tau}_{+}\!\in\!S_{+}(\vq),\;
    \boldsymbol{\tau}_{-}\!\in\!S_{-}(\vq)
    \quad\text{s.t.}\quad
    G_{\mathrm{OPD}}(\vq,\boldsymbol{\tau}_{+})
    <
    G_{\mathrm{OPD}}(\vq,\boldsymbol{\tau}_{-})\,.
    \label{eq: violation_def}
\end{equation}
By the definition of the prompt-level margin in~\cref{eq: prompt_level_margin}, the worst inverted pair is exactly the one attaining $\min_{\boldsymbol{\tau}\in S_{+}}G_{\mathrm{OPD}}$ and $\max_{\boldsymbol{\tau}\in S_{-}}G_{\mathrm{OPD}}$. 
Hence, a violation occurs on $\vq$ if and only if
\begin{equation}
    m(\vq)
    =
    \min_{\boldsymbol{\tau}\in S_{+}(\vq)}\!G_{\mathrm{OPD}}(\vq,\boldsymbol{\tau})
    -
    \max_{\boldsymbol{\tau}\in S_{-}(\vq)}\!G_{\mathrm{OPD}}(\vq,\boldsymbol{\tau})
    <
    0\,,
    \label{eq: violation_margin}
\end{equation}

\noindent
which is the operational condition we use throughout. The margin $m(\vq)$ therefore serves as a certificate: $m(\vq)\!\ge\!0$ guarantees that no inverted pair exists, whereas $m(\vq)\!<\!0$ signals that at least one violation exists.

\custompara{Consequence: violations reinforce incorrect trajectories.}
Recall from~\cref{subsec:Teacher_Perspective} that $G_{\mathrm{OPD}}(\vq,\boldsymbol{\tau})$ is the teacher's average log-likelihood preference over the student along $\boldsymbol{\tau}$: a larger value encourages the student to move toward $\boldsymbol{\tau}$, while a smaller one discourages it. 
When order consistency holds, every correct trajectory carries a larger $G_{\mathrm{OPD}}$ than every incorrect one.
The trajectories the student is encouraged to move toward are therefore exactly the outcome-correct ones, in agreement with the outcome-induced ordering in~\cref{eq: rl_order}.
A violation reverses this relationship. 
When $m(\vq)\!<\!0$, there exists a correct trajectory $\boldsymbol{\tau}_{+}$ whose return is even smaller than that of some incorrect trajectory $\boldsymbol{\tau}_{-}$, so the update encourages the student toward the incorrect $\boldsymbol{\tau}_{-}$ more strongly than toward the correct $\boldsymbol{\tau}_{+}$. 
In other words, a violation indicates that the teacher supervision is unreliable in the sense of~\cref{subsec:preliminary}.
Following such supervision biases the update toward incorrect trajectories and hinders the student's optimization.
\subsection{Statistical Evidence for Order-Consistency Violations}
\label{supp:subsec:violation_statistics}

Having formalized order-consistency violations in~\cref{supp:subsec:consistency_violation}, we now provide a statistical study demonstrating that they are prevalent rather than incidental. 
Unless otherwise stated, we use Qwen3-4B as the student and Qwen3-30B-A3B-Instruct-2507 as the teacher, and sample on-policy trajectories on the textual math training data.

\begin{figure}[t!]
    \centering
    \includegraphics[width=\linewidth]{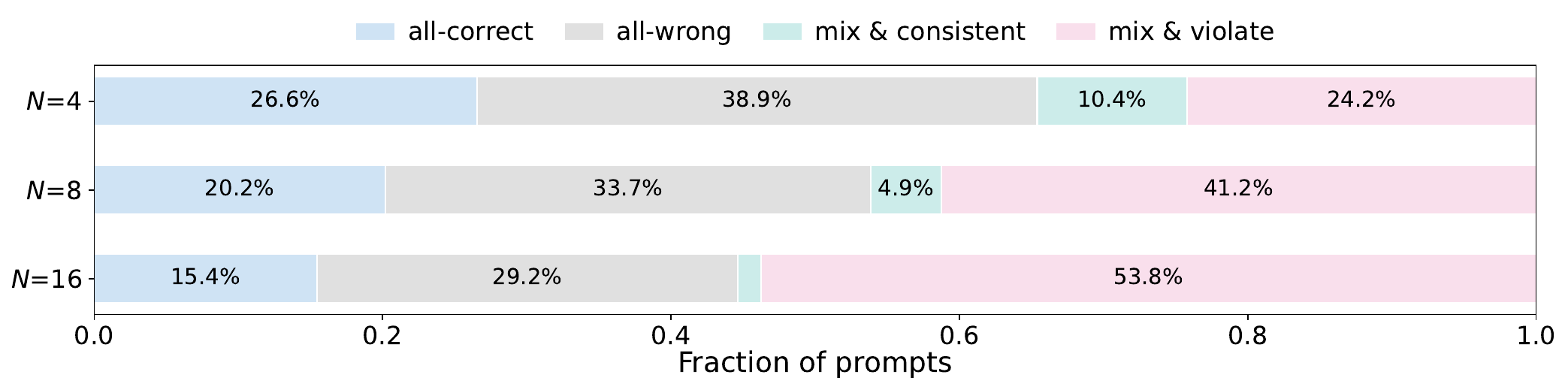}
    \vspace{-1.6em}
    \caption{
        \textbf{Prevalence of order-consistency violations across rollout sizes.}
        For each prompt, we sample $N$ on-policy trajectories from the student on the math training data and classify them into one of four groups.
        \textit{All-correct} and \textit{all-wrong} are degenerate prompts with no comparable positive--negative pair;
        \textit{mix \& consistent} denotes non-degenerate prompts satisfying $m(\vq)\!\ge\!0$;
        \textit{mix \& violate} denotes non-degenerate prompts with $m(\vq)\!<\!0$.
        The violating fraction already reaches $24.2\%$ at $N\!=\!4$ and grows rapidly with $N$, exceeding the consistent fraction.
    }
    \vspace{-1mm}
    \label{fig: prompt_fraction_stacked}
\end{figure}

\custompara{Violations are prevalent and grow with rollout size.}
We first count how many prompts fail the condition $m(\vq)\!\ge\!0$ at the prompt level. 
For each prompt we draw $N$ on-policy trajectories. Following~\cref{supp:subsec:consistency_violation}, we then partition the prompts into four groups: two degenerate ones, \textit{all-correct} and \textit{all-wrong}, for which no comparable positive--negative pair exists and order consistency is vacuous; \textit{mix \& consistent}, the non-degenerate prompts for which the lowest-scoring correct trajectory still outranks the highest-scoring incorrect one (\myie $m(\vq)\!\ge\!0$); and \textit{mix \& violate}, the non-degenerate prompts with $m(\vq)\!<\!0$. 
As shown in~\cref{fig: prompt_fraction_stacked}, at $N\!=\!4$, $24.2\%$ of prompts violate the condition, substantially more than the $10.4\%$ that remain consistent. 
The violating fraction increases sharply as $N$ grows, since more sampled trajectories provide more chances to expose an inverted pair, reaching roughly half of all prompts at $N\!=\!16$. 
This confirms that failing to guarantee $m(\vq)\!\ge\!0$ is a widespread phenomenon across prompts.

\begin{figure}[t!]
    \centering
    \begin{subfigure}[t]{0.50\linewidth}
        \centering
        \includegraphics[width=\linewidth]{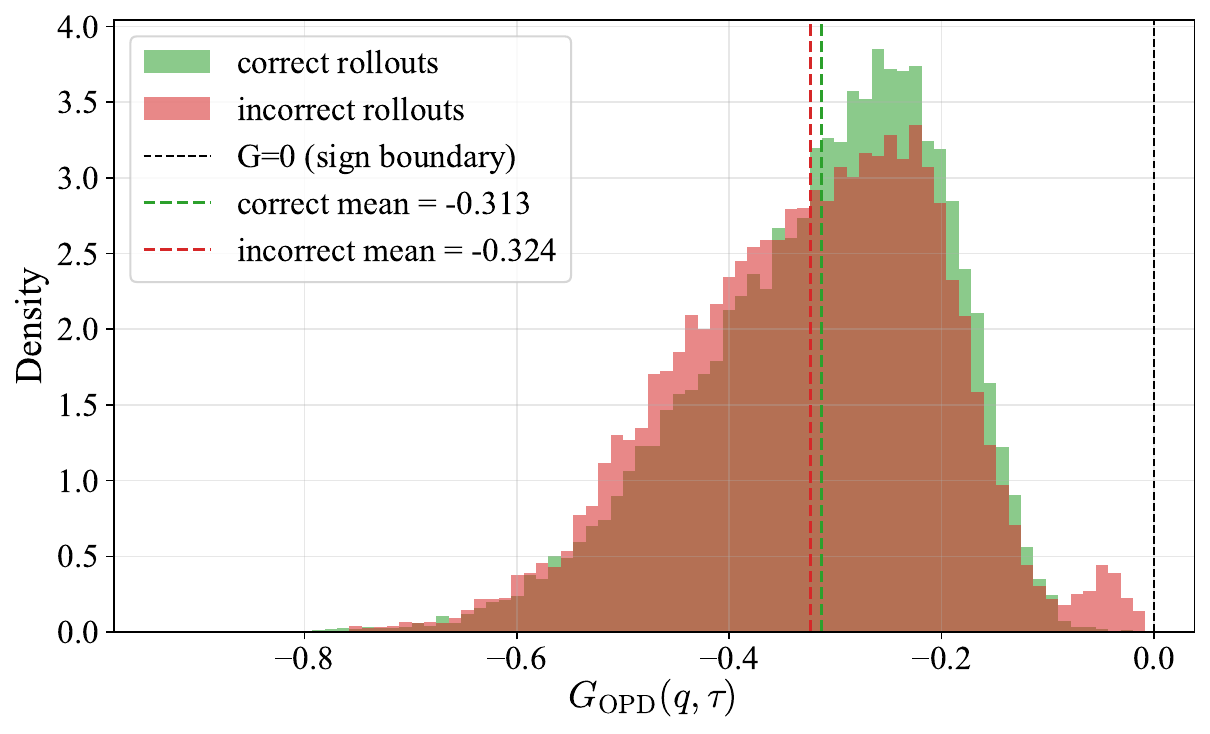}
        \vspace{-1.6em}
        \caption{Distribution of $G_{\mathrm{OPD}}$ for correct vs. incorrect rollouts.}
        \label{fig: pos_neg_overlap}
    \end{subfigure}
    \hfill
    \begin{subfigure}[t]{0.475\linewidth}
        \centering
        \includegraphics[width=\linewidth]{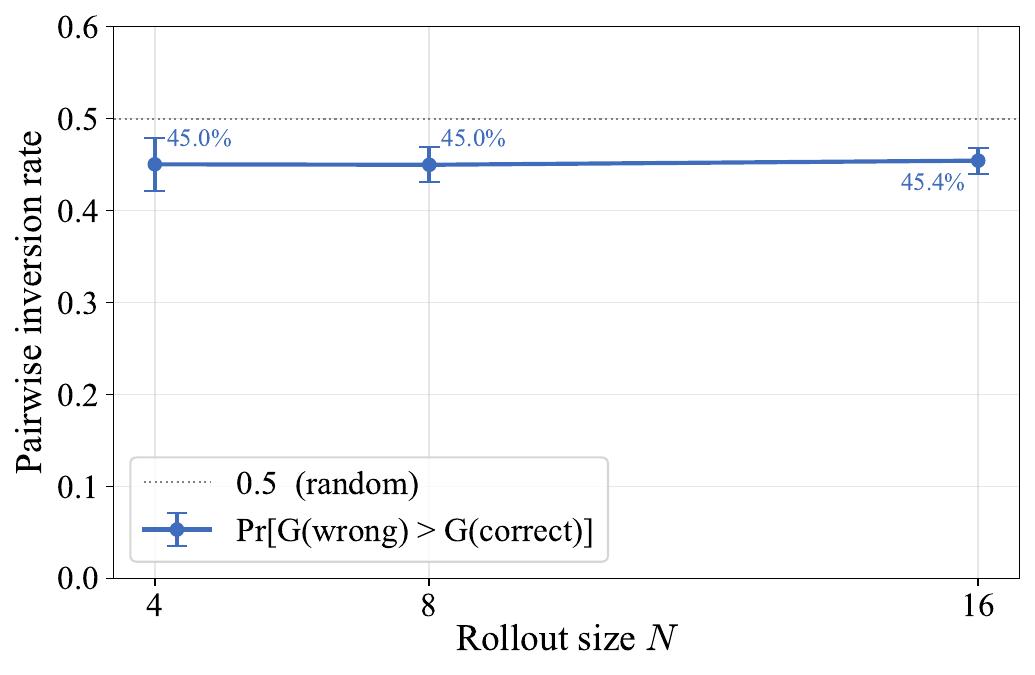}
        \vspace{-1.6em}
        \caption{Pairwise inversion rate vs. rollout size $N$.}
        \label{fig: pairwise_mismatch}
    \end{subfigure}
    \vspace{-0.6em}
    \caption{
        \textbf{Teacher supervision is globally selective but locally unreliable.}
        (\subref{fig: pos_neg_overlap}) Aggregated over all rollouts, correct trajectories attain a slightly higher mean $G_{\mathrm{OPD}}$ than incorrect ones, yet the two distributions overlap heavily; the anomalous mass of incorrect rollouts near $G_{\mathrm{OPD}}\!=\!0$ corresponds to degenerate repetitive generations.
        (\subref{fig: pairwise_mismatch}) Among all comparable correct--incorrect pairs, the fraction in which an incorrect trajectory outranks a correct one stays around $45\%$ and is insensitive to $N$, indicating that the per-prompt ordering in~\cref{eq: opd_order} is frequently violated.
    }
    \vspace{-1mm}
    \label{fig: teacher_reliability_stats}
\end{figure}

\custompara{The teacher is globally selective yet biased by degenerate patterns.}
We next examine the trajectory-level returns aggregated over the entire rollout set, shown in~\cref{fig: pos_neg_overlap}. 
On average, correct rollouts attain a slightly higher $G_{\mathrm{OPD}}$ than incorrect ones, indicating that the teacher does possess a certain degree of discriminative and selective ability at the global level. 
However, the two distributions overlap heavily, and we observe an anomalous concentration of incorrect rollouts near $G_{\mathrm{OPD}}\!=\!0$. 
Inspecting these cases reveals that they are predominantly degenerate generations that fail to terminate under high-temperature decoding and collapse into repetitive or otherwise low-information continuations~\citep{Demystifying_OPD_2026}. On the repetitive spans of such trajectories, the teacher tends to assign disproportionately high token-level rewards, which inflates their trajectory-level return toward zero. 
Such locally favored trajectories are precisely the ones that can outrank correct trajectories and cause the violations analyzed in~\cref{supp:subsec:consistency_violation}.
Accumulated over training, they drive up the response length and repetition rate of the student, one of the very failure modes our method is designed to prevent.

\custompara{Per-prompt orderings are frequently inverted.}
While the teacher is selective in aggregate, the ordering between correct and incorrect trajectories \emph{within an individual prompt} is far less reliable, which is exactly the quantity that order consistency concerns. 
To measure this directly, we consider all comparable pairs, \myie every pair consisting of one correct and one incorrect trajectory under the same prompt, and compute the pairwise inversion rate $\Pr[\,G_{\mathrm{OPD}}(\vq,\boldsymbol{\tau}_{-})\!>\!G_{\mathrm{OPD}}(\vq,\boldsymbol{\tau}_{+})\,]$. 
As shown in~\cref{fig: pairwise_mismatch}, this rate remains around $45\%$ across all rollout sizes and is largely insensitive to $N$, so only about $55\%$ of comparable pairs satisfy the ordering required by~\cref{eq: opd_order}. 
In other words, a large fraction of positive--negative pairs fail to meet~\cref{eq: opd_order}, confirming that the teacher's local supervision is unreliable at the per-prompt level. 
This is the problem our approach sets out to solve, and we now describe the two outcome-guided margin calibration strategies designed for this purpose.

\subsection{Outcome-Guided Margin Calibration}
\label{supp:subsec:margin_calibration}

\begin{algorithm}[!t]
    \caption{Greedy Margin Mask}
    \label{alg:greedy_mask}
    \begin{algorithmic}[1] 
        \Statex \textbf{Inputs:}
        \Statex \quad Prompt $\boldsymbol{q}$ with rollout group $\{\boldsymbol{\tau}_i\}_{i=1}^{N}$, outcome rewards $\{R_{i}\}_{i=1}^{N}$ with $R_{i} \!\in\! \{0,1\}$, min retention ratio $\rho$,
        \Statex \quad trajectory-level distillation returns $\{G_{\mathrm{OPD}}(\boldsymbol{q},\boldsymbol{\tau}_i)\}_{i=1}^{N}$, target margin $\delta$, mode $\in\{\mathrm{MinMax},\mathrm{Mean}\}$.
        \Statex \textbf{Output:} Keep-mask $\{k_i\}_{i=1}^{N} \in \{0,1\}^{N}$ \mycomment{$k_i\!=\!1$ means ``keep trajectory $\boldsymbol{\tau}_i$'' and $k_i\!=\!0$ means ``drop it''}.
        \Statex
        \Statex \textbf{Notation:} For any two subsets $A\!\subseteq\!S_{+}(\boldsymbol{q})$ and $B\!\subseteq\!S_{-}(\boldsymbol{q})$, we define the prompt-level margin
        \Statex \quad $\displaystyle \textsc{Margin}(A,B;\mathrm{MinMax})=\min_{\boldsymbol{\tau}\in A}\! G_{\mathrm{OPD}}(\boldsymbol{q},\boldsymbol{\tau}) - \max_{\boldsymbol{\tau}\in B}\! G_{\mathrm{OPD}}(\boldsymbol{q},\boldsymbol{\tau})$,
        \Statex \quad $\displaystyle \textsc{Margin}(A,B;\mathrm{Mean})=\mean_{\boldsymbol{\tau}\in A}\! G_{\mathrm{OPD}}(\boldsymbol{q},\boldsymbol{\tau}) - \mean_{\boldsymbol{\tau}\in B}\! G_{\mathrm{OPD}}(\boldsymbol{q},\boldsymbol{\tau})$,

        \Statex
        \Function{GreedyMarginMask}{$\boldsymbol{q},\{\boldsymbol{\tau}_i, R_{i}, G_{\mathrm{OPD}}(\boldsymbol{q},\boldsymbol{\tau}_i)\}_{i=1}^{N}, \delta, \rho, \text{mode}$}
        \State \mylinecomment{Step 1: split the group by outcome correctness.}
        \State $S_{+}(\boldsymbol{q}) \leftarrow \{\boldsymbol{\tau}_i \mid R_{i}=1\}$,\quad $S_{-}(\boldsymbol{q}) \leftarrow \{\boldsymbol{\tau}_i \mid R_{i}=0\}$
        \State $N_{+} \leftarrow |S_{+}(\boldsymbol{q})|$,\quad $N_{-} \leftarrow |S_{-}(\boldsymbol{q})|$
        \State $k_i \leftarrow 1, \quad \forall i=1,\ldots,N$ \mycomment{initialize: keep all trajectories}

        \If{$N_{+} = 0$ \textbf{or} $N_{-} = 0$}
        \State \Return $\{k_i\}_{i=1}^{N}$ \mycomment{ordering is not defined; no masking}
        \EndIf

        \State
        \State \mylinecomment{Step 2: sort each side so that the most ordering-violating trajectory is at the front.}
        \State $L_{+}(\boldsymbol{q}) \leftarrow$ sort $S_{+}(\boldsymbol{q})$ by $G_{\mathrm{OPD}}(\boldsymbol{q},\cdot)$ ascending \mycomment{$L_{+}(\boldsymbol{q})[1]$ = correct trajectory with lowest return}
        \State $L_{-}(\boldsymbol{q}) \leftarrow$ sort $S_{-}(\boldsymbol{q})$ by $G_{\mathrm{OPD}}(\boldsymbol{q},\cdot)$ descending \mycomment{$L_{-}(\boldsymbol{q})[1]$ = incorrect trajectory with highest return}

        \State
        \State \mylinecomment{Step 3: iteratively drop the trajectory whose removal increases the margin the most.}
        \While{$\textsc{Margin}(L_{+}(\boldsymbol{q}),L_{-}(\boldsymbol{q});\text{mode}) < \delta$}
            \If{$|L_{+}(\boldsymbol{q})| \le \lceil \rho N_{+}\rceil$ \textbf{and} $|L_{-}(\boldsymbol{q})| \le \lceil \rho N_{-}\rceil$}
            \State \textbf{break} \mycomment{minimum retention ratio reached on both sides}
            \EndIf
            \State
            \State \mylinecomment{Margin gain when the worst correct trajectory $L_{+}(\boldsymbol{q})[1]$ is dropped.}
            \State $\Delta_{+} \leftarrow \textsc{Margin}(L_{+}(\boldsymbol{q})\!\setminus\!\{L_{+}(\boldsymbol{q})[1]\},L_{-}(\boldsymbol{q});\text{mode}) - \textsc{Margin}(L_{+}(\boldsymbol{q}),L_{-}(\boldsymbol{q});\text{mode})$
            \State \mylinecomment{Margin gain when the best incorrect trajectory $L_{-}(\boldsymbol{q})[1]$ is dropped.}
            \State $\Delta_{-} \leftarrow \textsc{Margin}(L_{+}(\boldsymbol{q}),L_{-}(\boldsymbol{q})\!\setminus\!\{L_{-}(\boldsymbol{q})[1]\};\text{mode}) - \textsc{Margin}(L_{+}(\boldsymbol{q}),L_{-}(\boldsymbol{q});\text{mode})$

            \If{$\max(\Delta_{+},\Delta_{-}) \le 0$}
            \State \textbf{break} \mycomment{no single removal can further improve the margin}
            \EndIf

            \If{$\Delta_{+} > \Delta_{-}$ \textbf{and} $|L_{+}(\boldsymbol{q})| > \lceil \rho N_{+}\rceil$}
            \State $\boldsymbol{\tau}_{\mathrm{drop}} \leftarrow \textsc{PopFront}(L_{+}(\boldsymbol{q}))$ \mycomment{greedy drop on the positive side}
            \Else
            \State $\boldsymbol{\tau}_{\mathrm{drop}} \leftarrow \textsc{PopFront}(L_{-}(\boldsymbol{q}))$ \mycomment{greedy drop on the negative side}
            \EndIf
            \State $k_{\,\mathrm{idx}(\boldsymbol{\tau}_{\mathrm{drop}})} \leftarrow 0$ \mycomment{exclude this trajectory from the subsequent gradient update}
        \EndWhile
        \State \Return $\{k_i\}_{i=1}^{N}$
        \EndFunction
    \end{algorithmic}
\end{algorithm}

In this section, we describe the details of the two outcome-guided margin calibration strategies introduced in~\cref{subsec:Teacher_Perspective}: \textit{Margin Mask} and \textit{Margin Shift}. 
Both strategies operate on the trajectory-level distillation returns $\{G_{\mathrm{OPD}}(\boldsymbol{q},\boldsymbol{\tau}_i)\}_{i=1}^{N}$ within a rollout group of a prompt $\boldsymbol{q}$, with the common
goal of enforcing the order-consistency condition $m(\boldsymbol{q})\!\ge\!\delta$ (\cref{eq: ge_condition}). 
They differ in how they repair violations: \textit{Margin Mask} removes the most adversarial trajectories until the condition holds, whereas \textit{Margin Shift} applies a minimal additive correction to restore the margin in closed form.

\custompara{Margin choices: MinMax vs. Mean.}
Following the prompt-level margin in~\cref{eq: prompt_level_margin}, we define the margin between $S_{+}(\boldsymbol{q})$ and $S_{-}(\boldsymbol{q})$ in two modes: the MinMax mode uses $\min_{\boldsymbol{\tau}\in S_{+}}\!{G}_{\mathrm{OPD}}\!-\!\max_{\boldsymbol{\tau}\in S_{-}}\!{G}_{\mathrm{OPD}}$ and characterizes the worst-case ordering violation; the Mean mode uses $\mathrm{mean}_{\boldsymbol{\tau}\in S_{+}}\!{G}_{\mathrm{OPD}}\!-\!\mathrm{mean}_{\boldsymbol{\tau}\in S_{-}}\!{G}_{\mathrm{OPD}}$ and reflects the average-case ordering tendency. MinMax is more conservative (it forces every positive to outrank every negative), while Mean is more lenient and less sensitive to individual outliers.

\custompara{Detailed implementation of margin mask.}
The margin mask strategy discards unreliable trajectories until the prompt-level margin is restored. 
We implement its fine-grained, data-efficient variant as \textit{Greedy Margin Mask}, which removes the single most adversarial trajectory in each iteration rather than discarding the entire group. 
Specifically, given the rollout group $\{\boldsymbol{\tau}_i\}_{i=1}^{N}$ of prompt $\boldsymbol{q}$ with trajectory-level returns $\{G_{\mathrm{OPD}}(\boldsymbol{q},\boldsymbol{\tau}_i)\}_{i=1}^{N}$, we sort the positives in ascending order of $G_{\mathrm{OPD}}$ (so the worst correct trajectory comes first) and the negatives in descending order (so the best incorrect trajectory comes first). 
At each iteration, we compute the margin improvement obtained by removing the front of each sorted list and greedily dropping the side that yields the larger improvement. 
The iteration terminates once (i) the target margin $m(\boldsymbol{q})\!\ge\!\delta$ is satisfied, (ii) no further beneficial removal exists, or (iii) a minimum retention ratio $\rho\!\in\!(0,1)$ is reached to prevent excessive data loss. 
The masked trajectories are excluded from the subsequent gradient update by setting their trajectory-level return to zero, \myie $\widetilde{G}_{\mathrm{OPD}}(\boldsymbol{q},\boldsymbol{\tau}_i)\!=\!k_i\!\cdot\!G_{\mathrm{OPD}}(\boldsymbol{q},\boldsymbol{\tau}_i)$, where $k_i\!\in\!\{0,1\}$ is the keep mask. 
In distributed training, the trajectory-level statistics are aggregated across all ranks via AllReduce so that the masking is deterministic and consistent across devices. The procedure is in~\cref{alg:greedy_mask}.

\custompara{Detailed implementation of margin shift.}
The margin shift strategy applies a minimal additive correction to the trajectory-level returns so that the margin exactly meets the target $\delta$, rather than discarding any sample. 
Given the rollout group $\{\boldsymbol{\tau}_i\}_{i=1}^{N}$ of prompt $\boldsymbol{q}$, we first compute the current margin $m(\boldsymbol{q})$ with the chosen mode (Mean by default). 
If $m(\boldsymbol{q})\!<\!\delta$, we define the required shift as $\lambda(\boldsymbol{q})\!=\!\delta\!-\!m(\boldsymbol{q})\!>\!0$ and distribute it across trajectories in one of three directions: (i) Lift: add $\lambda(\boldsymbol{q})$ to every positive trajectory, \myie $\widetilde{G}_{\mathrm{OPD}}(\boldsymbol{q},\boldsymbol{\tau})\!=\!G_{\mathrm{OPD}}(\boldsymbol{q},\boldsymbol{\tau})\!+\!\lambda(\boldsymbol{q})\mathbf{1}\{R(\boldsymbol{q},\boldsymbol{\tau})\!=\!1\}$, which matches~\cref{eq: margin_shift_lift} in the main text; (ii) Suppress: subtract $\lambda(\boldsymbol{q})$ from every negative trajectory, \myie $\widetilde{G}_{\mathrm{OPD}}(\boldsymbol{q},\boldsymbol{\tau})\!=\!G_{\mathrm{OPD}}(\boldsymbol{q},\boldsymbol{\tau})\!-\!\lambda(\boldsymbol{q})\mathbf{1}\{R(\boldsymbol{q},\boldsymbol{\tau})\!=\!0\}$; and (iii) Spread: split the correction symmetrically, adding $\lambda(\boldsymbol{q})/2$ to positives and subtracting $\lambda(\boldsymbol{q})/2$ from negatives. All three variants (a)~preserve the relative ordering within $S_{+}(\boldsymbol{q})$ and within $S_{-}(\boldsymbol{q})$ respectively, and (b)~guarantee that the calibrated margin equals $\delta$, \myie $\min_{\boldsymbol{\tau}\in S_{+}}\!\widetilde{G}_{\mathrm{OPD}}\!-\!\max_{\boldsymbol{\tau}\in S_{-}}\!\widetilde{G}_{\mathrm{OPD}}\!=\!\delta$. 
In distributed training, the aggregation of trajectory-level statistics and the computation of $\lambda(\boldsymbol{q})$ are done via AllReduce to ensure consistency across devices. The procedure is in~\cref{alg:margin_shift}.

\begin{algorithm}[!t]
    \caption{Margin Shift}
    \label{alg:margin_shift}
    \begin{algorithmic}[1]
        \Statex \textbf{Inputs:}
        \Statex \quad Prompt $\boldsymbol{q}$ with rollout group $\{\boldsymbol{\tau}_i\}_{i=1}^{N}$, outcome rewards $\{R_{i}\}_{i=1}^{N}$ with $R_{i}\!\in\!\{0,1\}$,
        \Statex \quad trajectory-level distillation returns $\{G_{\mathrm{OPD}}(\boldsymbol{q},\boldsymbol{\tau}_i)\}_{i=1}^{N}$,
        \Statex \quad target margin $\delta$, mode $\in\{\mathrm{MinMax},\mathrm{Mean}\}$, direction $\in\{\mathrm{Lift},\mathrm{Suppress},\mathrm{Spread}\}$.

        \Statex \textbf{Output:} Calibrated trajectory-level returns $\{\widetilde{G}_{\mathrm{OPD}}(\boldsymbol{q},\boldsymbol{\tau}_i)\}_{i=1}^{N}$

        \Statex
        \Function{MarginShift}{$\boldsymbol{q},\{\boldsymbol{\tau}_i, R_{i}, G_{\mathrm{OPD}}(\boldsymbol{q},\boldsymbol{\tau}_i)\}_{i=1}^{N}, \delta, \text{mode}, \text{direction}$}
        \State \mylinecomment{Step 1: split the group by outcome correctness.}
        \State $S_{+}(\boldsymbol{q}) \leftarrow \{\boldsymbol{\tau}_i \mid R_{i}=1\}$,\quad $S_{-}(\boldsymbol{q}) \leftarrow \{\boldsymbol{\tau}_i \mid R_{i}=0\}$

        \If{$S_{+}(\boldsymbol{q})=\emptyset$ \textbf{or} $S_{-}(\boldsymbol{q})=\emptyset$}
        \State \Return $\{\widetilde{G}_{\mathrm{OPD}}(\boldsymbol{q},\boldsymbol{\tau}_i) \leftarrow G_{\mathrm{OPD}}(\boldsymbol{q},\boldsymbol{\tau}_i)\}_{i=1}^{N}$ \mycomment{ordering is not defined}
        \EndIf

        \State \
        \State \mylinecomment{Step 2: summarize each side and compute the prompt-level margin $m(\boldsymbol{q})$.}
        \If{mode $=$ $\mathrm{MinMax}$}
        \State $\displaystyle G_{+}(\boldsymbol{q}) \leftarrow \min_{\boldsymbol{\tau}\in S_{+}(\boldsymbol{q})} G_{\mathrm{OPD}}(\boldsymbol{q},\boldsymbol{\tau})$ \mycomment{worst-scoring correct trajectory}
        \State $\displaystyle G_{-}(\boldsymbol{q}) \leftarrow \max_{\boldsymbol{\tau}\in S_{-}(\boldsymbol{q})} G_{\mathrm{OPD}}(\boldsymbol{q},\boldsymbol{\tau})$ \mycomment{best-scoring incorrect trajectory}
        \Else
        \State $\displaystyle G_{+}(\boldsymbol{q}) \leftarrow \mean_{\boldsymbol{\tau}\in S_{+}(\boldsymbol{q})} G_{\mathrm{OPD}}(\boldsymbol{q},\boldsymbol{\tau})$ \mycomment{average correct score}
        \State $\displaystyle G_{-}(\boldsymbol{q}) \leftarrow \mean_{\boldsymbol{\tau}\in S_{-}(\boldsymbol{q})} G_{\mathrm{OPD}}(\boldsymbol{q},\boldsymbol{\tau})$ \mycomment{average incorrect score}
        \EndIf
        \State $m(\boldsymbol{q}) \leftarrow G_{+}(\boldsymbol{q})-G_{-}(\boldsymbol{q})$

        \State \
        \State \mylinecomment{Step 3: additive correction when the margin is below the target.}
        \State $\widetilde{G}_{\mathrm{OPD}}(\boldsymbol{q},\boldsymbol{\tau}_i) \leftarrow G_{\mathrm{OPD}}(\boldsymbol{q},\boldsymbol{\tau}_i),\quad \forall i=1,\ldots,N$ \mycomment{start from the uncalibrated returns}
        \If{$m(\boldsymbol{q}) < \delta$}
        \State $\lambda(\boldsymbol{q}) \leftarrow \delta - m(\boldsymbol{q})$ \mycomment{amount by which the margin falls short of $\delta$}
        \If{direction $=$ $\mathrm{Lift}$}
        \State $\widetilde{G}_{\mathrm{OPD}}(\boldsymbol{q},\boldsymbol{\tau}) \mathrel{+}= \lambda(\boldsymbol{q}),\quad \forall \boldsymbol{\tau}\in S_{+}(\boldsymbol{q})$ \mycomment{pull all correct trajectories up}
        \ElsIf{direction $=$ $\mathrm{Suppress}$}
        \State $\widetilde{G}_{\mathrm{OPD}}(\boldsymbol{q},\boldsymbol{\tau}) \mathrel{-}= \lambda(\boldsymbol{q}),\quad \forall \boldsymbol{\tau}\in S_{-}(\boldsymbol{q})$ \mycomment{push all incorrect trajectories down}
        \Else
        \State $\widetilde{G}_{\mathrm{OPD}}(\boldsymbol{q},\boldsymbol{\tau}) \mathrel{+}= \lambda(\boldsymbol{q})/2,\quad \forall \boldsymbol{\tau}\in S_{+}(\boldsymbol{q})$ \mycomment{split: half up on the positive side, \ldots}
        \State $\widetilde{G}_{\mathrm{OPD}}(\boldsymbol{q},\boldsymbol{\tau}) \mathrel{-}= \lambda(\boldsymbol{q})/2,\quad \forall \boldsymbol{\tau}\in S_{-}(\boldsymbol{q})$ \mycomment{\ldots and half down on the negative side}
        \EndIf
        \EndIf
        \State \Return $\{\widetilde{G}_{\mathrm{OPD}}(\boldsymbol{q},\boldsymbol{\tau}_i)\}_{i=1}^{N}$
        \EndFunction

    \end{algorithmic}
\end{algorithm}

\section{Training Details}
\label{supp:sec:training_details}

In this section, we present details related to training, including the training setup (\cref{supp:subsec:training_setup}), the training datasets (\cref{supp:subsec:training_data}), the training reward acquisition (\cref{supp:subsec:reward_acquisition}), the training pseudocode (\cref{supp:subsec:training_pseudocode}), the training dynamics (\cref{supp:subsec:training_dynamics}), and the training complexity analysis (\cref{supp:subsec:training_complexity}).
These details are provided to enhance the reproducibility of~\texttt{Uni-OPD}.

\subsection{Training Setup}
\label{supp:subsec:training_setup}

To support multi-teacher OPD for both LLMs and MLLMs, we build~\texttt{Uni-OPD} upon a widely used training framework, Miles\footnote{\url{https://github.com/radixark/miles}}. 
Specifically, we use Megatron-LM\footnote{\url{https://github.com/nvidia/megatron-lm}}~\citep{megatron_lm_2019} as the training backend and SGLang\footnote{\url{https://github.com/sgl-project/sglang}} as the rollout inference engine. 
For teacher models, we deploy them as independent SGLang services that can be accessed via HTTP from arbitrary locations to obtain token-level rewards, enabling flexible teacher extensions and scalable multi-teacher integration. 

Each teacher is served behind a pool of SGLang endpoints with client-side shuffled round-robin load balancing, and a lightweight task-to-teacher routing table dispatches every prompt to the teacher best matched to its domain (\myeg math reasoning or code generation), so that new teachers or new tasks can be plugged in by simply extending the registry without touching the training loop. 
Because each teacher only needs to expose its prefill-time \texttt{input\_token\_logprobs}, no gradient, KV cache, or parameter is shared with the student, which keeps teachers fully stateless and decouples their deployment from the trainer. 
As a result, teacher scoring overlaps with student generation and contributes negligible overhead to the overall training throughput.

\custompara{General training hyperparameters.}
All general training settings, including the batch size, rollout numbers, learning rate schedule, optimizer choice, and so on, are identical to those used in ExOPD\footnote{\url{https://github.com/RUCBM/G-OPD}}~\citep{G_OPD_2026}, ensuring a fair and controlled comparison. The prompts used for training are provided in~\cref{prompt: training_templates}.
\begin{custom_template_box}{Training Prompt Template}

    \noindent\textbf{Math Reasoning} \\
    \small
    \texttt{<|im\_start|>user} \\
    \texttt{\{question\}} \\
    Please reason step by step, and put your final answer within \verb|\boxed{}|.\texttt{<|im\_end|>}\\
    \texttt{<|im\_start|>assistant}

    \promptsep

    \noindent\textbf{Code Reasoning} \\
    \small
    \texttt{<|im\_start|>user} \\
    \texttt{\{question\}} \\
    Write Python code to solve the problem. Present the code in \\
    \texttt{```python} \\
    \texttt{Your code} \\
    \texttt{```} \\
    at the end. \\
    You need to think first then write the Python code.\texttt{<|im\_end|>} \\
    \texttt{<|im\_start|>assistant}

    \promptsep

    \noindent\textbf{Multimodal Math Reasoning} \\
    \small
    \texttt{<|im\_start|>user} \\
    \texttt{<image>} \\
    \texttt{\{question\}} \\
    Please solve the problem step by step and put your answer in one \verb|\boxed{}|.\texttt{<|im\_end|>}\\
    \texttt{<|im\_start|>assistant}
    \label{prompt: training_templates}
\end{custom_template_box}

\custompara{RL training setup.} Teacher models are trained using reinforcement learning (RL). Detailed training settings of the teacher models are provided in~\cref{tab: RL_training_setup}.

\begin{table}[htbp]
    \centering
    \small
    \caption{
        \textbf{Teacher model training configuration with GRPO.}
    }
    \vspace{-2mm}
    \label{tab: RL_training_setup}
    \setlength{\tabcolsep}{10pt}
    {\renewcommand{\arraystretch}{0.95}
        \resizebox{0.85\columnwidth}{!}{
            \begin{tabular}{llc}
                \toprule
                \textbf{Group}   &
                \textbf{Setting} &
                \textbf{Value}
                \\
                \midrule
                \multirow{4}{*}{\textit{Model}}
                                 & \multirow{2}{*}{Base model}
                                 & LLM: Math, Code: Qwen3-4B                                              \\
                                 &
                                 & MLLM: Math, Logic, Document: Qwen3-VL-4B-Inst.                         \\

                                 & \multirow{2}{*}{Training steps}
                                 & LLM: Math, Code: 500, 300                                              \\
                                 &
                                 & MLLM:  Math, Logic, Document: 300, 300, 160                            \\
                \midrule
                \multirow{8}{*}{\textit{Optimization}}
                                 & Tensor Parallelism (TP)                        & 2                     \\
                                 & Micro batch size / GPU                         & 1                     \\
                                 & Training batch size                            & 128                   \\
                                 & Learning rate                                  & $1\times10^{-6}$      \\
                                 & Warm-up steps                                  & 0                     \\
                                 & LR schedule                                    & Constant              \\
                                 & ZeRO stage                                     & 3                     \\
                                 & Optimizer                                      & Adam                  \\

                \midrule
                \multirow{2}{*}{\textit{Sequence}}
                                 & Max prompt length                              & 2048                  \\
                                 & Max response length                            & 16384                 \\

                \midrule
                \multirow{5}{*}{\textit{RL Algorithm}}
                                 & Advantage estimator                            & GRPO                  \\
                                 & GRPO clip ratio                                & 0.2                   \\
                                 & Use KL in reward                               & False                 \\
                                 & KL loss coefficient                            & 0.0                   \\
                                 & Entropy coefficient                            & 0.0                   \\

                \midrule
                \multirow{4}{*}{\textit{Rollout}}
                                 & Samples per prompt ($N$)                       & 8                     \\
                                 & Temperature                                    & 1.0                   \\
                                 & Top-$p$                                        & 0.95                  \\
                                 & Top-$k$                                        & 50                    \\

                \midrule
                \multirow{1}{*}{\textit{Hardware}}
                                 & GPUs                                           & $16\times$ NVIDIA H20 \\

                \bottomrule
            \end{tabular}
        }
    }
\end{table}

\custompara{OPD training setup.} For OPD, we inherit most hyperparameters (\myeg learning rate, optimizer, and sequence lengths) from the teacher RL setup in~\cref{tab: RL_training_setup}, so that the student is trained under the same optimization regime as its teachers. 
The OPD-specific entries, including the training batch size, the number of on-policy samples per prompt, the online correctness-aware filter, and the margin calibration configuration, are summarized in~\cref{tab: OPD_training_setup}. 
Concretely, we use a training batch size of 64 and sample $N\! = \!16$ on-policy rollouts per prompt, which we find provides a good trade-off between return estimation quality and computational efficiency (see the ablation in~\cref{tab: ablation_rollout_num}). The online correctness-aware filter is applied in sample filter mode with a target correct-to-incorrect ratio of $1{:}1$ within each training batch, following~\cref{supp:subsec:online_data_balancing}. 
For margin calibration (\cref{supp:subsec:margin_calibration}), we adopt \emph{group-level mean} normalization in both domains, while the shift direction and target margin are tuned per domain: for the textual domain, we use Spread with $\delta\!=\!0.4$, and for the multimodal domain, we use Lift with $\delta\!=\!0$.

\begin{table}[t!]
    \centering \small
    \caption{
        \textbf{OPD training configuration.}
        Most hyperparameters inherit from the teacher RL setup in~\cref{tab: RL_training_setup}; only the entries that differ between OPD and RL are listed here.
    }
    \vspace{-2mm}
    \label{tab: OPD_training_setup}
    \setlength{\tabcolsep}{10pt}
    {\renewcommand{\arraystretch}{0.95}
        \begin{tabular}{llcc}
            \toprule
            \textbf{Group}   &
            \textbf{Setting} &
            \textbf{Textual} &
            \textbf{Multimodal}
            \\
            \midrule
            \multirow{2}{*}{\textit{Optimization}}
                             & Training batch size      & 64              & 64            \\
                             & Samples per prompt ($N$) & 16              & 16            \\
            \midrule
            \multirow{2}{*}{\textit{Online filter}}
                             & Filter mode              & Sample filter   & Sample filter \\
                             & Correct/Incorrect ratio  & $1{:}1$         & $1{:}1$       \\
            \midrule
            \multirow{4}{*}{\textit{Margin calibration}}
                             & Scope                    & Group           & Group         \\
                             & Mode                     & Mean            & Mean          \\
                             & Direction                & \textsc{Spread} & \textsc{Lift} \\
                             & Target margin $\delta$   & $0.4$           & $0$           \\

            \bottomrule
        \end{tabular}
    }
    \vspace{-2mm}
\end{table}

\subsection{Training Data}
\label{supp:subsec:training_data}

\custompara{Textual math reasoning data.}
We use a subset of the DeepMath dataset~\citep{he2025deepmath} with difficulty level $\!\geq\!6$ to train mathematical reasoning ability, comprising 57.0K samples.

\custompara{Textual code generation data.} 
We use the Code subset of the Eurus-2-RL-Data dataset~\citep{cui2025process} with 25.3K samples to train code generation ability.

\custompara{Multimodal math reasoning data.}
For multimodal math reasoning tasks, we draw 14.8K samples from the OpenMMReasoner-RL dataset\footnote{\url{https://huggingface.co/datasets/OpenMMReasoner/OpenMMReasoner-RL-74K}}, covering MMK12, WeMath-Standard, and WeMath-Pro subsets.

\custompara{Multimodal logic reasoning data.}
We collect 14.8K samples spanning AlgoPuzzle, PuzzleVQA, and ThinkLite-VL-Hard subsets from the OpenMMReasoner-RL-74K dataset.

\custompara{Multimodal document understanding data.}
 We include 14.6K document understanding samples, obtained by 15\% sampling from the TQA subset of OpenMMReasoner with ChartQA~\citep{masry2022chartqa} and InfographicsVQA~\citep{mathew2022infographicvqa} training sets.

\subsection{Training Reward Acquisition}
\label{supp:subsec:reward_acquisition}

In this section, we describe how training rewards are obtained for different data sources. For textual math reasoning tasks, we use the rule-based verifier provided by DeepMath\footnote{\url{https://github.com/zwhe99/DeepMath}} to determine whether generated answers are correct. For textual code generation tasks, we use the rule-based verifier provided by PRIME\footnote{\url{https://github.com/PRIME-RL/PRIME}} to evaluate the correctness of generated code. For multimodal tasks, we use the verifier released by OpenMMReasoner\footnote{\url{https://github.com/EvolvingLMMs-Lab/OpenMMReasoner}} to assess whether generated answers are correct.

\subsection{Training Pseudocode}
\label{supp:subsec:training_pseudocode}

The full training procedure of \texttt{Uni-OPD} is summarized in~\cref{alg:opd}. In brief, the procedure (1) samples a prompt batch with offline difficulty-aware balancing (\cref{supp:subsec:offline_data_balancing}); (2) rolls out $N$ trajectories per prompt and computes the trajectory-level distillation return ${G}_{\mathrm{OPD}}$ from teacher--student log-probability differences~(\cref{eq: distillation_return}); (3) applies online correctness-aware balancing across the batch (\cref{supp:subsec:online_data_balancing}); (4) calibrates $G_{\mathrm{OPD}}$ via the prompt-level margin $m(\boldsymbol{q})$~(\cref{eq: prompt_level_margin}) using either \textsc{Greedy Margin Mask} (\cref{alg:greedy_mask}) or \textsc{Margin Shift} (\cref{alg:margin_shift}); and (5) broadcasts the calibrated returns to token-level advantages and updates the student $\pi_{\boldsymbol{\theta}}$.

\begin{algorithm}[!t]
    \caption{
        \texttt{Uni-OPD}: Outcome-guided Policy Distillation with Margin Calibration
    }
    \label{alg:opd}
    \small
    \begin{algorithmic}[1]
        \Statex \textbf{Input:}
        \Statex \quad Teacher $\pi_{\text{T}}$, student $\pi_{\boldsymbol{\theta}}$, dataset $\mathcal{D}$, group size $N$, target margin $\delta$, calibration mode $\!\in\!\{\textsc{Mask},\textsc{Shift}\}$, learning rate $\eta$.
        \Statex \textbf{Output:} Updated student parameters $\boldsymbol{\theta}$.

        \Statex
        \Function{UniOPD}{$\pi_{\text{T}}, \pi_{\boldsymbol{\theta}}, \mathcal{D}, N, \delta, \text{mode}, \eta$}
        \State \mylinecomment{\textbf{Offline difficulty-aware data balancing} (once before training; see~\cref{supp:subsec:offline_data_balancing}).}
        \State Sample a prompt batch $\mathcal{B}\subset\mathcal{D}$ with rebalanced difficulty distribution

        \State \
        \While{not converged}
        \State \mylinecomment{\textbf{Rollout and token-level scoring} (per prompt).}
        \ForAll{prompt $\boldsymbol{q}\in\mathcal{B}$}
        \State Rollout $N$ trajectories $\{\boldsymbol{\tau}_i\}_{i=1}^{N}\sim\pi_{\boldsymbol{\theta}}(\cdot\mid\boldsymbol{q})$
        \For{$i=1,\ldots,N$}
        \State Obtain outcome reward $R_{i}=R(\boldsymbol{q},\boldsymbol{\tau}_i)\in\{0,1\}$
        \ForAll{token $o_{t}\in\boldsymbol{\tau}_i$}
        \State $r^{\mathrm{OPD}}_{t}(\boldsymbol{\tau}_i) \leftarrow \log\pi_{\text{T}}(o_{t}\mid\boldsymbol{q},\boldsymbol{o}_{<t}) - \log\pi_{\boldsymbol{\theta}}(o_{t}\mid\boldsymbol{q},\boldsymbol{o}_{<t})$ \mycomment{token-level OPD reward}
        \EndFor
        \State \mylinecomment{Trajectory-level distillation return (\cref{eq: distillation_return}).}
        \State $G_{\mathrm{OPD}}(\boldsymbol{q},\boldsymbol{\tau}_i) \leftarrow \dfrac{1}{|\boldsymbol{\tau}_i|}\sum_{t=1}^{|\boldsymbol{\tau}_i|} r^{\mathrm{OPD}}_{t}(\boldsymbol{\tau}_i)$
        \EndFor
        \State Partition: $S_{+}(\boldsymbol{q})\leftarrow\{\boldsymbol{\tau}_i\mid R_{i}=1\}$,\quad $S_{-}(\boldsymbol{q})\leftarrow\{\boldsymbol{\tau}_i\mid R_{i}=0\}$ \mycomment{correct / incorrect trajectory sets}
        \EndFor

        \State \
        \State \mylinecomment{\textbf{Online correctness-aware data balancing} (across the batch; see~\cref{supp:subsec:online_data_balancing}).}
        \State $\mathcal{B}\leftarrow\textsc{OnlineCorrectnessAwareDataBalancing}\bigl(\mathcal{B},\{R_{i}\}_{\boldsymbol{q},i}\bigr)$

        \State \
        \State \mylinecomment{\textbf{Outcome-guided margin calibration} (per prompt; \cref{eq: prompt_level_margin,eq: ge_condition}).}
        \ForAll{prompt $\boldsymbol{q}\in\mathcal{B}$}
        \State Compute prompt-level margin $m(\boldsymbol{q})=\min_{\boldsymbol{\tau}\in S_{+}(\boldsymbol{q})} {G}_{\mathrm{OPD}}(\boldsymbol{q},\boldsymbol{\tau}) - \max_{\boldsymbol{\tau}\in S_{-}(\boldsymbol{q})} G_{\mathrm{OPD}}(\boldsymbol{q},\boldsymbol{\tau})$
        \If{mode $=\textsc{Mask}$}
        \State $\{k_{\boldsymbol{q},i}\}_{i=1}^{N}\leftarrow\textsc{GreedyMarginMask}(\boldsymbol{q},\{\boldsymbol{\tau}_i,R_{i},G_{\mathrm{OPD}}(\boldsymbol{q},\boldsymbol{\tau}_i)\}_{i=1}^{N},\delta,\rho,\text{mode})$ \mycomment{\cref{alg:greedy_mask}}
        \State $\widetilde{G}_{\mathrm{OPD}}(\boldsymbol{q},\boldsymbol{\tau}_i)\leftarrow k_{\boldsymbol{q},i}\cdot G_{\mathrm{OPD}}(\boldsymbol{q},\boldsymbol{\tau}_i),\quad \forall i=1,\ldots,N$ \mycomment{zero out masked trajectories}
        \Else
        \State $\{\widetilde{G}_{\mathrm{OPD}}(\boldsymbol{q},\boldsymbol{\tau}_i)\}_{i=1}^{N}\leftarrow\textsc{MarginShift}(\boldsymbol{q},\{\boldsymbol{\tau}_i,R_{i},G_{\mathrm{OPD}}(\boldsymbol{q},\boldsymbol{\tau}_i)\}_{i=1}^{N},\delta,\text{mode},\text{direction})$ \mycomment{\cref{alg:margin_shift}}
        \EndIf
        \EndFor

        \State \
        \State \mylinecomment{\textbf{Token-level broadcasting and policy update.}}
        \ForAll{prompt $\boldsymbol{q}\in\mathcal{B}$, rollout $i=1,\ldots,N$, token $o_t\in\boldsymbol{\tau}_i$}
        \State $\widetilde{A}_{t}(\boldsymbol{q},\boldsymbol{\tau}_i)\leftarrow \widetilde{G}_{\mathrm{OPD}}(\boldsymbol{q},\boldsymbol{\tau}_i)$ \mycomment{broadcast calibrated trajectory return to all tokens}
        \EndFor
        \State $\mathcal{L}(\boldsymbol{\theta})= -\,\mathbb{E}_{\boldsymbol{q},\boldsymbol{\tau}_i,t}\!\left[\widetilde{A}_{t}(\boldsymbol{q},\boldsymbol{\tau}_i)\,\log\pi_{\boldsymbol{\theta}}(o_{t}\mid\boldsymbol{q},\boldsymbol{o}_{<t})\right]$
        \State $\boldsymbol{\theta}\leftarrow\boldsymbol{\theta}-\eta\,\nabla_{\boldsymbol{\theta}}\mathcal{L}(\boldsymbol{\theta})$ \mycomment{one gradient step on the student}
        \EndWhile
        \State \Return $\boldsymbol{\theta}$
        \EndFunction
    \end{algorithmic}
\end{algorithm}

\subsection{Training Dynamics}
\label{supp:subsec:training_dynamics}

\begin{figure}[!t]
    \centering
    \includegraphics[width=1.0\linewidth]{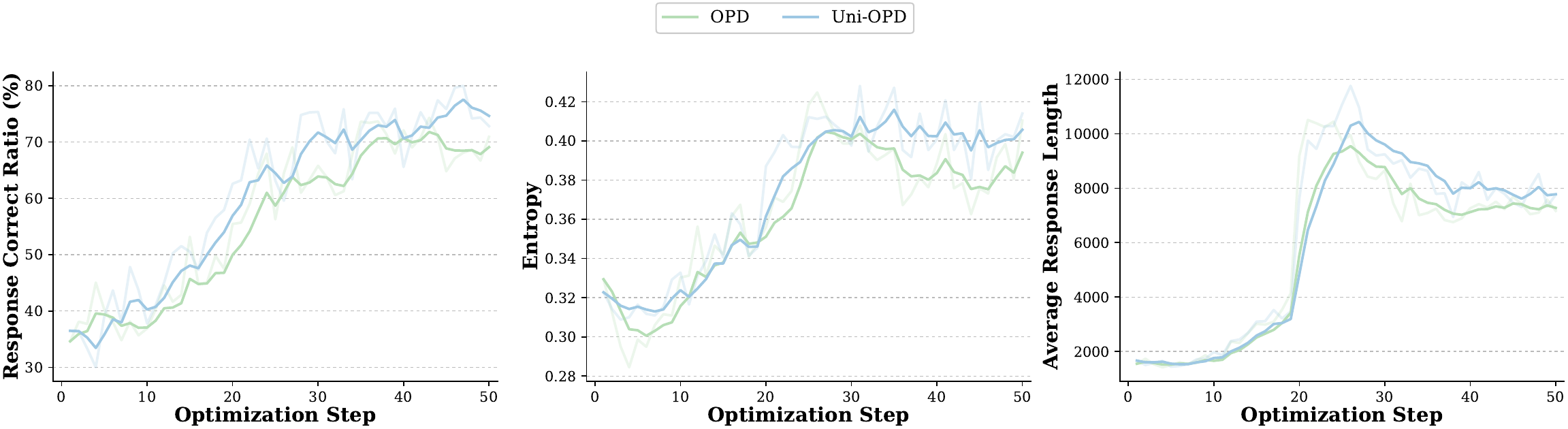}
    \vspace{-1.6em}
    \caption{\textbf{Training dynamics of OPD and \texttt{Uni-OPD} for multi-teacher distillation.} 
    We track three indicators along the optimization trajectory: response correctness (\%), 
policy entropy, and average response length.}
    \label{fig:train_dynamics}
    \vspace{-2mm}
\end{figure}

\Cref{fig:train_dynamics} demonstrates the effectiveness of \texttt{Uni-OPD} along three complementary axes. 
From a comparable starting point ($\sim$35\% correct, entropy $\sim$0.33, length $\sim$1.6k), \texttt{Uni-OPD} converges to a substantially higher response-correct ratio than OPD, peaking at $80.6\%$ versus $75.2\%$ and averaging $75.5\%$ over the final 10 steps versus OPD's $69.1\%$ (+6.4 absolute points). 
Crucially, this accuracy gain is not obtained by sacrificing exploration: policy entropy rises mildly under both methods, with \texttt{Uni-OPD} maintaining a marginally higher steady-state value, ruling out the entropy-collapse failure mode that typically plagues teacher-driven training. 
Meanwhile, the average response length grows from $\sim$1.6k to $\sim$8k tokens, with \texttt{Uni-OPD} producing slightly longer outputs than OPD (7.8k vs.\ 7.1k), indicating that the model learns to perform more elaborate reasoning rather than collapsing to short, high-confidence shortcuts. 
Together, these trends suggest that \texttt{Uni-OPD} provides a 
consistent improvement over OPD without adverse effects on exploration 
or response length.

\subsection{Training Complexity}
\label{supp:subsec:training_complexity}

Beyond vanilla OPD, \texttt{Uni-OPD} introduces lightweight components on top of the standard per-iteration cost during training: \emph{online correctness-aware data balancing} (per batch; \cref{supp:subsec:online_data_balancing}), and \emph{outcome-guided margin calibration} via Margin Mask / Shift (per prompt; \cref{supp:subsec:margin_calibration}). 
Let $B$ be the training batch size (number of prompts) and $N$ be the rollout group size. The online balancing only resamples prompts based on their precomputed $\{R_{i}\}$, costing $O(BN)$ per iteration. 
Margin Mask and Margin Shift both operate on the $N$ trajectory-level returns within each prompt group: Margin Shift is $O(N)$ per prompt, while the greedy variant of Margin Mask is at most $O(N^{2})$ per prompt in the worst case (typically $N\!\le\!16$ in our setup).

In contrast, the dominant per-iteration cost of OPD comes from two stages whose complexity scales linearly with the total number of rollout tokens $T_{\text{tok}}\! = \!\sum_{i=1}^{BN} |\boldsymbol{\tau}_i|$ and cubically with the hidden size $d$: (i) sampling $BN$ on-policy rollouts from the student, and (ii) running a teacher prefill pass over these rollouts to obtain token-level log-probabilities, each of order $O(T_{\text{tok}}\,d^{2})$ for transformer forward passes. 
Typical numbers in our setup ($B\! = \!64$, $N\! = \!16$, average length $\sim$8k) give $T_{\text{tok}}$ on the order of $8\!\times\!10^{6}$ tokens per iteration. 
All of \texttt{Uni-OPD}'s additional computation scales with the number of trajectories rather than the number of tokens, involves only scalar comparisons and additions, and is therefore several orders of magnitude cheaper than the rollout and teacher-scoring stages that OPD already pays. 
In practice, we observe that enabling all three components adds less than $1\%$ wall-clock overhead per iteration relative to vanilla OPD, while delivering the accuracy improvements reported in~\cref{supp:subsec:training_dynamics} and the main experiments. Thus~\texttt{Uni-OPD} offers a favorable accuracy--compute trade-off: a negligible compute surcharge in exchange for consistently better final performance.
\section{Evaluation Details}
\label{supp:sec:evaluation_details}

\subsection{Evaluation Benchmarks}
\label{supp:subsec:evaluation_benchmarks}

We evaluate our \texttt{Uni-OPD} on a comprehensive benchmark suite spanning textual and multimodal capabilities, organized along five capability axes:
\begin{itemize}[
        label=\raisebox{0.5ex}{\tiny$\bullet$},
        leftmargin=1em,
        itemsep=2pt,
        parsep=0pt,
        topsep=0pt,
        partopsep=0pt
    ]
    \item \textbf{Textual Math Reasoning:}
          \begin{itemize}[label=-, leftmargin=1.2em, itemsep=1pt]
              \item AIME (2024/2025)~\citep{aime2024}: A prestigious high school mathematics competition featuring challenging problems that test deep mathematical reasoning.
              \item HMMT25 (Feb \& Nov)~\citep{balunovic2025matharena}: Contest-level benchmarks designed to rigorously evaluate advanced reasoning across algebra, geometry, combinatorics, and other domains.
          \end{itemize}
    \item \textbf{Textual Code Generation:}
          \begin{itemize}[label=-, leftmargin=1.2em, itemsep=1pt]
              \item HumanEval+~\citep{liu2023your}: A set of 164 hand-written programming problems evaluating functional correctness, covering language understanding, reasoning, algorithms, and basic mathematics.
              \item MBPP+~\citep{liu2023your}: A collection of $\sim$1,000 crowd-sourced Python tasks targeting entry-level programming skills, including fundamentals and standard library usage.
              \item LiveCodeBench (v6)~\citep{jain2024livecodebench}: A contamination-free and continuously updated benchmark assessing not only code generation but also execution, self-repair, and test prediction.
          \end{itemize}

    \item \textbf{Multimodal Math Reasoning:}
          \begin{itemize}[label=-, leftmargin=1.2em, itemsep=1pt]
              \item MathVision~\citep{wang2024measuring}: A curated dataset of 3,040 visual problems sourced from real competitions, spanning 16 disciplines and multiple difficulty levels for evaluating multimodal mathematical reasoning.
              \item DynaMath~\citep{DynaMath_2025}: A dynamically generated benchmark based on 501 seed question generators, enabling diverse and scalable evaluation through multiple sampled variants.
              \item WeMath~\citep{qiao2025we}: A large-scale benchmark with 6.5K visual math problems organized into 67 hierarchical knowledge concepts, designed to analyze problem-solving processes. 
          \end{itemize}
    \item \textbf{Multimodal Logic Reasoning:}
          \begin{itemize}[label=-, leftmargin=1.2em, itemsep=1pt]
              \item LogicVista~\citep{xiao2024logicvista}: A benchmark for evaluating multimodal logical reasoning across 5 task types and 9 capabilities using annotated multiple-choice questions with human reasoning.
              \item VisuLogic~\citep{xu2025visulogic}: A challenging visual reasoning benchmark focusing on reasoning directly from visual inputs, with tasks that are difficult to express textually and expose gaps in current MLLMs.
          \end{itemize}
    \item \textbf{Document Understanding:}
          \begin{itemize}[label=-, leftmargin=1.2em, itemsep=1pt]
              \item AI2D~\citep{kembhavi2016diagram}: A diagram understanding benchmark focusing on parsing diagram structure and reasoning over relationships between components via question answering.
              \item ChartQA~\citep{masry2022chartqa}: A benchmark for question answering over charts, requiring complex visual and logical reasoning over both chart structure and underlying data.
              \item DocVQA~\citep{mathew2021docvqa}: A large-scale document visual question answering dataset over document images, emphasizing structural and textual understanding.
              \item InfoVQA~\citep{mathew2022infographicvqa}: A benchmark on infographic understanding that requires joint reasoning over layout, text, and visual elements with an emphasis on multi-step reasoning.
          \end{itemize}
\end{itemize}


\subsection{Evaluation Setup}
\label{supp:subsec:evaluation_setup}

\custompara{Textual evaluations.}
For all textual evaluations, we use a sampling temperature of 1.0, top-$p$ of 1.0, a maximum generation length of 16{,}384 tokens, and a fixed random seed of 42. We use the vLLM inference engine to perform sampling.
For math reasoning benchmarks, we sample $N = 32$ solutions per problem, while for code generation benchmarks, we sample $N = 4$ solutions per problem.
For evaluation, we adopt \texttt{Math-Verify}\footnote{\url{https://github.com/huggingface/Math-Verify}} as a rule-based verifier for math reasoning tasks.
For code generation, we use the \texttt{EvalPlus}\footnote{\url{https://github.com/evalplus/evalplus}} and \texttt{LiveCodeBench}\footnote{\url{https://github.com/livecodebench/livecodebench}} frameworks to assess functional correctness.
For all main results, we report the average accuracy across sampled solutions (\myie pass@1), and compute pass@k as:

\vspace{-1.0em}
\begin{equation}
    \text{pass}@k = 1 - \frac{\binom{N - c}{k}}{\binom{N}{k}}\, ,
    \label{eq: pass_at_k}
\end{equation}
\par\nobreak\vspace{-0.5em}

where \(N\) is the number of samples and \(c\) is the number of correct solutions.

\begin{table*}[!t]
    \centering \small
    \caption{
        \textbf{Reported evaluation metrics for different benchmark datasets.}
        We summarize the primary metrics used for performance reporting across math, logic, and document understanding tasks.
    }
    \vspace{-2mm}
    \label{tab: mm_eval_metrics}
    \setlength{\tabcolsep}{6.5pt}
    \begin{tabular}{lcccc}
        \toprule
        \textbf{Category} & \textbf{Tasks}            & \textbf{Filter}  & \textbf{$N$-Shot} & \textbf{Reported Metric}   \\
        \midrule
        \multirow{3}{*}{\makecell{Multimodal \nextline Math Reasoning}}
                          & MathVision Test           & none             & 0                 & mathvision\_standard\_eval \\
                          & DynaMath Reasoning        & none             & 0                 & dynamath\_average          \\
                          & WeMath TestMini Reasoning & none             & 0                 & acc\_score                 \\
        \midrule
        \multirow{3}{*}{\makecell{Multimodal \nextline Logic Reasoning}}
                          & LogicVista Reasoning      & none             & 0                 & acc\_score                 \\
                          & LogicVista Reasoning      & none             & 0                 & format\_score              \\
                          & VisuLogic                 & none             & 0                 & visulogic\_acc             \\
        \midrule
        \multirow{4}{*}{\makecell{Document \nextline Understanding}}
                          & AI2D                      & flexible-extract & 0                 & exact\_match               \\
                          & ChartQA                   & none             & 0                 & relaxed\_human\_split      \\
                          & DocVQA Val                & none             & 0                 & anls                       \\
                          & InfoVQA Val               & none             & 0                 & anls                       \\
        \bottomrule
    \end{tabular}
\end{table*}
\custompara{Multimodal evaluations.}
For multimodal evaluations, we adopt the widely used \texttt{LMMs-Eval}\footnote{\url{https://github.com/evolvinglmms-lab/lmms-eval}}~\citep{LMMs_Eval_2024} framework and strictly follow its official evaluation protocols and configurations. The reported evaluation metrics are summarized in~\cref{tab: mm_eval_metrics}.

\clearpage
\section{Further Evaluations}
\label{supp:sec:further_evaluations}

\subsection{More Evaluation Results}
\label{supp:subsec:more_evaluation_results}

\begin{table*}[!t]
    \centering \small
    \caption{
        \textbf{Performance of \textit{Qwen3-1.7B Student} under math reasoning and code generation benchmarks.}
        Teacher models (\myie \textit{Qwen3-4B-Math-RL} and \textit{Qwen3-4B-Code-RL}) are developed through domain-specific RL. The performance of teacher models is denoted by the ``RL'' type.
    }
    \vspace{-2mm}
    \label{tab: llm_full_4B_teacher_1.7B_student}
    \setlength{\tabcolsep}{5pt}
    \begin{tabular}{lcccccccccc}
        \toprule
        \multirow{2.5}{*}{\vspace{-3mm}\textbf{Method}} &
        \multicolumn{5}{c}{\textbf{Math Reasoning}}     &
        \multicolumn{4}{c}{\textbf{Code Generation}}
        \\
        \cmidrule(lr){2-6}
        \cmidrule(lr){7-10}
                                                        &
        \makecell{AIME \nextline 2024}                  &
        \makecell{AIME \nextline 2025}                  &
        \makecell{HMMT \nextline 25 Feb.}               &
        \makecell{HMMT \nextline 25 Nov.}               &
        \textbf{Avg.}                                   &
        \makecell{Human \nextline Eval+}                &
        MBPP+                                           &
        LCB                                             &
        \textbf{Avg.}
        \\
        \midrule

        Student                                         & 13.9          & 11.1          & 5.6           & 4.9           & 8.9           & 61.9          & 53.4          & 11.9          & 42.4          \\
        Teacher                                         & 60.1          & 55.1          & 32.5          & 38.5          & 46.6          & 85.2          & 69.8          & 26.6          & 60.5          \\
        \midrule
        \multicolumn{10}{c}{\emph{\quad \textbf{Single--Teacher Distillation}}}                                                                                                                         \\

        OPD                                             & 42.3          & \textbf{35.4} & 18.4          & 19.1          & 28.8          & 71.8          & 58.2          & 26.7          & 52.5          \\
        \rowcolor{table-color}
        \texttt{Uni-OPD}                                & \textbf{42.6} & 35.1          & \textbf{20.8} & \textbf{20.9} & \textbf{29.9} & \textbf{73.0} & \textbf{60.0} & \textbf{28.1} & \textbf{53.7} \\
        \midrule
        \multicolumn{10}{c}{\emph{\quad \textbf{Multi--Teacher Distillation}}}                                                                                                                          \\
        OPD                                             & 40.3          & 32.4          & \textbf{20.0} & \textbf{20.3} & 28.3          & \textbf{73.2} & 59.1          & 25.7          & 52.7          \\
        \rowcolor{table-color}
        \texttt{Uni-OPD}                                & \textbf{44.0} & \textbf{35.1} & 19.5          & 19.8          & \textbf{29.6} & 72.9          & \textbf{60.5} & \textbf{28.0} & \textbf{53.8} \\
        \bottomrule
    \end{tabular}
\end{table*}
\begin{table*}[!t]
    \centering \small
    \caption{
        \textbf{Performance of \textit{Qwen3-VL-2B-Instruct Student} under math reasoning, logic reasoning, and document understanding benchmarks.}
        Teacher models (\myie \textit{Qwen3-VL-4B-Instruct-Math-RL}, \textit{Qwen3-VL-4B-Instruct-Logic-RL} and \textit{Qwen3-VL-4B-Instruct-Document-RL}) are developed through domain-specific RL. \textbf{Avg.} denotes the mean score within each category.
    }
    \vspace{-2mm}
    \label{tab: vlm_full_4B_Teacher_2B_Student}
    \setlength{\tabcolsep}{2.5pt}
    \begin{tabular}{l cccc cccc ccccc}
        \toprule
        \multirow{3}{*}{\textbf{Method}}             &
        \multicolumn{4}{c}{\textbf{Math Reasoning}}  &
        \multicolumn{4}{c}{\textbf{Logic Reasoning}} &
        \multicolumn{5}{c}{\textbf{Document Understanding}}                                                                                                                                                                                                                                                                     \\
        \cmidrule(lr){2-5}
        \cmidrule(lr){6-9}
        \cmidrule(lr){10-14}
                                                     & Math          & Dyna          & We            & \multirow{2}{*}{\textbf{Avg.}} & LogicVista    & LogicVista    & Visu          & \multirow{2}{*}{\textbf{Avg.}} & \multirow{2}{*}{AI2D} & Chart         & Doc           & Info          & \multirow{2}{*}{\textbf{Avg.}} \\
                                                     & Vision        & Math          & Math          &                                & Accuracy      & Format        & Logic         &                                &                       & QA            & VQA           & VQA           &                                \\
        \midrule
        Student                                      & 11.1          & 49.1          & 48.6          & 36.3                           & 32.4          & 59.1          & 6.4           & 32.6                           & 73.4                  & 66.1          & 92.8          & 72.4          & 76.2                           \\
        Teacher                                      & 47.2          & 65.3          & 79.5          & 64.0                           & 52.5          & 73.8          & 27.4          & 51.2                           & 82.5                  & 76.4          & 95.1          & 81.6          & 83.9                           \\
        \midrule
        \multicolumn{14}{c}{\emph{\quad \textbf{Single--Teacher Distillation}}}                                                                                                                                                                                                                                                 \\
        OPD                                          & 24.4          & 54.5          & 64.8          & 47.9                           & 35.3          & 61.6          & 26.8          & 41.2                           & 76.1                  & 66.0          & \textbf{93.0} & 72.2          & 76.8                           \\
        \rowcolor{table-color}
        \texttt{Uni-OPD}                             & \textbf{25.5} & \textbf{55.2} & \textbf{65.0} & \textbf{48.6}                  & \textbf{36.8} & \textbf{65.2} & \textbf{27.6} & \textbf{43.2}                  & \textbf{76.7}         & \textbf{66.6} & 92.9          & \textbf{72.6} & \textbf{77.2}                  \\
        \midrule
        \multicolumn{14}{c}{\emph{\quad \textbf{Multi--Teacher Distillation}}}                                                                                                                                                                                                                                                  \\
        OPD                                          & 15.2          & 50.2          & 57.6          & 41.0                           & 38.0          & 65.2          & \textbf{27.2} & 43.4                           & \textbf{76.2}         & 66.1          & 92.9          & 72.5          & 76.9                           \\
        \rowcolor{table-color}
        \texttt{Uni-OPD}                             & \textbf{18.7} & \textbf{51.2} & \textbf{58.7} & \textbf{43.9}                  & \textbf{42.0} & \textbf{69.8} & 27.0          & \textbf{46.3}                  & 76.0                  & \textbf{66.5} & \textbf{93.0} & \textbf{72.6} & \textbf{77.0}                  \\
        \bottomrule
    \end{tabular}
\end{table*}

\custompara{Single-teacher and multi-teacher distillation on LLMs and MLLMs.}
We further evaluate \texttt{Uni-OPD} under both single-teacher and multi-teacher distillation settings on LLMs and MLLMs.
As shown in~\cref{tab: llm_full_4B_teacher_1.7B_student,tab: vlm_full_4B_Teacher_2B_Student}, our method consistently outperforms the standard OPD baseline across all domains and settings.
On the LLM student (\myie \textit{Qwen3-1.7B}), \texttt{Uni-OPD} improves the average scores on both math reasoning and code generation under single-teacher and multi-teacher distillation.
On the MLLM student (\myie \textit{Qwen3-VL-2B-Instruct}), it delivers consistent gains across math reasoning, logic reasoning, and document understanding. Further, it narrows the gap to the teacher ensemble under multi-teacher distillation.
Consistent improvements in smaller students provide strong empirical evidence for our dual-perspective approach, confirming that student exploration and teacher reliability are indeed the fundamental drivers of successful and reliable distillation.

\begin{table*}[htbp]
    \centering \small
    \caption{
        \textbf{Performance of \textit{Qwen3-VL-4B-Instruct Student} under code generation and logic reasoning benchmarks.}
        Teacher models (\myie \textit{Qwen3-VL-4B-Instruct-Code-RL} and \textit{Qwen3-VL-4B-Instruct-Logic-RL}) are developed through domain-specific RL. The performance of teacher models is denoted by the ``RL'' type.
    }
    \vspace{-2mm}
    \label{tab: vlm_llm_opd_code_logic}
    \setlength{\tabcolsep}{8pt}
    \begin{tabular}{l cccc cccc}
        \toprule
        \multirow{3}{*}{\textbf{Method}}                  &
        \multicolumn{4}{c}{\textbf{Code Generation}}      &

        \multicolumn{4}{c}{\textbf{Logic Reasoning}}
        \\
        \cmidrule(lr){2-5} \cmidrule(lr){6-9}
                                                          &
        \multirow{2}{*}{\makecell{Human \nextline Eval+}} &
        \multirow{2}{*}{MBPP+}                            &
        \multirow{2}{*}{LCB}                              &
        \multirow{2}{*}{\textbf{Avg.}}                    & LogicVista    & LogicVista    & Visu          &
        \multirow{2}{*}{\textbf{Avg.}}
        \\
                                                          &               &               &               &

                                                          & Accuracy      & Format        & Logic         &
        \\
        \midrule
        \makecell[l]{Student}                             & 76.8          & 70.0          & 37.0          & 61.3          & 49.9          & 66.4          & 25.1          & 47.0          \\
        \makecell[l]{Teacher}                             & 82.2          & 70.5          & 40.1          & 64.3          & 52.5          & 73.8          & 27.4          & 51.2          \\
        \midrule
        \multicolumn{9}{c}{\emph{\quad \textbf{Multi--Teacher Distillation}}}                                                                                                             \\
        OPD                                               & 79.0          & 68.5          & 39.6          & 62.4          & 50.0          & 69.3          & 27.3          & 48.9          \\
        \rowcolor{table-color}
        \texttt{Uni-OPD}                                  & \textbf{79.4} & \textbf{69.2} & \textbf{41.4} & \textbf{63.3} & \textbf{52.0} & \textbf{73.8} & \textbf{28.0} & \textbf{51.3} \\
        \bottomrule
    \end{tabular}
\end{table*}

\custompara{Cross-modal distillation on code generation and logic reasoning.}
Beyond the cross-modal distillation on math reasoning and code generation, we further conduct cross-modal distillation on code generation and logic reasoning.
Specifically, we combine text-only code data with multimodal logic reasoning data, and jointly distill from two domain-specific teachers (\textit{Qwen3-VL-4B-Instruct-Code-RL} and \textit{Qwen3-VL-4B-Instruct-Logic-RL}) into a single \textit{Qwen3-VL-4B-Instruct} student.
As shown in~\cref{tab: vlm_llm_opd_code_logic}, \texttt{Uni-OPD} outperforms the standard OPD baseline on both the code generation and logic reasoning averages, with the largest gain on LCB (39.6 $\rightarrow$ 41.4) and LogicVista Accuracy (50.0 $\rightarrow$ 52.0).
These results confirm that \texttt{Uni-OPD} effectively integrates heterogeneous text-only and multimodal data under a single training run, further supporting its applicability to cross-modal distillation.

\subsection{Downstream Task Evaluation}
\label{supp:subsec:downstream_task_evaluation}

\begin{table*}[htbp]
    \centering \small
    \caption{
        \textbf{General downstream task performance.}
        Evaluation on 8 general benchmarks to ensure general-purpose capabilities are maintained after OPD.
    }
    \vspace{-2mm}
    \label{tab: downstream_task_evaluation}
    \setlength{\tabcolsep}{4pt}
    \begin{tabular}{lccccccccc}
        \toprule
        \textbf{Model}
                         &
        MMLU             &
        ARC              &
        HellaSwag        &
        TruthfulQA       &
        Winogrande       &
        GSM8K            &
        CommonsenseQA    &
        IFEval           &
        \textbf{Avg.}                                                                   \\
        \midrule
        Qwen3-4B         & 68.3 & 80.7 & 68.4 & 54.8 & 66.6 & 84.2 & 75.8 & 88.9 & 73.5 \\
        \midrule
        Math Teacher     & 68.4 & 80.8 & 68.5 & 54.3 & 66.0 & 86.7 & 75.4 & 89.2 & 73.7 \\
        Code Teacher     & 68.3 & 80.2 & 68.3 & 54.8 & 65.7 & 85.8 & 75.7 & 89.7 & 73.6 \\
        OPD              & 68.3 & 80.3 & 68.4 & 54.6 & 66.5 & 88.6 & 75.5 & 89.2 & 73.9 \\
        \rowcolor{table-color}
        \texttt{Uni-OPD} & 68.3 & 80.3 & 68.3 & 54.6 & 66.0 & 88.6 & 75.7 & 89.2 & 73.9 \\
        \bottomrule
    \end{tabular}
\end{table*}

\custompara{Evaluation on general capabilities.}
To assess the impact of OPD on general downstream performance of the policy model, we evaluate the models on a diverse set of benchmarks from the Hugging Face Open LLM Leaderboard~\citep{Open_LLM_Leaderboard_2023} following recent studies~\citep{Uni_DPO_2025, SimPO_2024}. Specifically, we report results on MMLU~\citep{MMLU_2020}, ARC~\citep{ARC_2018}, HellaSwag~\citep{HellaSwag_2019}, TruthfulQA~\citep{TruthfulQA_2022}, Winogrande~\citep{Winograde_2012}, GSM8K~\citep{GSM8K_2021}, CommonsenseQA~\citep{CommonsenseQA_2019}, and IFEval~\citep{IFEval_2023}. We strictly follow the standard evaluation protocols provided by the lm-evaluation-harness system\footnote{\url{https://github.com/EleutherAI/lm-evaluation-harness}}. For IFEval, we report the inst\_level\_loose\_acc.

The results are presented in~\cref{tab: downstream_task_evaluation}. Overall, \texttt{Uni-OPD} not only outperforms OPD and domain-specific teachers on math reasoning and code generation benchmarks demonstrated in the main text, but also retains strong performance across a wide range of downstream tasks. These results suggest that OPD serves as a general and effective framework for improving LLM performance beyond task-specific settings.

\subsection{Further Ablation}
\label{supp:subsec:further_ablation}

\begin{table*}[htbp]
    \centering \small
    \caption{
        \textbf{Effectiveness validation of margin shift across different hyperparameters.}
        We conduct single-teacher distillation experiments with a Qwen3-4B Student using individual math and code teachers.
    }
    \vspace{-2mm}
    \label{tab: ablation_shift}
    \setlength{\tabcolsep}{4.5pt}
    \begin{tabular}{l ccccc cccc}
        \toprule
        \multirow{3}{*}{\textbf{Configuration}}     &
        \multicolumn{5}{c}{\textbf{Math Reasoning}} &
        \multicolumn{4}{c}{\textbf{Code Generation}}
        \\
        \cmidrule(lr){2-6}
        \cmidrule(lr){7-10}
                                                    &
        \makecell{AIME \nextline 2024}              &
        \makecell{AIME \nextline 2025}              &
        \makecell{HMMT \nextline 25 Feb.}           &
        \makecell{HMMT \nextline 25 Nov.}           &
        \textbf{Avg.}                               &
        \makecell{Human \nextline Eval+}            &
        MBPP+                                       &
        LCB                                         &
        \textbf{Avg.}
        \\
        \midrule
        OPD (no shift)                              & 57.9 & 52.4 & 30.2 & 37.8 & 44.6          & 82.6 & 68.8 & 25.7 & 59.0          \\
        Global + Mean + Lift                        & 61.8 & 55.2 & 34.8 & 39.4 & 47.8          & 85.7 & 71.4 & 25.7 & 60.9          \\
        Global + MinMax + Lift                      & 62.4 & 57.3 & 32.2 & 38.2 & 47.5          & 85.8 & 71.8 & 26.7 & 61.4          \\
        Group + MinMax + Spread                     & 63.4 & 56.7 & 33.4 & 39.0 & 48.1          & 86.9 & 70.6 & 26.7 & 61.4          \\
        \midrule
        \rowcolor{table-color}
        Group + Mean + Spread (ours)                & 62.7 & 56.3 & 34.4 & 39.2 & \textbf{48.2} & 88.3 & 72.3 & 26.7 & \textbf{62.4} \\
        \bottomrule
    \end{tabular}
\end{table*}

\custompara{Hyperparameter analysis for margin shift.}
As shown in~\cref{tab: ablation_shift}, we compare four variants of margin shift against the OPD baseline across math reasoning and code generation benchmarks. The shift scope (Global vs. Group), normalization mode (Mean vs. MinMax), and shift direction (Lift vs. Spread) are ablated systematically. All shift variants consistently outperform the vanilla OPD baseline, demonstrating the general effectiveness of margin shift.
Among the variants, \textit{Group + Mean + Spread} achieves the best average performance on both code generation (62.4) and math reasoning (48.2), indicating that group-level mean normalization with bidirectional shifting provides a more calibrated return signal.
Applying the shift to both correct and incorrect responses (Spread) proves beneficial over unidirectional shifting (Lift), and group-level statistics generalize better than global ones when reward distributions vary across prompts.
Furthermore, we observe that MinMax-based normalization and global-scope statistics are susceptible to outlier return values, as extreme return values within a batch can distort the shift magnitude and destabilize training.
In contrast, group-level mean normalization produces more robust and consistent return estimates, contributing to stable optimization throughout training.

\begin{wraptable}{r}{0.48\textwidth}
    \centering
    \small
    \caption{
        \textbf{The effects of rollout number.} The global batch size is fixed at $N \times bs = 1024$ throughout.
    }
    \vspace{-2.5mm}
    \label{tab: ablation_rollout_num}
    \setlength{\tabcolsep}{3.5pt}
    \rowcolors{1}{white}{white}
    \begin{tabular}{lccccc}
        \toprule
        \textbf{Method}                   &
        \makecell{AIME \nextline 2024}    &
        \makecell{AIME \nextline 2025}    &
        \makecell{HMMT \nextline 25 Feb.} &
        \makecell{HMMT \nextline 25 Nov.} &
        \textbf{Avg.}
        \\
        \midrule
        Student (4B)                      & 23.0 & 19.3 & 12.3 & 9.2  & 15.9 \\
        \midrule
        \multicolumn{6}{c}{\textbf{\textit{OPD}}}                            \\
        $N=4$, $bs=256$                   & 60.1 & 55.1 & 32.5 & 29.6 & 44.3 \\
        $N=8$, $bs=128$                   & 59.8 & 52.9 & 29.6 & 35.8 & 44.5 \\
        $N=16$, $bs=64$                   & 57.9 & 52.4 & 30.2 & 37.8 & 44.6 \\
        $N=32$, $bs=32$                   & 58.3 & 51.2 & 30.6 & 36.9 & 44.3 \\
        \midrule
        \multicolumn{6}{c}{\textbf{\textit{OPD + Margin shift}}}             \\
        $N=4$, $bs=256$                   & 57.9 & 52.4 & 33.2 & 37.8 & 45.3 \\
        $N=8$, $bs=128$                   & 62.5 & 55.4 & 31.9 & 39.2 & 47.3 \\
        \rowcolor{table-color}
        $N=16$, $bs=64$                   & 62.7 & 56.3 & 34.4 & 39.2 & 48.2 \\
        $N=32$, $bs=32$                   & 63.1 & 55.4 & 34.2 & 39.6 & 48.1 \\
        \bottomrule
    \end{tabular}
\end{wraptable}
\noindent \textbf{Hyperparameter analysis for rollout number $N$.} \xspace 
As shown in \cref{tab: ablation_rollout_num}, we ablate the rollout number $N$ in OPD while keeping the global batch size fixed at 1024 (\myie $N \times bs = 1024$), so that increasing $N$ comes at the cost of a smaller per-step batch size $bs$. For the OPD baseline, performance remains largely stable across all values of $N$ (44.3--44.6 avg.), suggesting that the base method is relatively insensitive to this trade-off. In contrast, OPD with margin shift benefits notably from larger rollout groups: average performance improves from 45.3 at $N{=}4$ to 48.2 at $N{=}16$, as more responses per prompt yield more reliable relative return estimation for the margin-based calibration. We find that increasing $N$ from 16 to 32 yields comparable performance. Considering return estimation quality, training stability, and computational efficiency, we therefore set $N{=}16$ as our default.
\section{Related Work}
\label{supp:sec:related_work}

\subsection{Multimodal Large Language Models}
\label{supp:subsec:multimodal_large_language_models}

Large Language Models (LLMs) have undergone rapid development in recent years~\citep{Llama_2_2023, GPT4_2023, Llama_3_2024, GPT_4o_2024, Qwen2_5_2024, Llama_3_1_2024, Qwen3_2025, GPT3_2020, Gemini_1_5_2024, Claude_2023, Claude2_2023, Claude3_2024, Deepseek_v3_2024, Deepseek_R1_2025, li2025logits}, significantly improving reasoning capabilities. Meanwhile, MLLMs have also seen substantial progress~\citep{radford2021learning, shao2024explore, wang2025declip, tian2019learning, liu2024typicalness, yang2024unified, Uni_DPO_2025, HunyuanOCR_2025}. Leveraging advances in LLMs, multimodal large language models (MLLMs) further integrate visual and textual representations through cross-modal learning, achieving strong multimodal understanding and generation capabilities.
A key driver of this success lies in the combination of large-scale self-supervised pre-training on diverse corpora and subsequent high-quality supervised fine-tuning (SFT), which enables LLMs and MLLMs to exhibit strong generalization and emergent capabilities in real-world tasks~\citep{Qwen2_VL_2024, Qwen_VL_2023, Qwen2_5_VL_2024, LLaVA_v1_2023, LLaVA_v1_5_2024, LLaVA_NeXT_2024, InstructBLIP_2023, OpenAI_GPT4V_2023, MiniGPT_4_2024, qu2025does, yang2023improved, zhong2024lyra, yang2023lidar, yang2025visionzip, lai2024lisa, peng2025mitigating, hou2026seeing, ChartArena_2026}.
Building upon these foundations, KD has emerged as an important paradigm for transferring sophisticated reasoning capabilities from teacher models to more efficient students.
Among various distillation strategies, OPD has recently emerged as a mainstream post-training paradigm for both LLMs and MLLMs.
In the on-policy setting, however, the effectiveness of distillation is tied to both the quality of student exploration and the reliability of teacher feedback.
In this work, we present a dual-perspective optimization strategy from both the student and teacher sides to improve data suitability and training stability in OPD.

\subsection{Reinforcement Learning}
\label{supp:subsec:reinforcement_learning}

By optimizing trajectories sampled from the current policy, on-policy RL alleviates distribution mismatch and is often instantiated with verifiable or outcome-based rewards in reasoning tasks.
Notable methods include GRPO~\citep{shao2024deepseekmath} for critic-free grouped optimization and GSPO~\citep{zheng2025group} for sequence-level stable optimization.
Recently, some works have also combined RLVR with OPD, such as Self-Distilled RLVR~\citep{yang2026self} and OpenClaw-RL~\citep{wang2026openclaw}.
In our work, we use GRPO to obtain stronger domain-specific teachers and use the corresponding reward models as global guidance for return calibration in OPD.

\subsection{On-Policy Distillation}
\label{supp:subsec:on_policy_distillation}

Early OPD work, such as MiniLLM~\citep{gu2023minillm} and GKD~\citep{agarwal2024policy}, establishes the basic paradigm of using teacher feedback on student-generated trajectories under a reverse KL objective.
Recent studies further broaden this paradigm from multiple perspectives.
In self-distillation methods, OPSD~\citep{zhao2026self} uses privileged information;
SDFT~\citep{shenfeld2026self} allows the student to absorb knowledge from retrieved demonstrations while reducing forgetting.
SDPO~\citep{hubotter2026reinforcement} treats the current model itself as a self-teacher;
OPCD~\citep{ye2026policy} internalizes context knowledge into model parameters by minimizing reverse KL between the student and a context-conditioned teacher on the student's trajectories.
Regarding teacher access, black-box OPD~\citep{ye2025black} introduces a discriminator-guided framework that does not require teacher logits.
Several works also focus on improving optimization and efficiency.
ExOPD~\citep{G_OPD_2026} reformulates OPD as weighted dense RL;
Fast and Effective OPD~\citep{zhang2026fast} improves efficiency through prefix-only distillation;
KDFlow~\citep{zhang2026kdflow} provides an extensible distillation framework supporting both off-policy and on-policy training;
MiMo-V2-Flash~\citep{xiao2026mimo} introduces multi-teacher OPD, enabling effective capability merging across domains.
Revisiting OPD~\citep{fu2026revisit} identifies three failure modes of sampled-token OPD and proposes teacher top-$K$ local support matching for more stable training.
CoPD~\citep{Gu2026CoEvolvingPD} interleaves branch-specific RLVR with bidirectional mutual OPD, keeping parallel experts behaviorally close enough for effective capability transfer.
Li et al.~\citep{li2026rethinking} rethink OPD in terms of its phenomenology, mechanisms, and training recipes.

Recently, OPD has also begun to extend beyond text-only settings.
VOLD~\citep{bousselham2025vold} transfers reasoning ability from text teachers to vision-language students through a two-stage pipeline that combines cold-start alignment with GRPO and OPD.
Video-OPD~\citep{li2026video} adapts OPD to long-video grounding and introduces a curriculum that filters unreliable teacher signals.
X-OPD~\citep{cao2026x} further extends OPD to speech through cross-modal alignment.
In contrast, our work focuses on developing a unified OPD framework with an open recipe for both LLMs and MLLMs.
\section{Case Studies}
\label{supp:sec:case_studies}

We provide qualitative case studies of \texttt{Uni-OPD}, standard OPD, and the Student model across both LLM and MLLM benchmarks, covering textual math reasoning, code generation, logical reasoning, multimodal math reasoning, and chart understanding.

\begin{figure}
    \centering
    \includegraphics[width=1.0\linewidth]{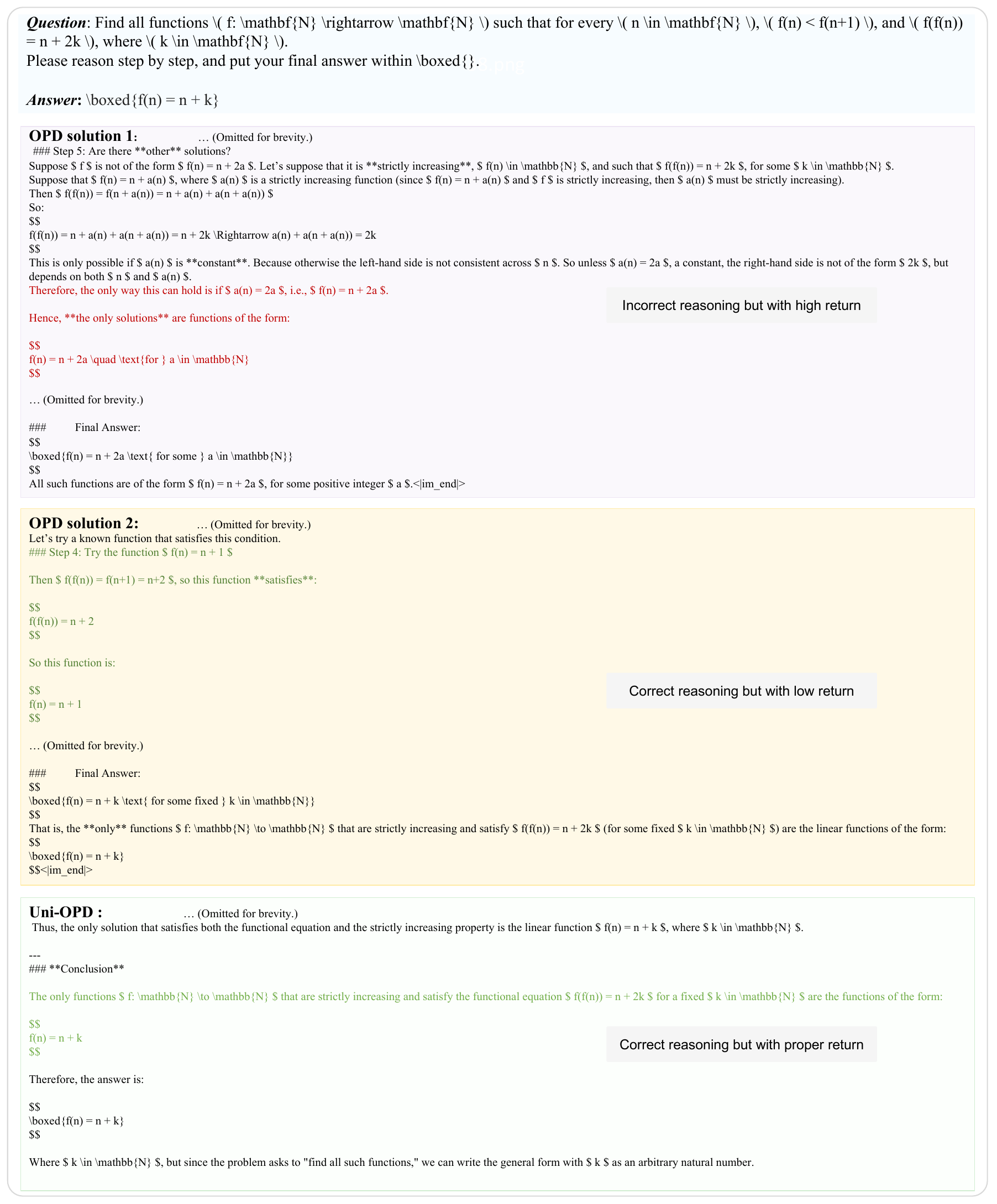}
    \vspace{-1.6em}
    \caption{
        \textbf{Comparison of math reasoning outputs between OPD and \texttt{Uni-OPD}.}
        In this case, standard OPD assigns \emph{high} returns to incorrect reasoning trajectories and \emph{low} returns to correct ones.
        In contrast, our \texttt{Uni-OPD} performs outcome-guided margin calibration to restore order consistency between correct and incorrect trajectories, yielding a reliable supervision signal that ultimately improves both efficiency and correctness of the generated solutions.
        On this question, we further measure \texttt{pass@1} accuracy over $64$ rollouts: standard OPD reaches $79.69\%$, while our \texttt{Uni-OPD} attains $82.81\%$, further validating the effectiveness of the proposed strategy.}
    \label{fig:example_math_adv_three_results}
\end{figure}

\begin{figure}[htbp]
    \centering
    \includegraphics[width=1.0\linewidth]{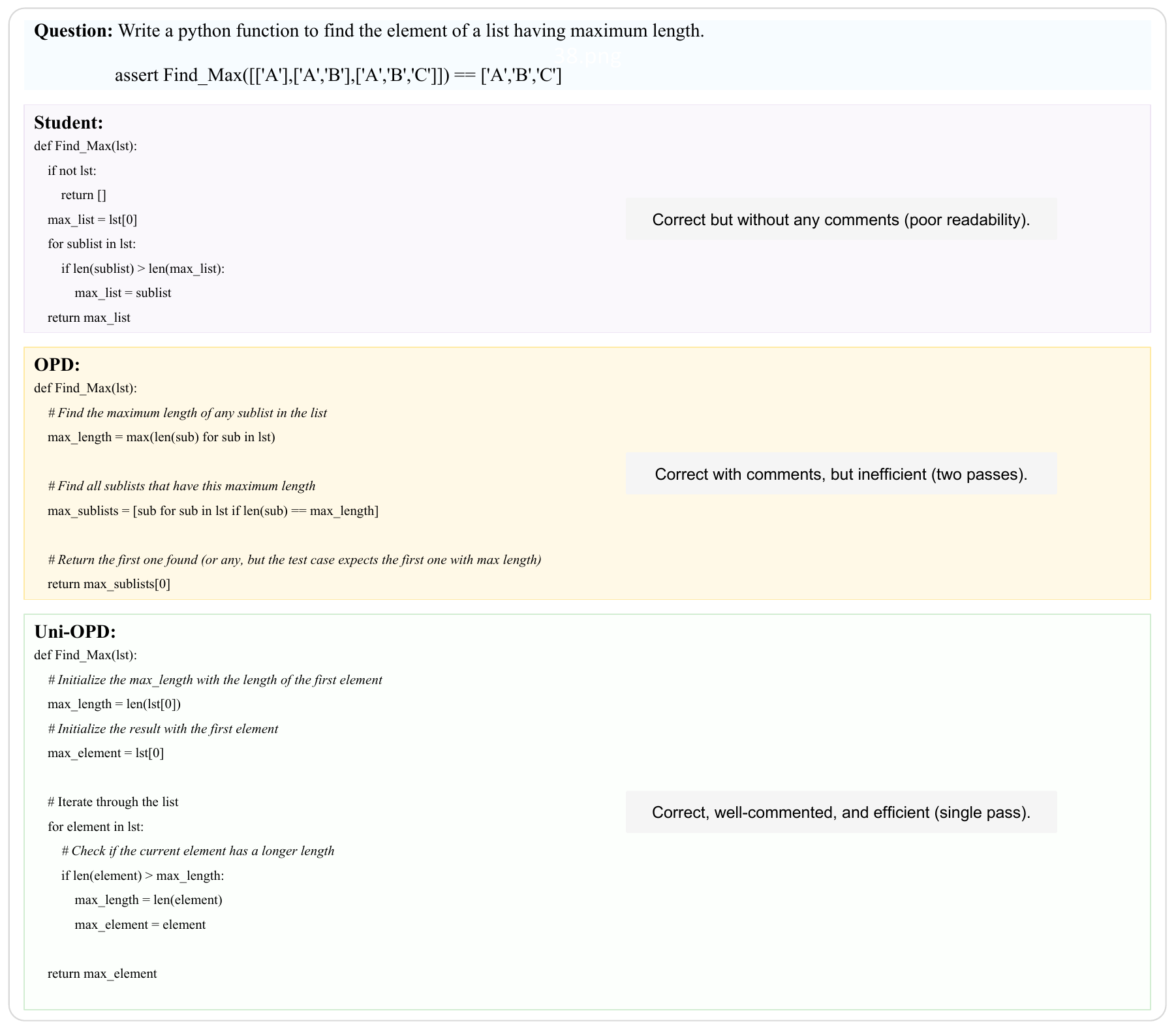}
    \vspace{-1.6em}
    \caption{
        \textbf{Comparison of code generation for the \texttt{Find\_Max} task.}
        While the Student model produces correct logic with limited readability, the OPD baseline introduces redundant computation (two passes) despite adding comments. Our \texttt{Uni-OPD} generates a superior solution that is both computationally efficient (single pass) and well-commented, demonstrating its effectiveness in aligning with complex task requirements.
    }
    \label{fig:example_code_generation}
\end{figure}

We first revisit the math reasoning case in~\cref{fig:example_math_adv_three_results}, and provide a detailed output comparison of standard OPD and our \texttt{Uni-OPD}.
Standard OPD assigns \emph{high} returns to incorrect trajectories and \emph{low} returns to correct ones.
Furthermore, the code generation case in~\cref{fig:example_code_generation} highlights \texttt{Uni-OPD}'s ability to balance algorithmic efficiency and code readability.
These case studies demonstrate how our dual-perspective optimization--specifically by restoring order consistency through margin calibration--leads to more reliable and high-quality model outputs.

\begin{figure}[htbp]
    \centering
    \includegraphics[width=1.0\linewidth]{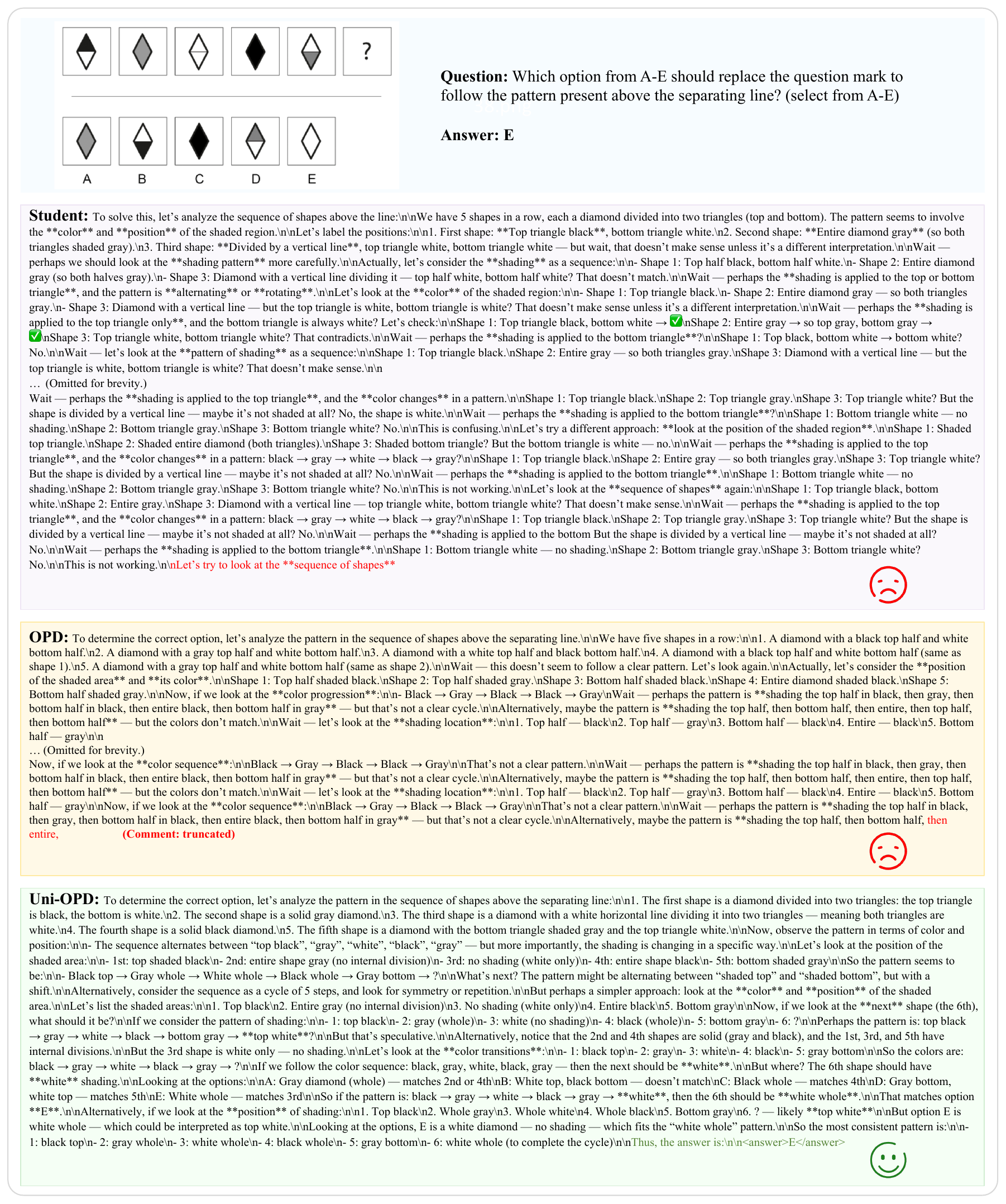}
    \vspace{-1.6em}
    \caption{
        \textbf{Example output of LogicVista.}
        The Student model produces an incorrect reasoning trace and arrives at the wrong answer. Standard OPD overthinks the problem, generating an excessively long response that is truncated without producing a final answer. In contrast, \texttt{Uni-OPD} reasons concisely and correctly answers the question.
    }
    \label{fig:example_logicvista_uniopd_correct_effective}
\end{figure}

\begin{figure}
    \centering
    \includegraphics[width=1.0\linewidth]{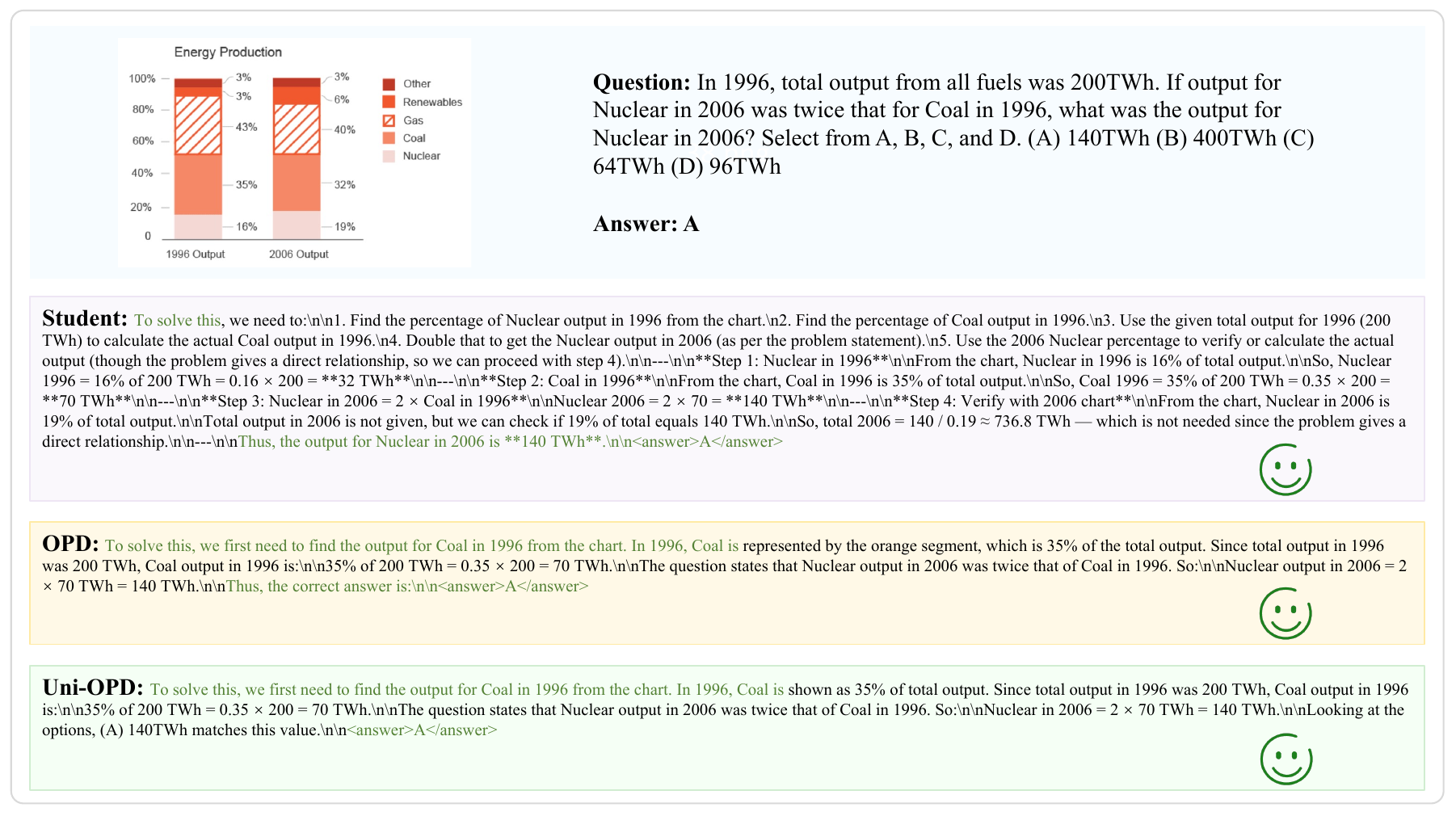}
    \vspace{-1.6em}
    \caption{
        \textbf{Example output of LogicVista.}
        All three models correctly answer this multi-step arithmetic reasoning question. OPD and \texttt{Uni-OPD} both reason concisely, with \texttt{Uni-OPD} being slightly more token-efficient.
    }
    \label{fig:example_logicvista_all_correct}
\end{figure}

\begin{figure}
    \centering
    \includegraphics[width=1.0\linewidth]{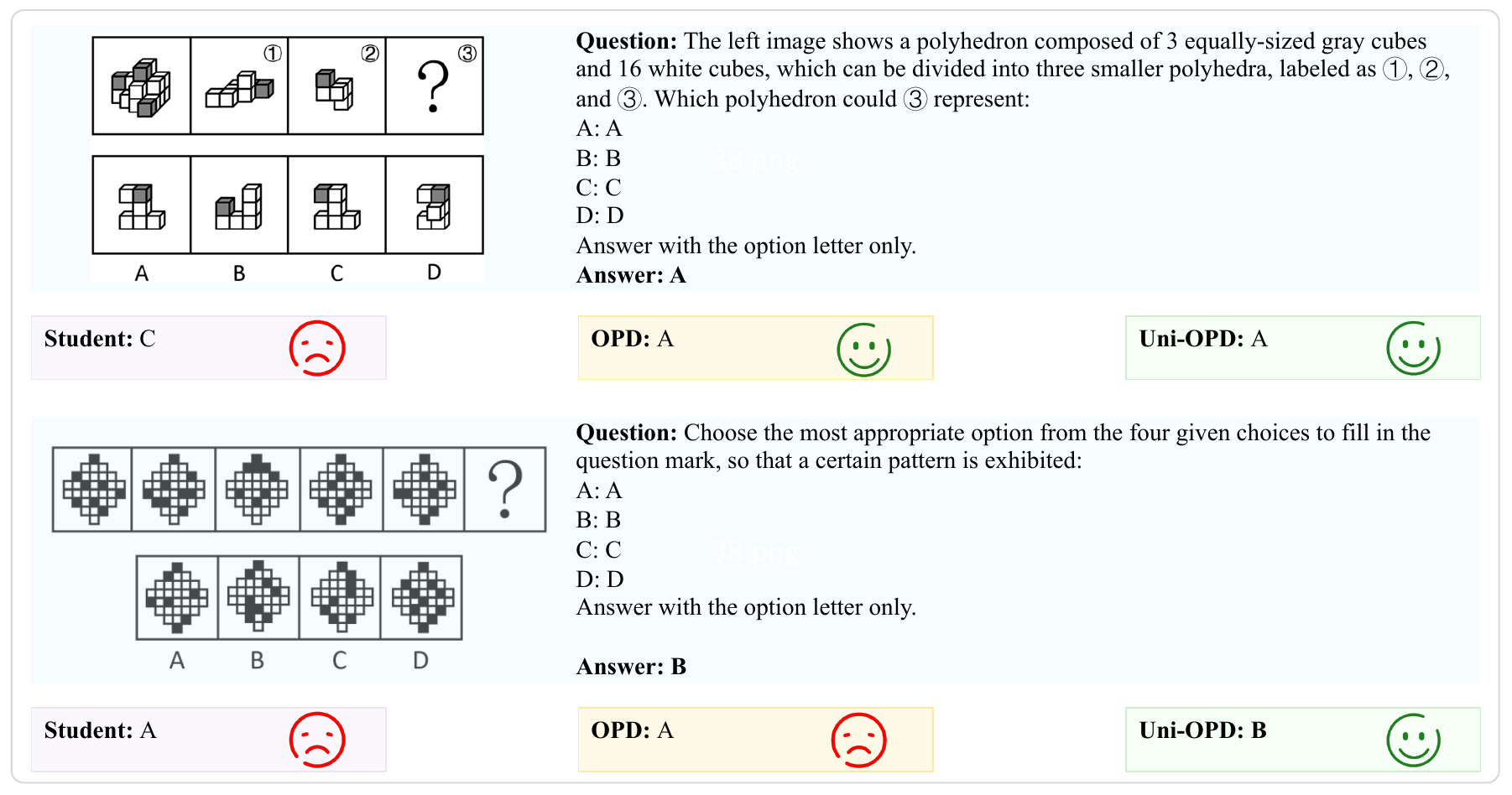}
    \vspace{-1.6em}
    \caption{
        \textbf{Example output of VisuLogic.}
        \texttt{Uni-OPD} correctly answers both questions, demonstrating that our training recipe encourages student exploration to improve its ability for challenging visual reasoning problems.
    }
    \label{fig:example_visulogic}
\end{figure}

\begin{figure}
    \centering
    \includegraphics[width=1.0\linewidth]{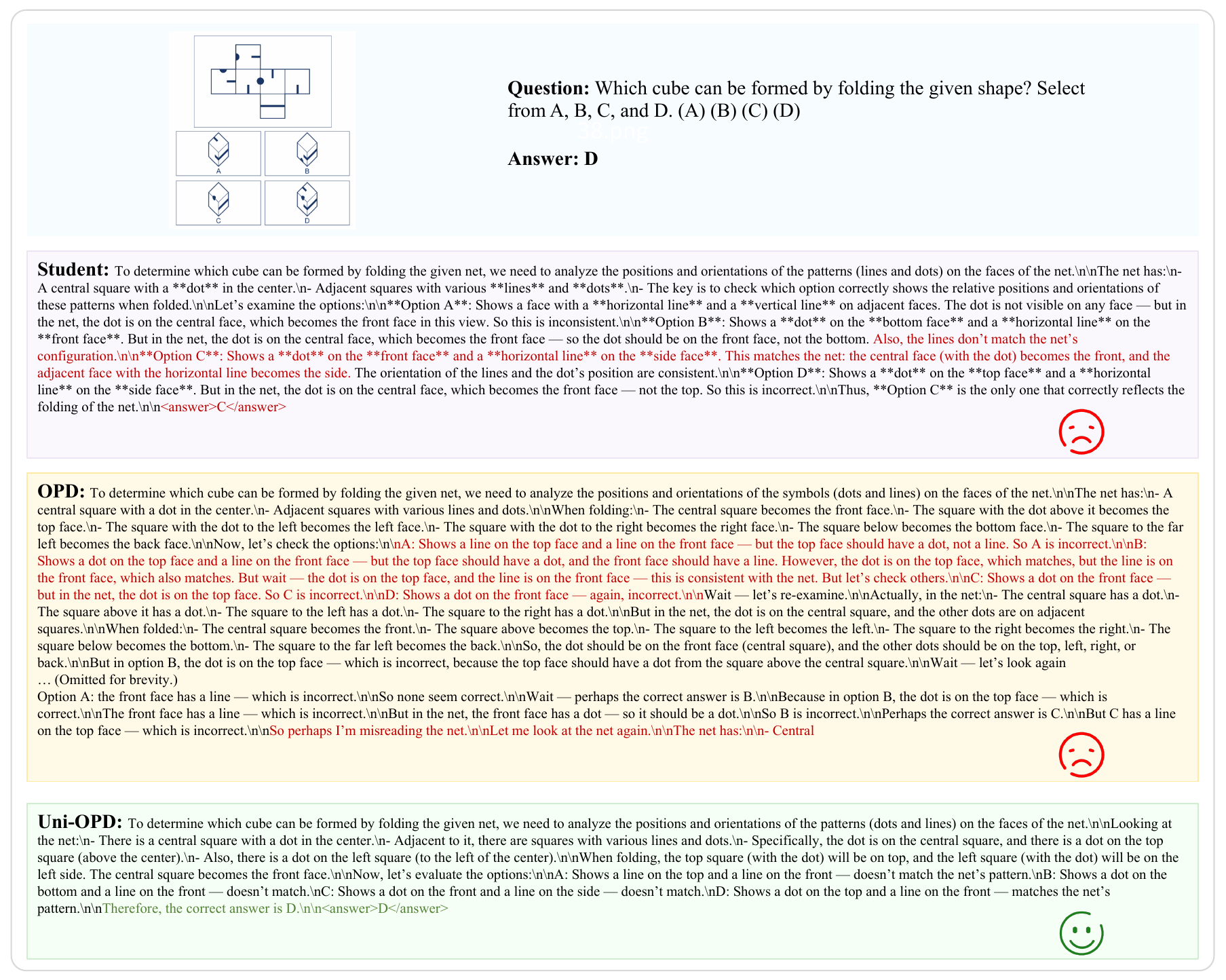}
    \vspace{-1.6em}
    \caption{
        \textbf{Example output of LogicVista.}
        On this challenging visual pattern reasoning puzzle, both the Student model and OPD fail to produce a final answer due to overthinking. \texttt{Uni-OPD}, however, identifies the correct pattern and selects the right answer.
    }
    \label{fig:example_logicvista_uniopd_correct}
\end{figure}

\begin{figure}
    \centering
    \includegraphics[width=1.0\linewidth]{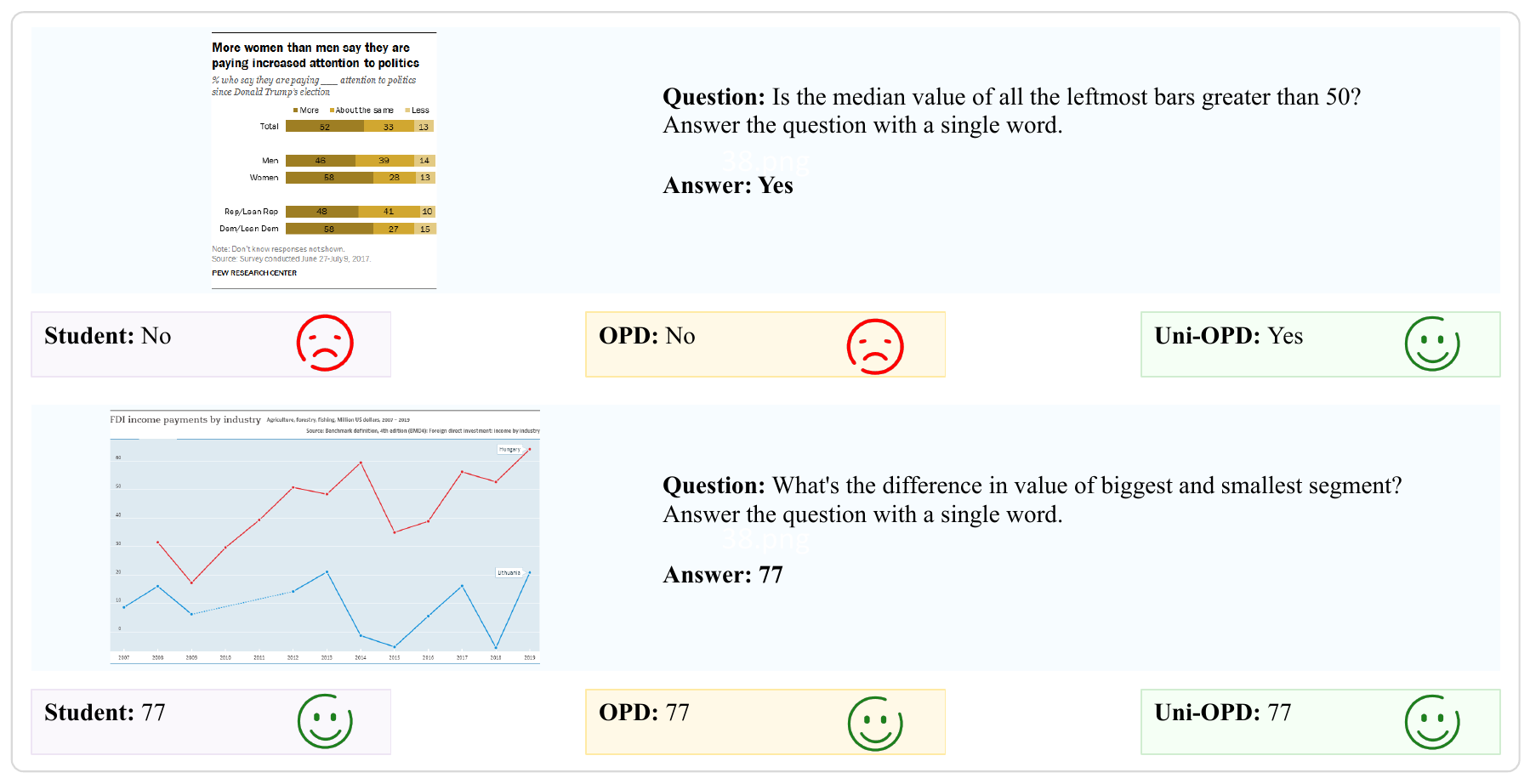}
    \vspace{-1.6em}
    \caption{
        \textbf{Example output of ChartQA.}
        All models answer the simpler chart question correctly, while only \texttt{Uni-OPD} answers the more complex one correctly.
    }
    \label{fig:example_chartqa}
\end{figure}

\begin{figure}
    \centering
    \includegraphics[width=1.0\linewidth]{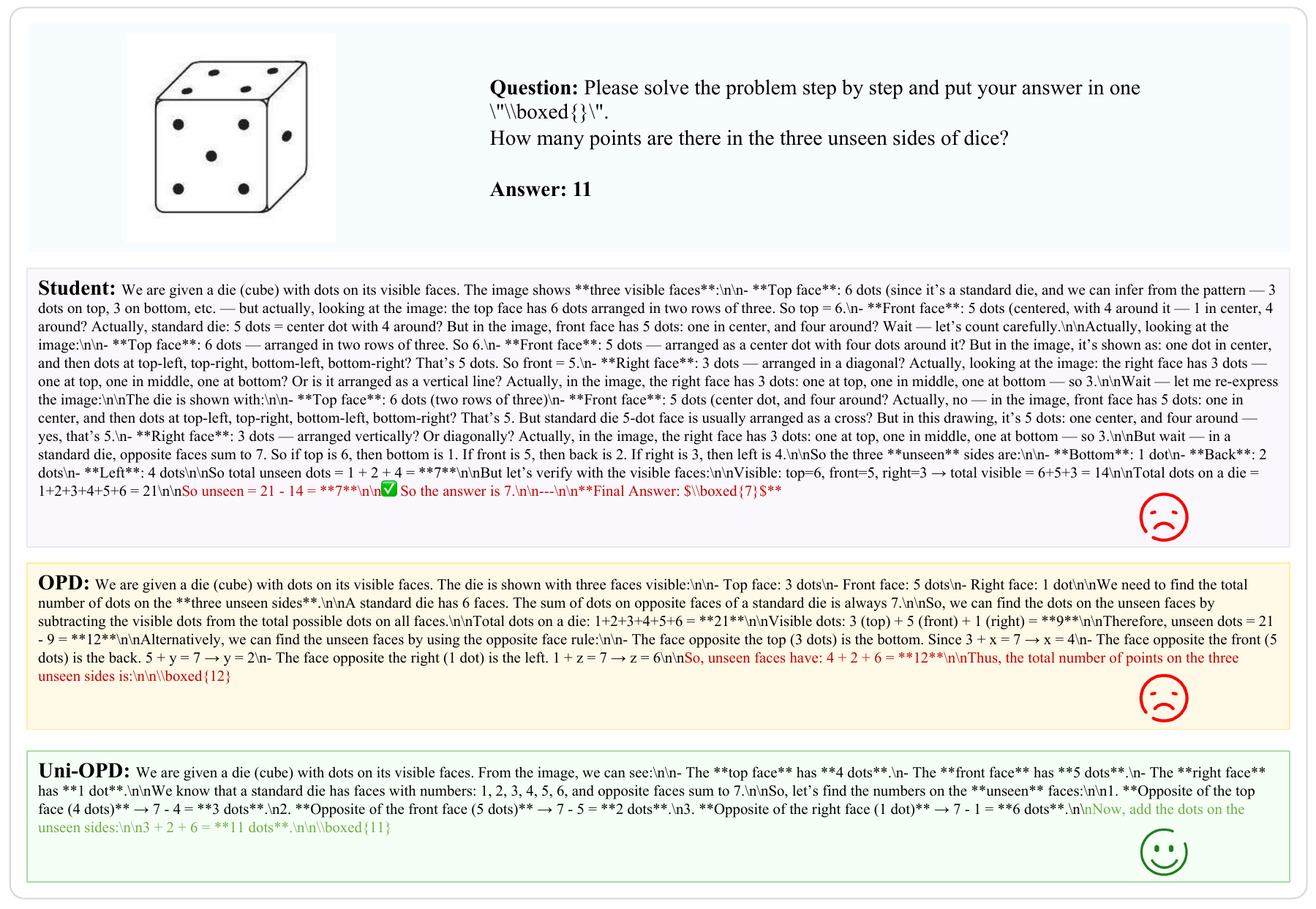}
    \vspace{-1.6em}
    \caption{
        \textbf{Example output of MathVision.}
        All three models follow the required format, but only \texttt{Uni-OPD} produces correct reasoning and reaches the right answer.
    }
    \label{fig:example_mathvision_uniopd_correct}
\end{figure}

\begin{figure}
    \centering
    \includegraphics[width=1.0\linewidth]{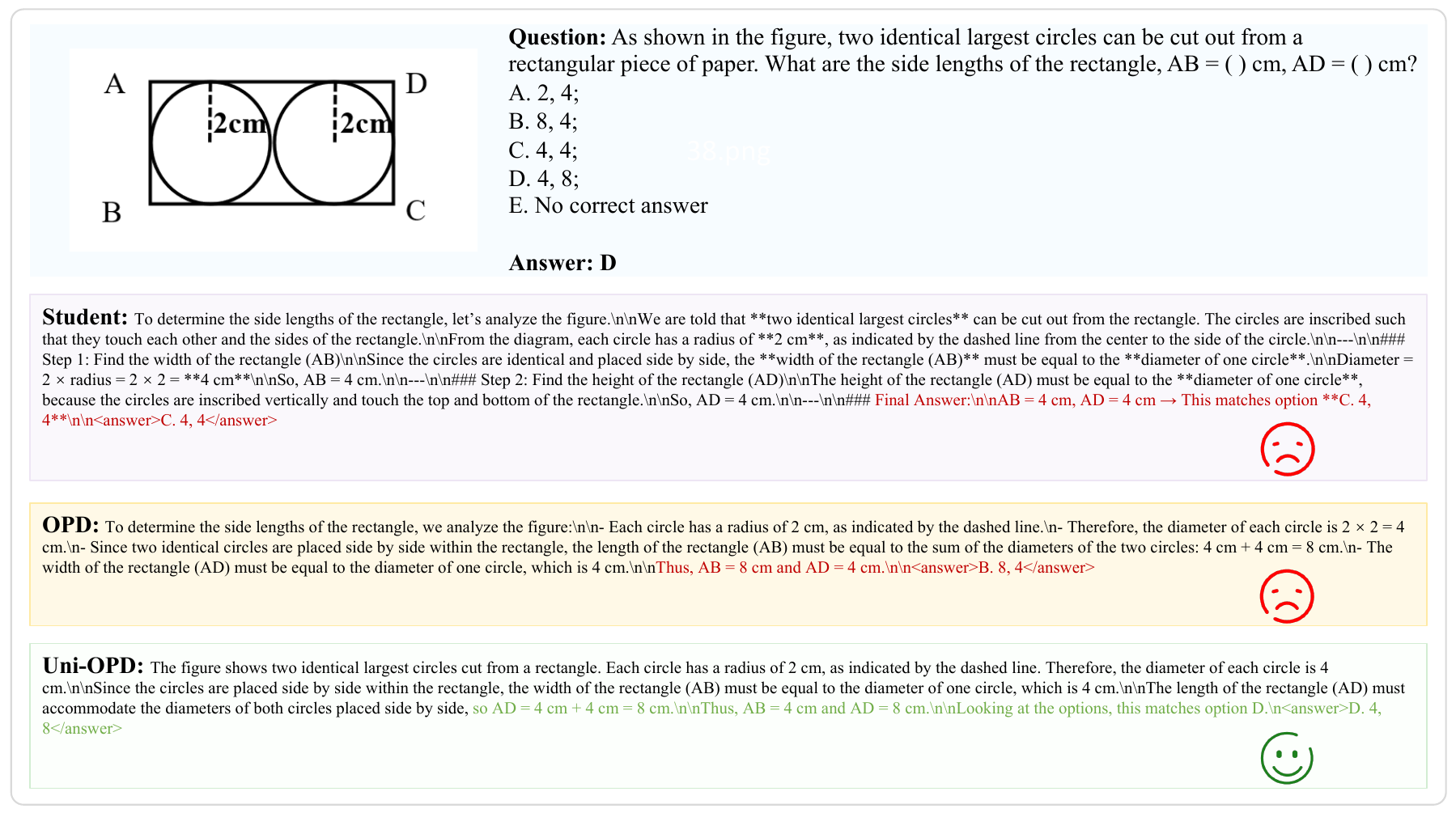}
    \vspace{-1.6em}
    \caption{
        \textbf{Example output of WeMath.}
        This geometry problem requires correctly identifying which side accommodates two circle diameters.
        Both the Student model and OPD confuse the orientation of AB and AD, while \texttt{Uni-OPD} correctly answers the question.
    }
    \label{fig:example_wemath_uniopd_correct}
\end{figure}

Across the multimodal case studies in~\crefrange{fig:example_logicvista_uniopd_correct_effective}{fig:example_wemath_uniopd_correct}, our observations reveal three consistent patterns: (a) \texttt{Uni-OPD} demonstrates superior efficiency on complex reasoning problems, producing more concise outputs while maintaining correctness, whereas the Student model and standard OPD frequently generate excessively long responses that are truncated before reaching a final answer;
(b) \texttt{Uni-OPD} achieves higher correctness than the Student model, often succeeding on questions where the Student model fails; and
(c) Our data-balancing strategies encourage exploration of informative student-generated states during training, improving \texttt{Uni-OPD}'s ability to tackle challenging visual and mathematical reasoning problems that the Student model cannot solve on its own.

\end{document}